\documentclass[twoside,11pt]{article}

\usepackage[T1]{fontenc}
\usepackage{amsmath,amsthm,mathtools}
\usepackage[preprint,nohyperref]{jmlr2e}
\usepackage{booktabs,microtype,array,float,enumitem,longtable}
\usepackage{etoc}
\usepackage[most]{tcolorbox}
\usepackage[colorlinks=false,allbordercolors={1 1 1}]{hyperref}
\hypersetup{
  pdftitle={Denoising Is Stein: Moment Access and a Resource Hierarchy for Learning Hidden Structure in Diffusion Models},
  pdfauthor={Rui Wang}
}
\AtBeginEnvironment{longtable}{\setlength{\tabcolsep}{2pt}}

\newtcolorbox{roadmap}[1][]{enhanced,breakable,colback=black!3,colframe=black!25,
  boxrule=0.5pt,arc=1.5pt,left=6pt,right=6pt,top=4pt,bottom=4pt,
  fonttitle=\bfseries\small,coltitle=black,colbacktitle=black!8,
  title={Roadmap#1},fontupper=\small}
\newcommand{\step}[1]{\par\smallskip\noindent\textbf{#1.}\enspace}
\numberwithin{equation}{section}
\numberwithin{theorem}{section}
\theoremstyle{definition}
\newtheorem{assumption}{Assumption}

\newcommand{\E}{\mathbb E}
\newcommand{\Prob}{\mathbb P}
\newcommand{\R}{\mathbb R}
\newcommand{\Id}{I}
\newcommand{\Sph}{\mathbb S}
\newcommand{\norm}[1]{\left\lVert #1\right\rVert}
\newcommand{\ip}[2]{\langle #1,#2\rangle}
\newcommand{\inner}[2]{\langle #1,#2\rangle}
\newcommand{\risk}{\mathcal R}
\newcommand{\Aff}{\mathcal A}
\newcommand{\Stein}{\mathcal T}
\newcommand{\He}{\mathrm{He}}
\DeclareMathOperator{\tr}{tr}
\DeclareMathOperator{\dvg}{div}
\DeclareMathOperator{\spanop}{span}

\newcommand{\HeightCells}{1116}
\newcommand{\HeightViolations}{0}
\newcommand{\HeightImproved}{265}
\newcommand{\HeightMaxRatio}{99\%}
\newcommand{\HeightAmbientCells}{30}
\newcommand{\HeightAmbientMin}{32\%}
\newcommand{\HeightAmbientMax}{98\%}

\ShortHeadings{Denoising Is Stein}{Rui Wang}
\firstpageno{1}

\begin{document}
\title{Denoising Is Stein: Moment Access\\
and a Resource Hierarchy for Learning\\
Hidden Structure in Diffusion Models}
\author{\name Rui Wang \email wangrui@microsoft.com\\
  \addr Vancouver, BC, Canada}
\maketitle
\thispagestyle{plain}

\begin{abstract}
We study score learning from denoising pairs with a hidden non-Gaussian
component. In one regime covariance reveals it; in another, isotropic
covariance hides it until the first distinguishing moment, of order \(k^\ast\).
Denoising-loss correlations are Stein discrepancies against a baseline
and, for polynomial fields, exactly encode moment differences.
Averaging noisy views of each clean point gives exact moment queries.

In the covariance-matched regime, orientation-independent dictionaries
satisfying an \(L^4/L^2\) condition need order \(d^{k^\ast-1}\) features.
The stated dot-product kernel ridge models need the same power in samples,
as does small-step online SGD on one unit up to logarithms. A two-stage
learner uses Gaussian-smoothed-loss curvature to recover the direction,
then sharpens it and fits a scalar readout. For even \(k^\ast\),
it obtains a fixed score gain over the Gaussian baseline from
\(O(d^{k^\ast/2}\log d)\) clean points. In the covariance-visible regime,
the frozen-feature threshold is \(d^J/J!\); projection onto principal
components makes fitted-model, score-risk, and generated-law distributions
independent of ambient dimension.

Experiments are consistent with the spectral transition. A width-8 network
trained with clean-point reuse shows a \(d^2\)-like
training-set threshold over \(d=64\)--\(192\), with fixed validation and
calibration overhead.
\end{abstract}
\begin{keywords}
diffusion models, denoising score matching, Stein discrepancies,
feature learning, sample complexity
\end{keywords}

\etocdepthtag.toc{mtmain}
\etocsettagdepth{mtmain}{subsection}
\etocsettagdepth{mtappendix}{section}
\section{Introduction}\label{sec:intro}

Diffusion models estimate the score \(\nabla\log p_t\) as Gaussian noise
progressively obscures a data distribution
\citep{ho2020ddpm,song2021sde}. Training provides \emph{denoising pairs}
\((X_0,X_t)\): a clean point and a noisy view. The learner predicts \(X_0\)
from \(X_t\), thereby estimating the score
\citep{vincent2011connection}. If the data have a hidden non-Gaussian
direction, a successful denoiser must detect it through those pairs. The
amount of data needed depends on what the learner can represent and whether
it can use repeated views of the same clean point.

Existing theory studies parts of this question under different data models
and resource measures. Lower bounds for random features and kernels count
features or samples for fixed representations
\citep{ghorbani2021linearized,mei2022hypercontractivity}; analyses of trained
denoisers count fresh pairs for one architecture and optimizer
\citep{bardone2026easyhard}; statistical-query and cryptographic lower bounds
count queries or computation for arbitrary algorithms
\citep{dks2017,song2024crypto}. These results do not directly compare the
cost of learning the same hidden signal across learner classes. We ask:

\begin{quote}
\emph{Which moments of a hidden structure can a denoising learner access, and
how do representation size, distinct clean samples, noisy views and the
learner's access to geometry determine the cost of improving on the Gaussian
score?}
\end{quote}

We study two versions of one hidden-component model. Data are
\(X_0=UZ+\sigma_0(\Id-UU^\top)G_0\) with an unknown orthonormal
\(U\in\R^{d\times k}\), a non-Gaussian latent vector \(Z\in\R^k\) and an
independent Gaussian \(G_0\). In the \emph{covariance-visible} regime
(\(\sigma_0=0\)), the data lie on the subspace, their covariance reveals \(U\),
and principal components recover it from \(k+1\) points. In the
\emph{covariance-matched} regime, \(\mathrm{Cov}(Z)=\sigma_0^2\Id_k\), so
\(\mathrm{Cov}(X_0)=\sigma_0^2\Id_d\) carries no information about \(U\). The
first informative moment has order \(k^\ast\ge3\), the
\emph{generative exponent} of the latent law. This is non-Gaussian component
analysis \citep{dks2017,tan2018ngca} seen through a diffusion model.

\subsection{Main results}

\begin{figure}[t]
\centering
\includegraphics[width=\linewidth]{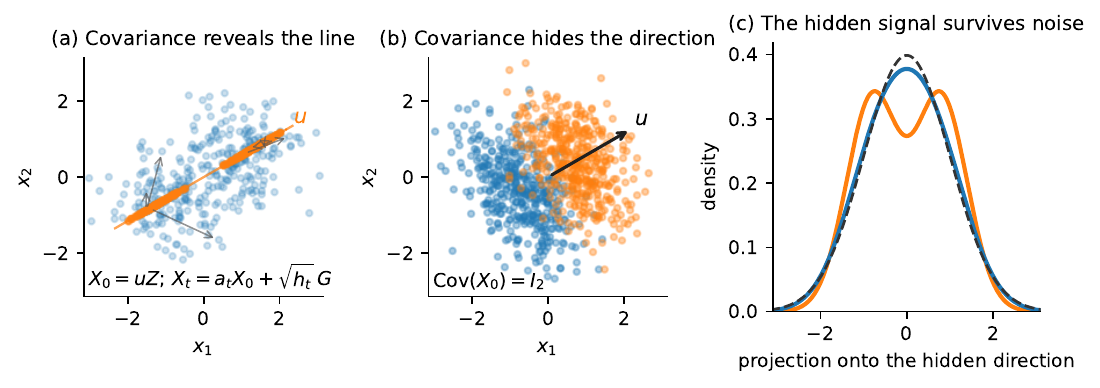}
\caption{Two hidden-component data regimes in two dimensions. (a) Clean
points \(X_0=uZ\) lie on a line; blue points are their noisy views at \(t=0.3\),
and short arrows connect selected pairs. The line is visible in the clean
covariance. (b) The clean covariance is \(I_2\). Blue and orange mark the
mixture components only for illustration; their labels are not observed by
the learner. (c) In the covariance-matched example, the clean marginal along
\(u\) (orange) differs from its moment-matched Gaussian (dashed black) even
though their means and variances agree. Diffusion at \(t=0.3\) weakens that
difference (blue).}
\label{fig:problem}
\end{figure}

\begin{figure}[t]
\centering
\includegraphics[width=0.82\linewidth]{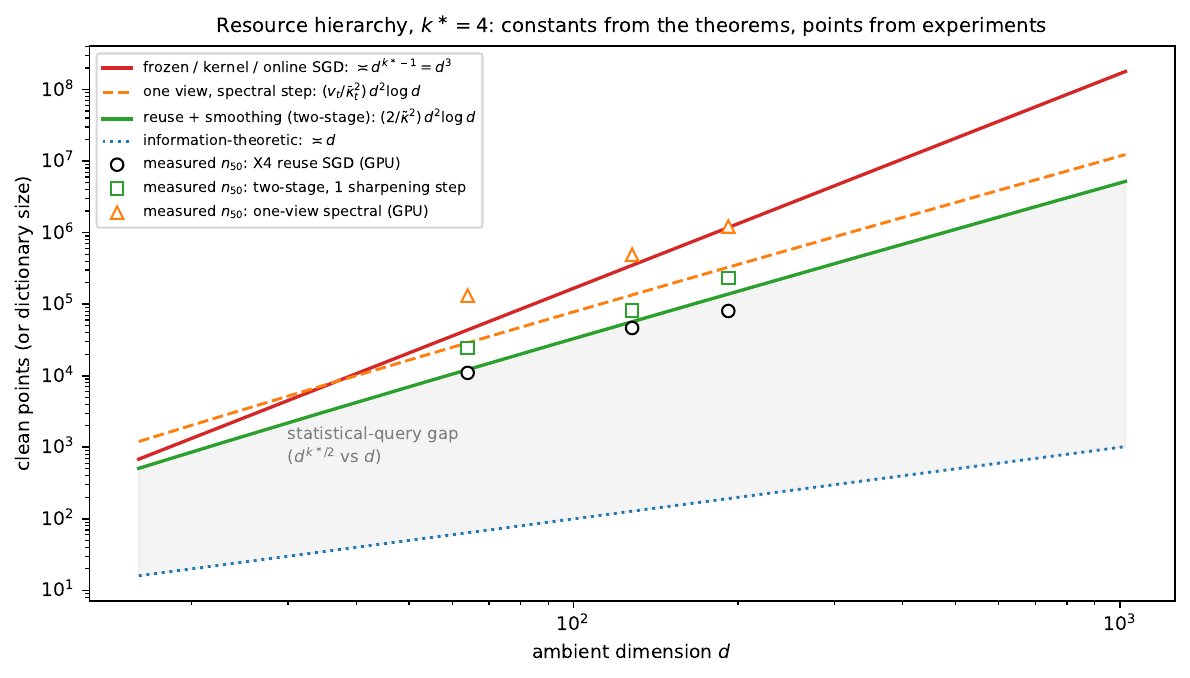}
\caption{The covariance-matched rows of Table~\ref{tab:hierarchy} on one scale,
for \(k^\ast=4\) and the two-component mixture with \(\mu=0.8\). Lines are the
leading terms of the theorems, with the constants \(2/\tilde\kappa^2\) (full reuse)
and \(v_t/\bar\kappa_t^2\) (one view, at \(t^\ast\)) of
Theorem~\ref{thm:multiview}; the logarithm is normalized to one at \(d=64\). The
red line is the dictionary-size threshold \(d^3/6\) of the frozen barrier, whose
exponent the kernel and online-SGD lower bounds share. Markers are measured 50\%
training-set thresholds from Section~\ref{sec:exp-cm}; they exclude the fixed
\(65{,}536\) validation and calibration points, and cover only \(d=64\)--\(192\).}
\label{fig:hierarchy}
\end{figure}

Figure~\ref{fig:problem} shows the observable difference between the two data
regimes: a line of clean points in one case, and an isotropic covariance that
masks a non-Gaussian direction in the other. Table~\ref{tab:hierarchy}
summarizes the resource costs for each learner. The results follow from a
moment-access calculus for the denoising loss (Section~\ref{sec:moments});
Figure~\ref{fig:hierarchy} compares the leading scales.

\begin{table}[t]
\centering\small
\caption{The resource hierarchy for improving the score at a fixed diffusion
time by a constant fraction of the Gaussian baseline's excess risk, with
\(k=1\) hidden direction. \(M\): dictionary size; \(n\): distinct clean points;
\(r\): noisy views per clean point. Costs are up to constants depending on the
latent law and the time, but not on \(d\). Every theorem row counts all clean
points the learner uses; the two-stage learner's total is
\(O(d^{k^\ast/2}\log d)+O(d)+O(1)\). The empirical row counts training points only;
that learner also uses a fixed set of \(65{,}536\) validation and calibration
points. Lower bounds hold for the stated class and under the stated hypotheses.
``Proved'' rows are theorems of this paper unless marked as imported.}
\label{tab:hierarchy}
\begin{tabular}{p{4.9cm}p{1.5cm}p{4.1cm}p{3.4cm}}
\toprule
Learner & Resource & Cost & Status \\
\midrule
\multicolumn{4}{l}{\emph{Covariance-matched regime, generative exponent \(k^\ast\) (even where stated)}}\\
Polynomial fields of degree \(<k^\ast-1\) & any & no gain at any size & proved (Cor.~\ref{cor:polybarrier})\\
Orientation-blind dictionary with bounded \(L^4/L^2\) constant & \(M\) & \(\Theta(d^{k^\ast-1})\) & proved under the \(L^4/L^2\) hypothesis (Thm.~\ref{thm:frozen}); open for random-feature spans\\
Dot-product kernel ridge regression on noisy inputs & \(n\) & \(\Omega(d^{k^\ast-1})\) & proved for ridge \(\gg d^{-(k^\ast-1)}\); at every ridge for the exponential kernel (Thm.~\ref{thm:kernel}, Lem.~\ref{lem:selfreg})\\
Online spherical SGD, one denoiser unit, small steps & \(n=\) steps & \(d^{k^\ast-1}\) up to \(\log^2d\) & proved via \citet{bgj2021} (Cor.~\ref{cor:online})\\
\(L^2\) statistical queries & tolerance & \(d^{-k^\ast/4+o(1)}\); the sample analogue \(n\gtrsim d^{k^\ast/2}\) is heuristic & tolerance bound proved under an output-norm assumption (Prop.~\ref{prop:sq})\\
One gradient step of a matrix score model, \(r\) views & \(n\) & \(O\bigl((1+\tfrac{\iota_t-1}r)\,d^{k^\ast/2}\log d\bigr)\) & proved (Thm.~\ref{thm:multiview})\\
Two-stage reuse learner (smoothed curvature, output sharpening, tied readout) & \(n\) & \(O(d^{k^\ast/2}\log d)\) & proved (Thm.~\ref{thm:twostage})\\
Reuse SGD on a width-8 network, \(k^\ast=4\) & training \(n\) & \(n_{50}\approx2.2\)--\(2.8\,d^2\) at \(d=64\)--\(192\), plus \(65{,}536\) fixed validation and calibration points & empirical \(d^2\)-like trend over this range (Sec.~\ref{sec:exp-cm}); mechanism open\\
\midrule
\multicolumn{4}{l}{\emph{Covariance-visible regime, product-mixture latents (\(k^\ast=4\) after the fitted Gaussian)}}\\
Orientation-blind dictionary & \(M\) & threshold location \(d^3/6\), sharp for this class; exact height for \(k=1\) & proved (Thms.~\ref{thm:floor}, \ref{thm:sharp}, \ref{thm:height})\\
Random features on the principal subspace & --- & laws of model, risk and samples free of \(d\) & proved (Thms.~\ref{thm:invariance}, \ref{thm:generation})\\
\bottomrule
\end{tabular}
\end{table}

\paragraph{Contributions.}
The Stein--Tweedie identity expresses each residual correlation as a
discrepancy between the
noisy law and a baseline (Theorem~\ref{thm:stein}); on polynomial fields,
these correlations recover exactly the corresponding moment differences
(Theorem~\ref{thm:bijection}). Thus a fitted Gaussian baseline removes the
moments it matches, shifting the first accessible signal in the
covariance-visible mixture from degree two to degree four
(Theorem~\ref{thm:baseline}, Corollary~\ref{cor:shift}). Conditioning on the
clean point gives an exact clean-data moment query
(Lemma~\ref{lem:condstein}). Averaging fresh noisy views of one point
reduces the forward-process variance of the corresponding gradient estimate.
Diffusion acts on the hidden
coordinate through an Ornstein--Uhlenbeck noise operator; the static
signal-to-noise ratio of the best low-degree correlation peaks at the time in
Theorem~\ref{thm:time}.

For an orientation-independent dictionary, the covariance-visible mixture
has a sharp feature threshold at \(d^3/6\), with higher odd-degree steps at
\(d^J/J!\) (Theorems~\ref{thm:floor}, \ref{thm:sharp}). For one hidden
direction, a one-sided change of measure identifies the height of the first
step (Theorem~\ref{thm:height}). The covariance-matched model gives the
\(d^{k^\ast-1}\) exponent for dictionaries satisfying the stated
\(L^4/L^2\) condition (Theorem~\ref{thm:frozen}). Kernel ridge regression on
noisy inputs has the same sample exponent under the conditions of
Theorem~\ref{thm:kernel}; for the exponential kernel this includes every
ridge (Lemma~\ref{lem:selfreg}). When covariance reveals the subspace,
principal-component projection instead yields fitted-model, score-risk and
generated-law distributions independent of ambient dimension
(Theorems~\ref{thm:invariance}, \ref{thm:generation}).

Learning the covariance-matched direction can be cheaper than fixing a
representation in advance. Small-step online SGD on one unit needs
\(d^{k^\ast-1}\) steps up to \(\log^2d\) (Corollary~\ref{cor:online}), while the
statistical-query bound gives tolerance \(d^{-k^\ast/4+o(1)}\) under its
output-norm assumption (Proposition~\ref{prop:sq}). Averaging \(r\) noisy
views of each clean point, one gradient step of the matrix score model
recovers the direction from
\(O((1+(\iota_t-1)/r)d^{k^\ast/2}\log d)\) clean points for even
\(k^\ast\) (Theorem~\ref{thm:multiview}). The same partial-trace matrix is
the curvature created by Gaussian smoothing of a single-unit loss
(Proposition~\ref{prop:smoothing}). A learner that uses this curvature,
sharpens the direction and fits a scalar readout obtains a fixed score gain
with \(O(d^{k^\ast/2}\log d)+O(d)+O(1)\) clean points
(Theorem~\ref{thm:twostage}). For \(k^\ast=4\), this separates it from the
restricted learner classes above by one power of \(d\), up to logarithms
and under their stated hypotheses (Corollary~\ref{cor:dimsep}).

The experiments check the predicted barriers and invariances in the
covariance-visible regime. In the covariance-matched regime, the partial
trace's measured overlaps are consistent with the proposed spectral
transition within Monte Carlo error.
A width-8 network trained with clean-point reuse has a \(d^2\)-like
training-set threshold over \(d=64\)--\(192\), with fixed validation and
calibration overhead (Section~\ref{sec:experiments}).

\subsection{Related work}\label{sec:related}

\paragraph{Low-dimensional structure in diffusion models.}
\citet{chen2023lowdim} give score approximation and estimation rates for data on
a linear subspace, \citet{oko2023minimax} prove minimax optimality of diffusion
estimators over Besov classes, and \citet{li2024adapting} show that samplers
adapt to unknown intrinsic dimension. Given an accurate score, sampling error is
controlled by \citet{chen2023sampling} and \citet{benton2024dlinear}. These are
upper bounds for adaptive estimators; our lower bounds identify what restricted
classes lack, and our upper bounds count each resource separately.

\paragraph{Fixed representations and kernels.}
Random-feature and kernel models in high dimension fit only polynomial
components of degree \(\ell\) with \(d^\ell\ll N\) features or samples
\citep{ghorbani2019limitations,ghorbani2021linearized,mei2022hypercontractivity},
and the spectrum of inner-product kernel matrices in polynomial regimes is
described by \citet{misiakiewicz2022spectrum} and \citet{lu2022equivalence}.
For denoising score matching, \citet{george2026randomfeatures} derive precise
learning curves for random features and \citet{george2026manifold} extend them
to manifold data. Here the inputs are non-Gaussian along a hidden direction,
and the barrier applies to every orientation-blind dictionary with arbitrary
coefficients. For one hidden direction, it also gives the exact risk height
relative to the moment-matched Gaussian denoiser.

\paragraph{Exponents, statistical queries and non-Gaussian component analysis.}
The information exponent governs online SGD for single-index models
\citep{bgj2021}, and the generative exponent governs statistical-query and
low-degree algorithms \citep{damian2024gaps}. Statistical-query lower bounds
for non-Gaussian component analysis are due to \citet{dks2017} and
\citet{dkrs2024}, spectral algorithms to \citet{vempala2011structure} and
\citet{tan2018ngca}, and score estimation for such data is cryptographically
hard in general \citep{song2024crypto}. \citet{bardone2026easyhard} show that a
denoiser trained online on a mixed cumulant model learns pair-wise statistics
at linear cost and the fourth cumulant at cubic cost, defining a diffusion
information exponent; \citet{szekely2024spiked} report a quadratic sample
threshold for supervised networks in a spiked cumulant model. Our
Theorem~\ref{thm:bijection} identifies the generative exponent as the
relevant one for denoising, and
Corollary~\ref{cor:shift} gives a mechanism for the staged learning of
\citet{bardone2026easyhard}.

\paragraph{Tensor methods, smoothing and reuse.}
The fourth-moment tensor of non-Gaussian component data is a spiked tensor, for
which unfolding reaches \(d^{k/2}\) \citep{montanari2014tensor} and gradient
descent or Langevin dynamics from random starts needs \(d^{k-1}\)
\citep{bgj2020tensor}. Gaussian smoothing reaches the unfolding threshold for
tensor PCA \citep{anandkumar2017homotopy} and the \(d^{k/2}\) rate for online
single-index learning \citep{damian2023smoothing}; SGD can be viewed as
optimizing a smoothed objective \citep{kleinberg2018alternative}. Reusing
batches changes what gradient methods can learn in supervised two-layer
networks \citep{dandi2024reuse,lee2024neural}. A link between one gradient step of a
diffusion objective and a spectral method was made for a covariance spike by
\citet{shah2023ddpm}. Here the conditional Stein identity connects reuse to
the denoising loss, and the curvature created by smoothing is the
partial-trace statistic. The resulting two-stage learner has explicit costs
for clean points, noisy views and arithmetic.

\paragraph{Stein's method and score matching.}
Theorem~\ref{thm:stein} is a standard consequence of Stein's lemma and
Tweedie's formula \citep{robbins1956empirical,efron2011tweedie}; its supremum
over a unit ball of fields is a Stein discrepancy \citep{gorham2015stein}, and
the equivalence of denoising score matching with score matching is due to
\citet{hyvarinen2005scorematching} and \citet{vincent2011connection}. We use
this identity to compare what different denoising learners can extract from
the same hidden-component model.
\section{Setting, learners and resources}\label{sec:setting}

\subsection{Diffusion pairs, denoisers and baselines}\label{sec:pairs}

Fix a time \(t>0\) and write \(a=a_t=e^{-t}\), \(h=h_t=1-e^{-2t}\). For clean data
\(X_0\sim P_0\) on \(\R^d\) with all moments finite and an independent
\(G\sim N(0,\Id_d)\), the noisy view is
\[
 X_t=a_tX_0+\sqrt{h_t}\,G ,
\]
the time-\(t\) marginal of the Ornstein--Uhlenbeck process
\(dX_t=-X_t\,dt+\sqrt2\,dB_t\). Write \(P_t\) for its law, \(p_t\) for its density,
\(m_t(x)=\E[X_0\mid X_t=x]\) for the denoiser and \(s_t=\nabla\log p_t\) for the
score. Tweedie's formula \citep{robbins1956empirical,efron2011tweedie} gives
\begin{equation}\label{eq:tweedie}
 s_t(x)=\frac{a_tm_t(x)-x}{h_t},
\end{equation}
and every denoiser estimate \(\widehat m\) defines a score estimate
\(\widehat s=(a_t\widehat m-x)/h_t\). Then
\(\E\|\widehat s(X_t)-s_t(X_t)\|^2=(a_t/h_t)^2\,\E\|\widehat m(X_t)-m_t(X_t)\|^2\), so
score risk and denoiser risk differ by the factor \((a_t/h_t)^2\). Regressing
\(X_0\) on \(X_t\) is denoising score matching \citep{vincent2011connection}: for
any \(f\),
\(\E\|X_0-f(X_t)\|^2=\E\|X_0-m_t(X_t)\|^2+\E\|m_t(X_t)-f(X_t)\|^2\).

\paragraph{Baselines and the Stein operator.}
A \emph{baseline} is a law \(Q_0\) with finite moments, with diffused law
\(Q_t\), smooth positive density \(q_t\) and denoiser \(m^Q_t\). Learners in this
paper improve on a baseline: they fit \(m^Q_t+f\) and are judged by how much of
the residual \(m_t-m^Q_t\) the correction \(f\) captures. The Langevin--Stein
operator of \(Q_t\) acts on \(C^1\) vector fields by
\(\Stein_{Q_t}\varphi=\dvg\varphi+\inner{\nabla\log q_t}{\varphi}\); for
\(Q_t=N(\bar m,C)\), \(\Stein_{Q_t}\varphi=\dvg\varphi-\inner{C^{-1}(x-\bar m)}{\varphi}\).

\paragraph{Hermite notation.}
\(\He_j\) is the probabilists' Hermite polynomial and \(h_j=\He_j/\sqrt{j!}\)
its orthonormal version under \(N(0,1)\). For \(x\in\R^d\), \(\He_j(x)\) also
denotes the degree-\(j\) Hermite tensor, the symmetric \(j\)-tensor-valued
polynomial with \(\He_j(x)[w^{\otimes j}]=\He_j(w^\top x)\) for unit \(w\).
Under \(N(0,\Id)\) it satisfies the isometry
\(\E\,\He_j(x)[A]\,\He_j(x)[B]=\sum_\pi\inner A{\pi B}\), summed over the \(j!\)
permutations of tensor slots. For even \(k^\ast=2p+2\) the \emph{partial trace}
is the matrix
\[
 M_p(x)=\He_{k^\ast}(x)\bigl[\,\cdot\,,\cdot\,,\Id^{\otimes p}\bigr]\in\R^{d\times d};
 \qquad
 M_1(x)=\mathcal H(x)=(|x|^2-d-4)\,xx^\top-(|x|^2-d-2)\,\Id .
\]

\subsection{Two data regimes}\label{sec:regimes}

\begin{assumption}[Hidden component]\label{ass:hidden}
Fix \(1\le k<d\), an orientation \(U\in\R^{d\times k}\) with orthonormal columns
\(u_1,\dots,u_k\), \(\sigma_0\ge0\), and a law \(\nu\) on \(\R^k\) with mean zero
and all moments finite. Clean data are
\(X_0=UZ+\sigma_0(\Id-UU^\top)G_0\) with \(Z\sim\nu\) and \(G_0\sim N(0,\Id_d)\)
independent.
\end{assumption}

\begin{assumption}[Covariance-visible regime]\label{ass:model}
In Assumption~\ref{ass:hidden}, \(\sigma_0=0\), so \(X_0=UZ\), and \(Z\) has
independent coordinates \(Z_j=\epsilon_j\mu+\tau\xi_j\) with \(\mu,\tau>0\),
\(\epsilon_j\) uniform on \(\{\pm1\}\) and \(\xi_j\sim N(0,1)\). Write
\(P_t^U\) for the law of \(X_t\).
\end{assumption}

\begin{assumption}[Covariance-matched regime]\label{ass:cm}
In Assumption~\ref{ass:hidden}, \(\sigma_0>0\) and \(\mathrm{Cov}(Z)=\sigma_0^2\Id_k\),
so that \(\mathrm{Cov}(X_0)=\sigma_0^2\Id_d\) and the isotropic Gaussian
\(Q_0=N(0,\sigma_0^2\Id_d)\), which does not depend on \(U\), is the
moment-matched baseline. The \emph{generative exponent} of \(\nu\) is
\[
 k^\ast(\nu)=\min\bigl\{j\ge1:\ \E_\nu[Z^{\otimes j}]\ne\E[(\sigma_0\xi)^{\otimes j}]\bigr\},
 \qquad \xi\sim N(0,\Id_k).
\]
Unless stated otherwise \(k=1\), \(U=u\), and we write
\(\bar\sigma_t^2=a_t^2\sigma_0^2+h_t\), \(\lambda_t=a_t\sigma_0/\bar\sigma_t\in(0,1)\), and
\(\tilde\kappa=\kappa_{k^\ast}(\nu)/\sigma_0^{k^\ast}\) for the normalized
\(k^\ast\)-th cumulant.
\end{assumption}

For the symmetric two-component mixture
\(\nu=\frac12N(-\mu,\tau^2)+\frac12N(\mu,\tau^2)\) with \(\mu^2+\tau^2=\sigma_0^2\),
the odd moments vanish, the variance matches and the fourth cumulant is
\(-2\mu^4\), so \(k^\ast=4\). With \(\sigma_0=1\) and \(\mu=0.8\),
\(\kappa_4=-0.8192\). The two regimes use the same latent family; what differs is
whether the covariance carries the orientation. Relative to an isotropic
baseline, the covariance-visible data have exponent two; relative to the
Gaussian fitted to their first two moments they have exponent four
(Corollary~\ref{cor:shift}). Each theorem states the regime to which it
applies. Under Assumption~\ref{ass:cm} the noisy marginal satisfies
\(u^\top X_t/\bar\sigma_t=\lambda_t(Z/\sigma_0)+\sqrt{1-\lambda_t^2}\,\xi\) with
independent \(\xi\sim N(0,1)\), and its orthogonal coordinates are exactly
\(N(0,\bar\sigma_t^2)\); this is the source of the time action in
Section~\ref{sec:moments}. Figure~\ref{fig:problem} shows the two regimes in
two dimensions. In higher dimensions, an orientation-independent feature
has a typical overlap of order \(d^{-1/2}\) with the hidden direction, which
underlies the frozen-feature bounds of Section~\ref{sec:frozen}.

\subsection{Learners and the resources they spend}\label{sec:resources}

We track four resources separately.
\begin{itemize}[leftmargin=*]
\item \emph{Representation size} \(M\): the dimension of a frozen dictionary.
\item \emph{Distinct clean points}: the total budget counts training, selection
(validation) and readout (calibration) points. Theorems state every part
explicitly (for example \(n_1\), \(n_2\), \(n_3\) in Theorem~\ref{thm:twostage}). In the
experiments, \(n\) denotes the \emph{training-set} size; every learner there also
uses one fixed set of \(32{,}768\) validation and one of \(32{,}768\) calibration
points, independent of \(n\), and the tables report this overhead next to each
training-set threshold.
\item \emph{Noisy presentations}: the total number of views
\((x_0,a_tx_0+\sqrt{h_t}\,g)\) used, with \(r\) views per clean point under
reuse; views of one clean point share \(x_0\) and use independent \(g\).
\item \emph{Computation}: arithmetic operations, reported where relevant.
\end{itemize}
A \(d^3\) barrier on dictionary size \(M\) imposes no \(d^3\) lower bound on
clean points \(n\). An algorithm using \(d^2\) clean points may require many
more noisy presentations.

\begin{definition}[Orientation-independent dictionary]\label{def:dictionary}
A \emph{dictionary} is a random linear space \(\mathcal S=\mathcal S(\omega)\) of
measurable functions \(\R^d\to\R\), of dimension at most \(M\), where \(\omega\) is
a random element whose law does not depend on \(U\) and which is independent of
\(U\) when \(U\) is random. An estimator \(\widehat m(\cdot,t)\) is
\emph{supported} by \(\mathcal S\) if, almost surely, every coordinate function
\(x\mapsto e_i^\top\widehat m(x,t)\) lies in \(\mathcal S+\Aff\), where
\(\Aff=\spanop\{1,x_1,\dots,x_d\}\).
\end{definition}

The element \(\omega\) may contain random weights, the latent training samples,
auxiliary seeds and any rotation-invariant statistic of the training inputs; it
may not contain the orientation. The coefficients that combine dictionary
elements, the affine part, the output directions and any selection among
candidates may all depend on the data and on \(U\). Selecting among several
dictionaries is covered by their sum. We assume that \(\mathcal S\) is spanned
by finitely many random functions that are jointly measurable in
\((\omega,x)\).

\paragraph{Gradient learners.} A gradient learner fits a parametric correction
\(f_\theta\) by descending the empirical denoising loss
\(\widehat L(\theta)=\overline{\|x_0-m^Q_t(x_t)-f_\theta(x_t)\|^2}\), the bar denoting
an average over the views used. \emph{Online} learners draw a fresh clean point
and a fresh view at each step; \emph{reuse} learners revisit \(n\) stored clean
points with fresh views; \emph{full-batch} learners average over all views at
once. Kernel ridge regression on the noisy inputs is defined in
Section~\ref{sec:kernels}.

\paragraph{Risk.} For an estimator \(\widehat s\) and an orientation \(U\), the
score risk at time \(t\) is
\(\risk_t(\widehat s;U)=\E\|\widehat s(X_t,t)-s_t(X_t)\|^2\), with \(X_t\)
independent of the training data and the expectation also over training data
and algorithmic randomness; it is not divided by \(d\). When \(U\) is
Haar-distributed on the Stiefel manifold, independently of everything else,
\begin{equation}\label{eq:haar}
 \bar\risk_t(\widehat s)=\E_U\,\risk_t(\widehat s;U)
\end{equation}
is the \emph{orientation-averaged} risk. In the covariance-matched regime we
write \(B=\|m_t-m^Q_t\|^2_{L^2(P_t)}\), which does not depend on \(u\), so the
moment-matched Gaussian baseline has score risk \((a_t/h_t)^2B\), and a learner
\emph{gains} \(\Delta\) if its denoiser risk is \(B-\Delta\).

\subsection{Estimators in the covariance-visible regime}\label{sec:model}

Training in the covariance-visible regime uses \(n\ge k+1\) clean examples, a
disjoint validation set of \(n_v\) clean examples, and \(r\) independent
corruptions of each example at each fitted time. Test points, generation
references and sampler noise are independent of these.

\paragraph{Shared Gaussian denoiser.} Let \(\widehat U\) be the top-\(k\) right
singular vectors of the centered training matrix and
\(\widehat\Pi=\widehat U\widehat U^\top\). Let \(\widehat\mu\in\R^k\) and
\(\widehat\Sigma\in\R^{k\times k}\) be the mean and covariance of
\(\widehat U^\top X_0^{(i)}\). The fitted Gaussian denoiser is the posterior mean
under \(N(\widehat U\widehat\mu,\widehat U\widehat\Sigma\widehat U^\top)\),
\[
 m_G(x,t)=\widehat U\Bigl[\widehat\mu+a_t\widehat\Sigma
 \bigl(a_t^2\widehat\Sigma+h_t\Id_k\bigr)^{-1}
 \bigl(\widehat U^\top x-a_t\widehat\mu\bigr)\Bigr];
\]
it is affine in \(x\) and available to every estimator.

\paragraph{Random-feature residual denoisers.} Draw \(W\in\R^{d\times p}\) and
\(b\in\R^p\) with independent \(N(0,1)\) entries, independently of all data. For
an input projector \(\Pi\) of rank \(q\) and a bandwidth multiplier \(\beta>0\),
\[
 \varphi_\beta(x)=\Bigl(1,\;p^{-1/2}\tanh\Bigl(\frac{W^\top\Pi x}
 {\sqrt q\,\beta\,\varsigma_t}+b\Bigr)\Bigr)\in\R^{p+1},
 \qquad
 \varsigma_t^2=h_t+\frac{a_t^2\tr\widehat\Sigma}{q}.
\]
The \emph{ambient} dictionary uses \(\Pi=\Id_d\), \(q=d\); the \emph{PCA}
dictionary uses \(\Pi=\widehat\Pi\), \(q=k\); the \emph{oracle} dictionary uses
\(\Pi=UU^\top\), \(q=k\), with the same \(W,b\). Each method fits
\(\widehat m(x,t)=m_G(x,t)+\widehat UB^\top\varphi_\beta(x)\), with
\(B\in\R^{(p+1)\times k}\) minimizing, for each ridge \(\lambda\) in a finite
grid \(\Lambda\),
\[
 \frac1N\sum_{i=1}^N\norm{\widehat U^\top\bigl(x_0^{(i)}-m_G(x_t^{(i)},t)\bigr)
 -B^\top\varphi_\beta(x_t^{(i)})}^2+\lambda\norm{B}_F^2
\]
over the \(N=nr\) corrupted training pairs. The pair
\((\lambda,\beta)\in\Lambda\times\mathcal B\) minimizes the same unpenalized error
on corrupted validation pairs, ties broken by grid order; \(L=|\mathcal B|\). No
method sees population scores or generated samples during tuning. Every method
has exactly \(k(p+1)\) readout coefficients.

\paragraph{Interpolation and sampling.} Denoisers are fitted on a finite
increasing time grid in \([t_0,T]\) and interpolated linearly in \(\log t\);
scores are obtained from~\eqref{eq:tweedie} at the interpolated time and are
never interpolated themselves. The sampler starts from \(x_T\sim N(0,\Id_d)\),
applies Euler--Maruyama steps of the reverse SDE
\(dx=[x+2\widehat s(x,t)]\,dt+\sqrt2\,d\bar B\) on a geometric grid from \(T\) to
\(t_0\), and returns \(\widehat x=e^{t_0}(x_{t_0}+h_{t_0}\widehat s(x_{t_0},t_0))\).
\section{Moment access: denoising correlations are Stein discrepancies}\label{sec:moments}

The denoising loss correlates a residual with fields of the noisy input.
These correlations query moments against a baseline. Conditioning on each
clean point shows how averaging fresh noisy views changes their variance.

\subsection{The Stein--Tweedie identity and polynomial queries}\label{sec:stein-id}

\begin{theorem}[Stein--Tweedie identity]\label{thm:stein}
Let \(\varphi\colon\R^d\to\R^d\) be \(C^1\) with
\(|\varphi(x)|+\|D\varphi(x)\|\le c(1+|x|^q)\), and let \(Q_0\) be a baseline with
\(|\nabla\log q_t(x)|\le c(1+|x|^q)\). Then
\begin{equation}\label{eq:stein}
 \E\inner{X_0-m^Q_t(X_t)}{\varphi(X_t)}=-\frac{h_t}{a_t}\,\E_{P_t}\bigl[\Stein_{Q_t}\varphi\bigr].
\end{equation}
\end{theorem}

The left side is what a gradient method computes: for a model \(m^Q_t+f_\theta\),
the population gradient of the denoising loss is
\(-2\E\inner{X_0-m^Q_t-f_\theta}{\partial_\theta f_\theta}\), and its
data-dependent part is a correlation of the residual with the fields
\(\varphi=\partial_\theta f_\theta\). The right side is a Stein discrepancy term
of the noisy law against \(Q_t\). The identity itself is standard. Its role
here is to specify which departures from a baseline the denoising loss can
reveal.

\begin{theorem}[Correlational denoising equals polynomial moment queries]\label{thm:bijection}
Let \(Q_t=N(\bar m,C)\) and \(D\ge1\). Write \(\mathcal P_D\) for polynomials of
degree at most \(D\) and \(\mathcal P_D^0\) for those with \(\E_{Q_t}q=0\).
\begin{enumerate}
\item For every polynomial field \(\varphi\) of degree at most \(D-1\),
\(\Stein_{Q_t}\varphi\in\mathcal P_D^0\).
\item The map \(u\mapsto\Stein_{Q_t}\nabla u\) is a linear bijection of
\(\mathcal P_D^0\) onto itself. Hence for every \(q\in\mathcal P_D\) there is a
polynomial \(u_q\) of degree at most \(D\) with
\begin{equation}\label{eq:sqcsq}
 \E_{P_t}q-\E_{Q_t}q=-\frac{a_t}{h_t}\,\E\inner{X_0-m^Q_t(X_t)}{\nabla u_q(X_t)}.
\end{equation}
\item \(\lambda_{\min}(C)\|q\|^2/D\le\|\nabla u_q\|^2_{L^2(Q_t)}\le\lambda_{\max}(C)\|q\|^2_{L^2(Q_t)}\)
for \(q\in\mathcal P_D^0\).
\end{enumerate}
Consequently, correlations of the denoising residual with polynomial fields of
degree at most \(D-1\) determine, and are determined by, the moment differences
between \(P_t\) and \(Q_t\) of order at most \(D\).
\end{theorem}

For supervised regression with Gaussian inputs, a correlational query sees only
the Hermite coefficients of the target, and correlational methods pay its
information exponent, which can exceed the generative exponent
\citep{damian2024gaps}. In denoising, the label \(X_0\) is tied to the input by a
known Gaussian channel, and Theorem~\ref{thm:bijection} shows that correlations
with the label realize every polynomial moment query of the input law. The
information exponent of the score as a regression target is therefore not a
barrier for denoising-based learning; whether a given architecture and optimizer
realize the required fields with controlled norm is the algorithmic question
that Sections~\ref{sec:frozen}--\ref{sec:training} answer.

\subsection{Conditioning on the clean point}\label{sec:condstein}

Conditioning the correlation on the clean point gives an exact clean-data
moment query.

\begin{lemma}[Conditional Stein identity]\label{lem:condstein}
Under Assumption~\ref{ass:cm}, let \(Q_t=N(0,\bar\sigma_t^2\Id)\) and \(\lambda=\lambda_t\).
\begin{enumerate}
\item If \(u(x)=\sum_{|\alpha|=j}c_\alpha H_\alpha(x/\bar\sigma_t)\) lies in the degree-\(j\)
Hermite chaos of \(Q_t\) and \(\tilde u(x)=\sum_{|\alpha|=j}c_\alpha H_\alpha(x/\sigma_0)\),
then for every \(x_0\in\R^d\)
\[
 \E\bigl[\inner{X_0-m^Q_t(X_t)}{\nabla u(X_t)}\bigm|X_0=x_0\bigr]
 =\frac{h_t\,j\,\lambda^{j}}{a_t\bar\sigma_t^2}\,\tilde u(x_0).
\]
\item For every unit \(w\) and \(j\ge1\),
\[
 \E\bigl[(X_0-m^Q_t(X_t))\,\He_{j-1}(w^\top X_t/\bar\sigma_t)\bigm|X_0=x_0\bigr]
 =\frac{h_t\,\sigma_0\lambda^{j-1}}{\bar\sigma_t^2}\,\He_j(x_0/\sigma_0)[w^{\otimes(j-1)},\cdot\,].
\]
\end{enumerate}
\end{lemma}

Taking expectations in part~1 recovers Theorem~\ref{thm:bijection} together
with the time action of Theorem~\ref{thm:noiseop} below. The content is the
conditioning: the noise-averaged denoising correlation is a deterministic
function of the clean point, namely a moment query on the \emph{clean} data
rescaled by \(\lambda^j\). A learner that revisits a clean point under many fresh
noise draws converges to that clean statistic, and the variance injected by the
forward process disappears; this is a Rao--Blackwellization performed by the
training protocol. Part~2 says that the noise-averaged gradient of the denoising
loss with respect to the output vector \(v\) of a unit
\(\He_{j-1}(w^\top x/\bar\sigma_t)\,v\), at \(v=0\), is one step of the tensor power method
on the clean Hermite tensor.

\subsection{The baseline principle}\label{sec:baseline}

\begin{theorem}[A baseline hides the moments it matches]\label{thm:baseline}
Let \(Q_0\) be Gaussian and suppose \(P_0\) and \(Q_0\) have equal moments of every
order below \(k\). Then for every \(t>0\):
\begin{enumerate}
\item \(\E\inner{X_0-m^Q_t(X_t)}{\varphi(X_t)}=0\) for every polynomial field
\(\varphi\) of degree at most \(k-2\);
\item if the moments of order \(k\) differ, some polynomial field of degree
\(k-1\) has nonzero correlation, and
\(\E_{P_t}X^{\otimes k}-\E_{Q_t}X^{\otimes k}=a_t^k(\E_{P_0}X^{\otimes k}-\E_{Q_0}X^{\otimes k})\).
\end{enumerate}
\end{theorem}

\begin{corollary}[Exact polynomial barrier]\label{cor:polybarrier}
Under Assumption~\ref{ass:cm} with the isotropic baseline, the best approximation
in \(L^2(P_t)\) of \(m_t-m^Q_t\) by polynomial fields of degree at most
\(k^\ast-2\) is zero. No polynomial dictionary of degree below \(k^\ast-1\), of any
size, improves the score on the Gaussian baseline, at any orientation, dimension
or time.
\end{corollary}

\begin{corollary}[Gaussian preconditioning shifts the exponent]\label{cor:shift}
Suppose \(P_0\) has an anisotropic covariance with range \(U\), as in
Assumption~\ref{ass:model}. Against an isotropic baseline the first moment
mismatch is at order two, and correlations with linear fields reveal \(U\).
Against the Gaussian fitted to the first two moments, correlations with fields
of degree at most one vanish; if the latent law is centrally symmetric, those
of degree two vanish as well; and if its fourth cumulant tensor is nonzero (for
the two-component mixture it is \(-2\mu^4\) per coordinate), the first nonzero
correlation has degree three and the exponent is four.
\end{corollary}

Corollary~\ref{cor:shift} is the mechanism behind the two regimes. In the
covariance-visible regime the Gaussian part of the score teaches the geometry:
correlations of degree one reveal the subspace at linear cost. A learner that
subtracts the fitted Gaussian, or a frozen representation that cannot use the
covariance, faces exponent four, and Section~\ref{sec:frozen} shows that it
pays a cubic price. It also explains the staged learning observed by
\citet{bardone2026easyhard}: once a model has fitted the Gaussian part of the
score, the correlations it can still exploit are of order at least three.

\subsection{Diffusion time}\label{sec:time}

\begin{theorem}[Exact action of the forward process]\label{thm:noiseop}
Under Assumption~\ref{ass:cm} with \(k=1\), let \(Y_t=u^\top X_t\) and
\(e_j(t)=\E\,h_j(Y_t/\bar\sigma_t)\). Then \(e_j(t)=\lambda_t^{\,j}e_j(0)\) for \(j\ge0\), so the
generative exponent is the same at every time. If moreover the likelihood ratio
\(r_0=d\,\mathrm{law}(Z/\sigma_0)/dN(0,1)\) lies in \(L^2(N(0,1))\), then
\(r_t=d\,\mathrm{law}(Y_t/\bar\sigma_t)/dN(0,1)=\sum_je_j(t)h_j\) in \(L^2\), the chi-square
divergence \(\chi^2_t=\sum_{j\ge1}e_j(t)^2\) is nonincreasing in \(t\),
\(dP_t/dQ_t(x)=r_t(u^\top x/\bar\sigma_t)\), and for unit directions \(u,v\)
\[
 \E_{Q_t}\bigl[(r_t(u^\top x/\bar\sigma_t)-1)(r_t(v^\top x/\bar\sigma_t)-1)\bigr]
 =\sum_{j\ge k^\ast}e_j(0)^2\bigl(\lambda_t^2\inner uv\bigr)^j .
\]
\end{theorem}

The two-component mixture satisfies the \(L^2\) hypothesis whenever
\(\tau<\sqrt2\,\sigma_0\), in particular whenever \(\mu^2+\tau^2=\sigma_0^2\); we
write \(\chi^2=\chi^2_0\). Pairwise correlations of this form control
statistical-query lower bounds for recovering \(u\) \citep{dks2017}. At time
\(t\) they are those of time zero with \(\inner uv\) multiplied by
\(\lambda_t^2<1\), so a learner that queries the noisy marginal is no stronger
than one that queries the data, and a learner that queries the denoising pair is
exactly as strong. No noise schedule lowers the exponent. Diffusion time
nevertheless matters for the \emph{constant}. Define, for a fixed unit \(u\),
\[
 S_t=\sup\Bigl\{\frac{|\E\inner{X_0-m^Q_t(X_t)}{\varphi(X_t)}|}{\mathrm{sd}_{Q}\inner{X_0-m^Q_t(X_t)}{\varphi(X_t)}}:\
 \varphi\ \text{polynomial of degree}\le k^\ast-1\Bigr\},
\]
the per-sample signal-to-noise ratio of the best low-degree denoising
correlation against the Gaussian null.

\begin{theorem}[Optimal geometry time]\label{thm:time}
Under Assumption~\ref{ass:cm} with \(k=1\) and \(k^\ast\ge3\), let \(\kappa\) be the
\(k^\ast\)-th cumulant of \(Z\). Then
\[
 S_t=\sqrt{\frac{k^\ast}{k^\ast!}}\;\frac{|\kappa|}{\sigma_0}\;
 \frac{\sqrt{h_t}\,a_t^{\,k^\ast-1}}{\bar\sigma_t^{\,k^\ast}},
\]
which has a unique maximum at
\(t^\ast=\frac12\log\{(k^\ast+\sigma_0^2-1)/(k^\ast-1)\}\). For \(\sigma_0=1\):
\(t^\ast=0.203\) for \(k^\ast=3\) and \(0.144\) for \(k^\ast=4\).
\end{theorem}

The optimal time depends on the data only through \(k^\ast\) and \(\sigma_0\).
It optimizes test fields of degree at most \(k^\ast-1\); higher-degree fields
can shift the optimum of the full objective. It also minimizes the leading
constant of the single-view spectral learner
(Theorem~\ref{thm:multiview}(5)). In the fixed-step SGD experiment of
Section~\ref{sec:exp-failures}, the escape-time ratio did not follow
\(S_t^{-2}\).

\begin{figure}[t]
\centering
\includegraphics[width=\linewidth]{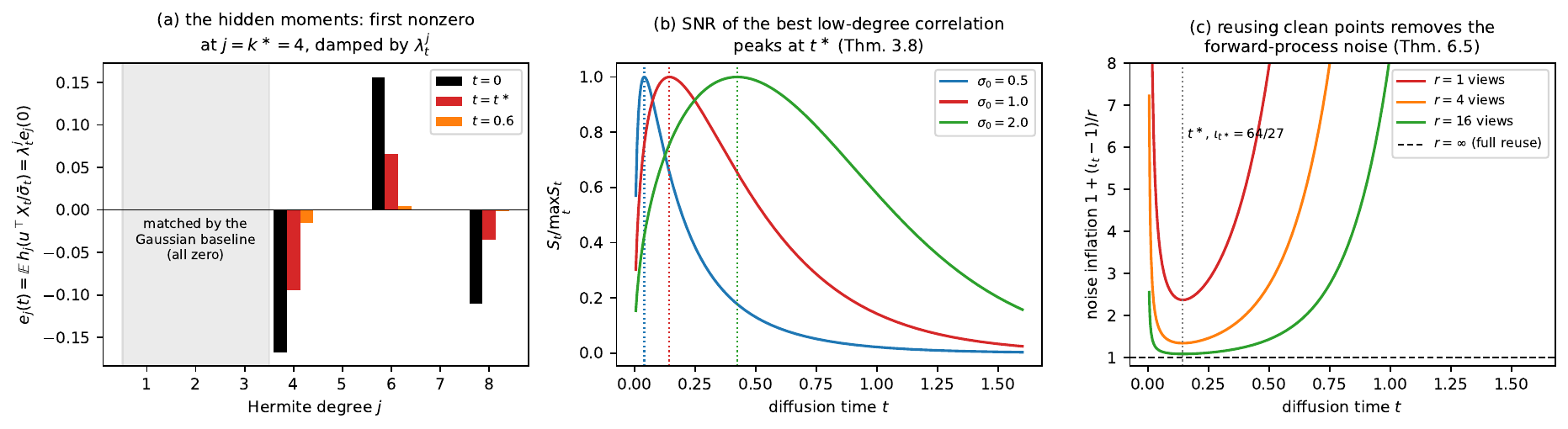}
\caption{Moment access in the covariance-matched regime (\(\mu=0.8\),
\(\sigma_0=1\)). (a)~Hermite moments \(e_j(t)\) of the hidden coordinate. The
Gaussian baseline matches degrees one to three, so they carry no signal
(Theorem~\ref{thm:baseline}); diffusion multiplies degree \(j\) by
\(\lambda_t^j\) (Theorem~\ref{thm:noiseop}), so it never creates a lower-degree
signal. (b)~The signal-to-noise ratio \(S_t\) of Theorem~\ref{thm:time} peaks at
\(t^\ast\), which depends only on \(k^\ast\) and \(\sigma_0\) (dotted lines).
(c)~The noise inflation factor \(1+(\iota_t-1)/r\) of
Theorem~\ref{thm:multiview}: one view per clean point pays at least \(64/27\) at
\(t^\ast\), and reusing each clean point under many fresh views removes the factor at
every time.}
\label{fig:moments}
\end{figure}
\section{Frozen representations pay \texorpdfstring{\(d^{k^\ast-1}\)}{d^(k*-1)}}\label{sec:frozen}

A frozen representation is a dictionary chosen independently of the hidden
orientation (Definition~\ref{def:dictionary}); its fitted coefficients may
depend on the data and on \(U\). In the covariance-visible regime, the
barrier starts at \(d^3/6\), with an exact risk height for one hidden direction
(Sections~\ref{sec:reductions}--\ref{sec:height}). Projection onto the
principal subspace removes ambient dimension from the fitted pipeline
(Section~\ref{sec:adaptive}). In the covariance-matched regime, the data law
has no positive density floor relative to the matched Gaussian.
Hypercontractivity instead gives the exponent \(d^{k^\ast-1}\)
(Section~\ref{sec:frozen-cm}). Throughout
Sections~\ref{sec:reductions}--\ref{sec:separation} the data follow
Assumption~\ref{ass:model}; in Proposition~\ref{prop:offsubspace} and the
discussion after Theorem~\ref{thm:generation} only, \(\sigma_0\) denotes the
scale of isotropic noise \(\sigma_0G_0\) added to all coordinates of
\(X_0=UZ\).

\begin{figure}[t]
\centering
\includegraphics[width=\linewidth]{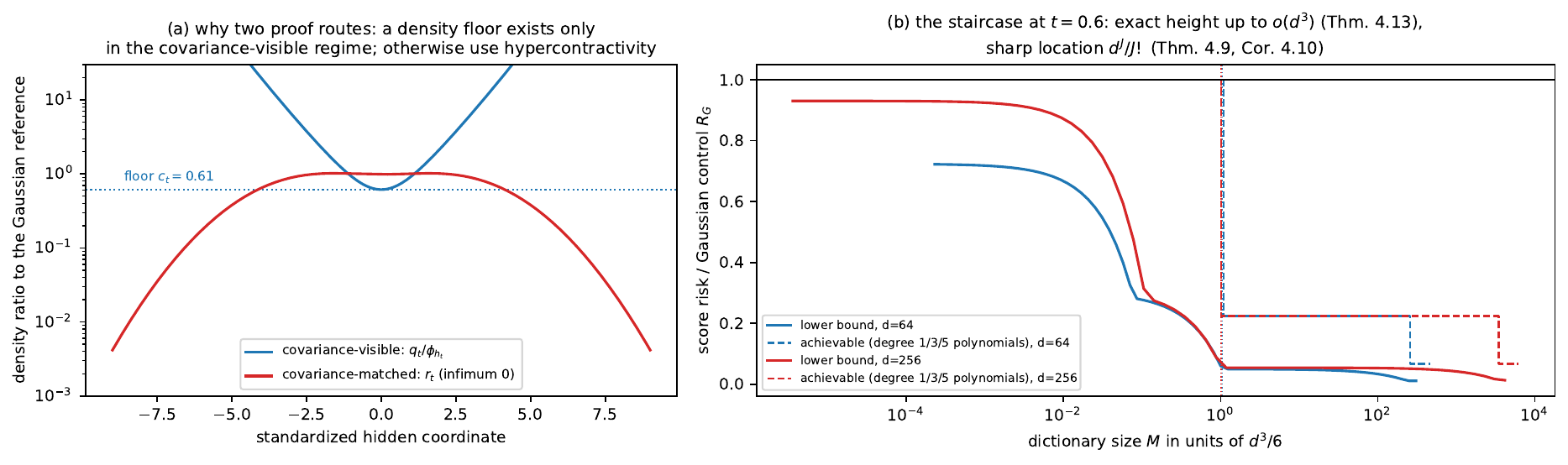}
\caption{Frozen representations. (a)~Density ratio of the hidden coordinate to its
Gaussian reference at \(t=0.6\). In the covariance-visible regime it is bounded
below by \(c_t\) (Lemma~\ref{lem:density}), which gives the barrier of
Theorem~\ref{thm:floor}; in the covariance-matched regime its infimum is zero, and
the one-sided change of measure (Lemma~\ref{lem:onesided}) combined with
hypercontractivity (Lemma~\ref{lem:comparison}) replaces the floor. (b)~The
staircase in the covariance-visible regime at \(t=0.6\), computed from the exact
constants. Solid: the best lower bound on the score risk of any orientation-blind
dictionary of size \(M\) (maximum of Theorems~\ref{thm:floor} and~\ref{thm:height}
and the level-5 floor). Dashed: the risk attained by polynomial dictionaries of
degree 1, 3 and 5 (Theorem~\ref{thm:sharp}). Dotted: \(N_{d,3}\). Below \(d^3/6\) the
lower bound approaches the Gaussian control as \(d\) grows (exact height); the
threshold sits at \(d^3/6\) (sharp location); between consecutive thresholds the
height is known only within the gap shown.}
\label{fig:frozen}
\end{figure}

A representation fixed before the orientation is revealed meets the hidden
direction through overlaps of order \(d^{-1/2}\). Its degree-\(J\) part
therefore needs about \(d^J/J!\) elements to align with the hidden signal.
The proofs use a spectral count on the sphere and a comparison with an
orientation-free Gaussian; the latter differs between the two regimes
(Figure~\ref{fig:frozen}).

\subsection{Two reductions}\label{sec:reductions}

\begin{lemma}[Denoiser reduction]\label{lem:reduction}
Under Assumption~\ref{ass:model}, let \(g_t(y)=\E[Z\mid a_tZ+\sqrt{h_t}\,\xi=y]\)
for \(\xi\sim N(0,\Id_k)\) independent of \(Z\). Then
\begin{enumerate}
\item \(m_t(x)=Ug_t(U^\top x)\) and \(s_t(x)=(a_tm_t(x)-x)/h_t\);
\item the normal component of the score is exactly linear:
\((\Id-UU^\top)s_t(x)=-(\Id-UU^\top)x/h_t\);
\item for any \(\widehat m\) and \(\widehat s\) related by~\eqref{eq:tweedie},
\(\risk_t(\widehat s;U)=(a_t/h_t)^2\,\E\norm{\widehat m(X_t,t)-m_t(X_t)}^2\).
\end{enumerate}
If \(\widehat m\) takes values in \(\spanop(U)\), its score error is purely
tangential.
\end{lemma}

\begin{lemma}[Diffused mixture coordinate]\label{lem:posterior}
For one coordinate \(Z=\epsilon\mu+\tau\xi\), let \(s^2=a_t^2\tau^2+h_t\). Then
\[
 g_t(y)=\frac{\mu h_t}{s^2}\tanh\Bigl(\frac{a_t\mu y}{s^2}\Bigr)
 +\frac{a_t\tau^2}{s^2}\,y ,
\]
and \(g_t\) is odd. Under Assumption~\ref{ass:model}, \(g_t\) acts
coordinatewise on \(\R^k\).
\end{lemma}

Let \(h_j\) be the orthonormal probabilists' Hermite polynomials and
\[
 c_j(t)=\E\bigl[g_t(\sqrt{h_t}\,\xi)h_j(\xi)\bigr],\qquad
 \kappa_t=\sum_{j\geq3}c_j(t)^2
 =\E[g_t(\sqrt{h_t}\xi)^2]-c_1(t)^2 ,
\]
for \(\xi\sim N(0,1)\). Since \(g_t\) is odd, \(c_j=0\) for even \(j\). The
quantity \(\kappa_t\) is the energy of \(g_t\) beyond its best linear
approximation in \(L^2(N(0,h_t))\). More generally, for odd \(J\geq3\) we write
\(\kappa_{J,t}=\sum_{j\ge J}c_j(t)^2\), so that \(\kappa_t=\kappa_{3,t}\). Every
\(\kappa_{J,t}\) is positive, because \(g_t\) is the sum of a linear function and
a bounded nonconstant function and is therefore not a polynomial. We also use the
derivative energy
\[
 J_t=\sum_{j\ge3}j\,c_j(t)^2=h_t\,\E\bigl[g_t'(\sqrt{h_t}\xi)^2\bigr]-c_1(t)^2 ,
\]
which is finite because \(g_t'\) is bounded; \(J_t/\kappa_t\geq3\) is the mean
Hermite degree of the nonlinear energy.

\subsection{A cubic barrier for orientation-blind dictionaries}\label{sec:floor}

\subsubsection{Density comparison}
The input law \(P_t^U\) depends on the hidden orientation. We compare it with
the orientation-free reference \(\gamma_t=N(0,h_t\Id_d)\).

\begin{lemma}[Density floor]\label{lem:density}
Let \(q_t\) be the density of \(a_tZ_1+\sqrt{h_t}\xi\) and \(\phi_{h}\) that of
\(N(0,h)\). Then
\(dP_t^U/d\gamma_t(x)=\prod_{j=1}^k\rho_t(u_j^\top x)\) with
\(\rho_t=q_t/\phi_{h_t}\), and
\[
 \rho_t(y)\geq c_t:=\E\exp\Bigl(-\frac{a_t^2Z_1^2}{2h_t}\Bigr)
 =\sqrt{\frac{h_t}{s^2}}\exp\Bigl(-\frac{a_t^2\mu^2}{2s^2}\Bigr),
\]
with equality at \(y=0\).
\end{lemma}

The constant does not depend on \(d\) or \(U\). It tends to one as \(t\to\infty\)
and to zero as \(t\to0\). The comparison is where the bound loses most. In
Table~\ref{tab:constants}, the limiting floor is \(3\%\) of the population
Gaussian control's risk at \(t=0.1\), \(15\%\) at \(t=0.3\) and \(50\%\) at
\(t=1\). Among isotropic Gaussian references \(N(0,\sigma^2\Id_d)\), only
\(\sigma^2=h_t\) gives a ratio bounded below uniformly in \(d\), because the
\(d-k\) nuisance coordinates have variance exactly \(h_t\). For \(k=1\),
Section~\ref{sec:height} removes this loss from the main term.

\subsubsection{The spectral floor}
Let \(N_{d,\ell}\) denote the dimension of degree-\(\ell\) spherical harmonics on
\(\Sph^{d-1}\); \(N_{d,1}=d\), \(N_{d,3}=d(d-1)(d+4)/6\), and in general
\(N_{d,\ell}=\binom{d+\ell-1}{\ell}-\binom{d+\ell-3}{\ell-2}=(1+O(1/d))\,d^\ell/\ell!\).

\begin{lemma}[Orientation-averaged approximation]\label{lem:spectral}
Let \(g\in L^2(N(0,h))\) be odd with normalized Hermite coefficients \(c_j\), so
that \(g(y)=\sum_jc_jh_j(y/\sqrt h)\). For \(u\in\Sph^{d-1}\), let
\(f_u(x)=g(u^\top x)\in L^2(\gamma)\), \(\gamma=N(0,h\Id_d)\). Let
\(\mathcal S\subset L^2(\gamma)\) be a random subspace of dimension at most
\(M\), independent of \(u\sim\mathrm{Unif}(\Sph^{d-1})\). Then, for every odd
\(J\geq3\),
\[
 \E\norm{f_u-\Pi_{\mathcal S}f_u}_{L^2(\gamma)}^2
 \;\geq\;\tr T-\sum_{i=1}^M\nu_i(T)
 \;\geq\;\Bigl(1-\frac{M}{N_{d,J}}\Bigr)_+\sum_{j\geq J}c_j^2\,w_{j,J}(d),
\]
where \(T=\E_u[f_u\otimes f_u]\) has eigenvalues
\(\nu_1\geq\nu_2\geq\cdots\), and the weights
\(w_{j,J}(d)=\sum_{\ell\ge J}N_{d,\ell}\,\eta_{j,\ell}\in[0,1]\) satisfy
\begin{equation}\label{eq:weights}
 \begin{aligned}
 w_{j,J}(d)&\;\geq\;\prod_{i=0}^{j-2}\frac{d-1+i}{d+2i}\;\geq\;1-\frac{j(j-1)}{2d}
 &&(j\geq J),\\
 w_{j,3}(d)&\;=\;1-\prod_{q=1}^{(j-1)/2}\frac{2q+1}{d+2q}\;\geq\;1-\frac3{d+2}
 &&(j\geq3\text{ odd}).
 \end{aligned}
\end{equation}
In particular the right side is at least
\((1-3/(d+2))(1-M/N_{d,3})_+\sum_{j\geq3}c_j^2\) for \(J=3\). The positive
eigenvalues of \(T\) are \(\Lambda_\ell=\sum_jc_j^2\eta_{j,\ell}\) on
degree-\(\ell\) harmonics, with multiplicity \(N_{d,\ell}\), where
\begin{equation}\label{eq:eta}
 \eta_{j,\ell}=\frac{j!}{r!\,2^r\prod_{q=0}^{(j+\ell)/2-1}(d+2q)},
 \qquad r=\frac{j-\ell}{2}\in\{0,1,\dots\},
\end{equation}
and \(\eta_{j,\ell}=0\) when \(j-\ell\) is odd or negative.
\end{lemma}

The first inequality is the exact (Ky Fan) optimum over \(M\)-dimensional
spans under the Gaussian reference; we call it the \emph{Ky Fan floor} and
evaluate it numerically. It is optimal for approximation by an unconstrained
span in \(L^2(\gamma)\), not for the score-risk problem itself, which also
incurs the density factor of Lemma~\ref{lem:density}. The second inequality is
explicit. Its \(J=3\) case uses only the degree-one eigenvalue and the
monotonicity of \(N_{d,\ell}\) in \(\ell\).

\subsubsection{The barrier}

Definition~\ref{def:dictionary} fixes the class of dictionaries.

\begin{theorem}[Cubic barrier, orientation average]\label{thm:floor}
Under Assumption~\ref{ass:model}, let \(U\) be Haar-distributed on the Stiefel
manifold and independent of the latent data. Let \(\widehat m\) be supported by
an orientation-independent dictionary of dimension at most \(M\), and let
\(\widehat s\) be given by~\eqref{eq:tweedie}. Then, for every \(t>0\),
\begin{equation}\label{eq:floor}
 \bar\risk_t(\widehat s)\;\geq\;
 F_d(M;t):=k\Bigl(\frac{a_t}{h_t}\Bigr)^2c_t^k
 \Bigl(1-\frac3{d+2}\Bigr)\Bigl(1-\frac{M+d+1}{N_{d,3}}\Bigr)_+\kappa_t .
\end{equation}
More generally, for every odd \(J\geq3\),
\(\bar\risk_t(\widehat s)\geq F^{(J)}_d(M;t):=k(a_t/h_t)^2c_t^k
(1-(M+d+1)/N_{d,J})_+\sum_{j\ge J}c_j(t)^2w_{j,J}(d)\), and
\(\bar\risk_t(\widehat s)\geq
F^{\mathrm{KF}}_d(M;t)=k(a_t/h_t)^2c_t^k[\tr T_t-\sum_{i\leq M+d+1}\nu_i(T_t)]\),
where \(T_t\) is the operator of Lemma~\ref{lem:spectral} for \(g=g_t\).
\end{theorem}

The bound concerns a single time \(t\), and it applies to integrated risks by
integration. The factor \(1-3/(d+2)\) is the part of the cubic energy that a
degree-one harmonic cannot capture. The factor involving \(M+d+1\) counts the
dictionary and the affine functions together. We refer to \(F_d\) as the
\emph{explicit floor} and to \(F^{\mathrm{KF}}_d\geq F^{(J)}_d\) as the Ky Fan
floor; \(F^{(3)}_d\geq F_d\) by~\eqref{eq:weights}.

\begin{corollary}[Geometry-blind random features]\label{cor:ambient}
The ambient random-feature estimator of Section~\ref{sec:model}, with any
finite ridge grid and \(L\) bandwidth multipliers, is supported by an
orientation-independent dictionary of dimension at most \(L(p+1)\). This holds
for any data-dependent choice of ridge and of bandwidth within the fixed grid
\(\mathcal B\), and for any coefficients. A bandwidth chosen outside a fixed
grid by an orientation-sensitive rule is not covered. Hence
\(\bar\risk_t\geq F_d(L(p+1);t)\).
\end{corollary}

The dictionary is \(\spanop\bigcup_{\beta\in\mathcal B}\{\text{coordinates of }
\varphi_\beta\}\). It depends on the training data only through
\(\tr\widehat\Sigma\), which equals the trace of the sample covariance of
\(Z^{(i)}\) and is therefore a function of the latent data.

\subsubsection{Every orientation and typical orientations}

Theorem~\ref{thm:floor} averages over orientations, and an arbitrary
orientation-independent dictionary need not have the same risk at every
orientation. Two results remove the average.

\begin{proposition}[Every orientation for isotropic random features]\label{prop:fixed}
For the ambient random-feature estimator with independent \(N(0,1)\) input
weights, the law of its score risk is the same for every fixed \(U\).
Consequently \(\risk_t(\widehat s;U)\geq F_d(L(p+1);t)\) for every orientation
\(U\), not only on average.
\end{proposition}

\begin{theorem}[Typical orientations]\label{thm:typical}
Under Assumption~\ref{ass:model}, let \(\widehat m\) be supported by an
orientation-independent dictionary of dimension at most \(M\), with \(\widehat
s\) given by~\eqref{eq:tweedie}. There is a function
\(\Phi_t:\Sph^{d-1}\to[0,\kappa_t]\), determined by the law of the dictionary,
such that for every orientation \(U\)
\[
 \risk_t(\widehat s;U)\;\geq\;\Bigl(\frac{a_t}{h_t}\Bigr)^2c_t^k
 \sum_{i=1}^k\Phi_t(u_i),
\]
\(\Phi_t\) is \(2\sqrt{\kappa_tJ_t}\)-Lipschitz, and its average over the
sphere is at least \((1-\frac3{d+2})(1-\frac{M+d+1}{N_{d,3}})_+\kappa_t\).
Consequently, for every \(s>0\), the set of orientations at which
\[
 \risk_t(\widehat s;U)\;<\;F_d(M;t)-k\Bigl(\frac{a_t}{h_t}\Bigr)^2c_t^k\,s
\]
has Haar measure at most \(k\exp\{-(d-2)s^2/(8\kappa_tJ_t)\}\).
\end{theorem}

With \(s=\eta\kappa_t\), the exceptional mass is at most
\(k\exp\{-(d-2)\eta^2/(8\bar\jmath_t)\}\), where
\(\bar\jmath_t=J_t/\kappa_t\) lies between \(3.1\) and \(4.9\) at the times we
use. The constant is conservative, but for fixed \(\eta\) and \(M=o(d^3)\) the
fraction of orientations at which an orientation-blind dictionary beats
\((1-\eta)F_\infty(t)\) tends to zero exponentially fast in \(d\).

\subsubsection{Sharpness: the polynomial staircase}

The barrier vanishes once \(M+d+1\geq N_{d,3}\approx d^3/6\). A deterministic
orientation-blind dictionary of that size resolves the cubic part of the
score at every orientation, fixing the threshold constant at \(1/6\).

\begin{theorem}[Matching polynomial dictionaries]\label{thm:sharp}
Fix \(t>0\) and an odd \(J\geq1\). Let \(\mathcal P_{d,J}\) be the space of
polynomials of degree at most \(J\) on \(\R^d\). It is deterministic, hence an
orientation-independent dictionary, it contains \(\Aff\), and its dimension is
\(D_{d,J}=\binom{d+J}{J}=N_{d,J}+O(d^{J-1})\). Let
\[
 \varepsilon_{J}(t)=\min_{\pi}\E\bigl[(g_t(Y)-\pi(Y))^2\bigr],\qquad
 Y\sim q_t,
\]
over univariate polynomials \(\pi\) of degree at most \(J\). Then for
\emph{every} orientation \(U\) there is an estimator \(\widehat m\) supported by
\(\mathcal P_{d,J}\) with
\[
 \risk_t(\widehat s;U)=k\Bigl(\frac{a_t}{h_t}\Bigr)^2\varepsilon_J(t),
\]
which does not depend on \(d\). Moreover \(\varepsilon_J(t)\) is nonincreasing
in \(J\), \(\varepsilon_J(t)\to0\) as \(J\to\infty\), and \(\varepsilon_1(t)\)
is the population Gaussian control's risk divided by \(k(a_t/h_t)^2\).
\end{theorem}

The estimator is \(\widehat m(x)=U\pi_J(U^\top x)\), with \(\pi_J\) the minimizer
applied coordinatewise. Its coefficients use the orientation, as the class of
Theorem~\ref{thm:floor} allows; its \emph{span} does not. The theorem therefore
shows that the barrier is representational: no lower bound for the class can
move the threshold, and a data-driven version needs only the exactly recovered
\(\widehat U\) and a one-dimensional polynomial regression.

\begin{corollary}[Polynomial staircase]\label{cor:staircase}
Fix \(t>0\), an odd \(J\geq3\) and \(\delta\in(0,1)\).
\begin{enumerate}
\item If \(M+d+1\leq(1-\delta)N_{d,J}\), every estimator supported by an
orientation-independent dictionary of dimension at most \(M\) has
\[
 \bar\risk_t(\widehat s)\;\geq\;\delta\,k\Bigl(\frac{a_t}{h_t}\Bigr)^2c_t^k
 \sum_{j\ge J}c_j(t)^2\prod_{i=0}^{j-2}\frac{d-1+i}{d+2i}
 \;\xrightarrow[d\to\infty]{}\;\delta\,k\Bigl(\frac{a_t}{h_t}\Bigr)^2c_t^k\,\kappa_{J,t}>0 .
\]
\item If \(M\geq D_{d,J}\), some estimator supported by an orientation-independent
dictionary of dimension \(M\) has \(\risk_t(\widehat s;U)\leq
k(a_t/h_t)^2\varepsilon_J(t)\) at every \(U\).
\end{enumerate}
Since \(N_{d,J}\) and \(D_{d,J}\) both equal \((1+O(1/d))\,d^J/J!\), the
dictionary size at which orientation-blind dictionaries resolve the degree-\(J\)
part of the score is \(d^J/J!\) to first order. For every \(J\) and every
\(\varepsilon<k(a_t/h_t)^2c_t^k\kappa_{J,t}\), an orientation-averaged risk at
most \(\varepsilon\) needs \(\Omega(d^J)\) dictionary elements, so no polynomial
budget in \(d\) suffices for all accuracies.
\end{corollary}

The lower level \(c_t^k\kappa_{J,t}\) and the upper level
\(\varepsilon_{J-2}(t)\) between consecutive thresholds are both
dimension-free, but they are not equal; Table~\ref{tab:constants} lists them.
For \(k=1\) and \(J=3\), Theorem~\ref{thm:height} below closes this gap: the
lower level of the first step is the upper level \(\varepsilon_1(t)\).
When \(F_\infty(t)\) exceeds the cubic level \(k(a_t/h_t)^2\varepsilon_3(t)\), the
two parts of the corollary force a drop. For the parameters of
Table~\ref{tab:constants} (\(\mu=1.5\), \(\tau=0.35\), \(k=1\)), the ratio is
\(5.4\) at \(t=1\), so for
large \(d\) every orientation-blind dictionary with at most \(0.8\,d^3/6\)
elements has orientation-averaged risk above the risk that the cubic dictionary
of size \((1+o(1))\,d^3/6\) attains at every orientation; at \(t=0.6\) the ratio
is \(1.4\) and the same holds below \(0.25\,d^3/6\). Section~\ref{sec:experiments}
traces the whole curve for random cubic and quintic sub-dictionaries.

\begin{corollary}[Cubic feature price]\label{cor:price}
Let \(B_d(t)=k(a_t/h_t)^2c_t^k(1-3/(d+2))\kappa_t\) and \(\varepsilon\geq0\).
\begin{enumerate}
\item \emph{Orientation average.} Any estimator supported by an
orientation-independent dictionary of dimension \(M\) with
\(\bar\risk_t\leq\varepsilon\) has
\(M\geq N_{d,3}(1-\varepsilon/B_d(t))-d-1\).
\item \emph{Every orientation.} For the ambient random-feature estimator, the
same bound with \(M=L(p+1)\) holds if \(\risk_t(\widehat s;U)\leq\varepsilon\)
at a single orientation.
\item \emph{Typical orientations.} If an estimator supported by an
orientation-independent dictionary has \(\risk_t(\widehat s;U)\le\varepsilon\)
on a set of orientations of Haar measure larger than
\(k\exp\{-(d-2)s^2/(8\kappa_tJ_t)\}\), then
\(M\geq N_{d,3}\bigl(1-(\varepsilon+k(a_t/h_t)^2c_t^ks)/B_d(t)\bigr)-d-1\).
\end{enumerate}
If the bracket is at least a fixed \(\delta>0\) (for example, a fixed
\(\varepsilon<F_\infty(t)\), a small \(s\) and all large \(d\)), the price is
\(\delta N_{d,3}-d-1=\Omega(d^3)\); by Theorem~\ref{thm:sharp} it is also
\(O(d^3)\) for every \(\varepsilon\geq k(a_t/h_t)^2\varepsilon_3(t)\).
\end{corollary}

\subsubsection{Exact height of the first step for one hidden direction}\label{sec:height}

The density comparison of Lemma~\ref{lem:density} is applied to the whole
squared error in Theorem~\ref{thm:floor}, and this costs the factor \(c_t^k\).
For one hidden direction the factor can be moved off the main term. The device
is to apply the change of measure only to the cross term between the target and
the dictionary.

\begin{lemma}[One-sided change of measure]\label{lem:onesided}
Let \(P\ll Q\) be probability measures on \(\R^d\) with \(\rho=dP/dQ\), let
\(\mathcal V\) be a finite-dimensional space of vector fields in
\(L^2(P)\cap L^2(Q)\) with \(\norm f^2_{L^2(P)}\geq c\norm f^2_{L^2(Q)}\) for all
\(f\in\mathcal V\) and some \(c>0\), and let \(r\in L^2(P)\) with
\(\rho r\in L^2(Q)\). Then
\[
 \inf_{f\in\mathcal V}\norm{r-f}^2_{L^2(P)}\;\geq\;\norm r^2_{L^2(P)}
 -\frac1c\,\bigl\|\Pi^Q_{\mathcal V}(\rho r)\bigr\|^2_{L^2(Q)},
\]
where \(\Pi^Q_{\mathcal V}\) is the orthogonal projection onto \(\mathcal V\) in
\(L^2(Q)\).
\end{lemma}

For \(k=1\) write \(U=u\), let \(q_t\) and \(\rho_t=q_t/\phi_{h_t}\) be as in
Lemma~\ref{lem:density}, and let
\[
 \alpha_t=\frac{a_t(\mu^2+\tau^2)}{a_t^2(\mu^2+\tau^2)+h_t},\qquad
 \hat g_t(y)=g_t(y)-\alpha_ty,\qquad
 \psi_t=\rho_t\,\hat g_t,\qquad
 \Psi_t=\E\bigl[\psi_t(\sqrt{h_t}\,\xi)^2\bigr],
\]
with \(\xi\sim N(0,1)\). Here \(\alpha_ty\) is the best linear approximation of
\(g_t\) in \(L^2(q_t)\), so \(\E_{Y\sim q_t}\hat g_t(Y)^2=\varepsilon_1(t)\) and
\(R_G(t)=(a_t/h_t)^2\varepsilon_1(t)\) is the score risk of the population
Gaussian control (Theorem~\ref{thm:sharp} with \(J=1\)).

\begin{theorem}[Exact height of the cubic barrier]\label{thm:height}
Under Assumption~\ref{ass:model} with \(k=1\), let \(u\) be uniform on
\(\Sph^{d-1}\) and independent of the latent data, and assume
\(a_t^2\tau^2<h_t\). Then \(\Psi_t<\infty\), and every estimator supported by an
orientation-independent dictionary of dimension at most \(M\) satisfies, with
\(D=M+d+1\),
\[
 \bar\risk_t(\widehat s)\;\geq\;\Bigl(\frac{a_t}{h_t}\Bigr)^2
 \Bigl[\varepsilon_1(t)-\frac{\Psi_t}{c_t}\Bigl(\frac3{d+2}+\frac D{N_{d,3}}\Bigr)\Bigr]
\]
and \(\bar\risk_t(\widehat s)\geq(a_t/h_t)^2[\varepsilon_1(t)-c_t^{-1}
\sum_{i\leq D}\nu_i(T^\psi_t)]\), where \(T^\psi_t\) is the operator of
Lemma~\ref{lem:spectral} built from \(\psi_t\) in place of \(g_t\). In
particular, if \(M=o(d^3)\) then \(\bar\risk_t(\widehat s)\geq(1-o(1))\,R_G(t)\).
\end{theorem}

Together with Theorem~\ref{thm:sharp} at \(J=1\) and \(J=3\), the theorem makes
the first step of the staircase exact in both location and height. The affine
dictionary attains \(R_G(t)\) at every orientation; no orientation-blind
dictionary with \(o(d^3)\) elements does asymptotically better on average; and
the cubic polynomials, \((1+o(1))\,d^3/6\) elements, attain
\((a_t/h_t)^2\varepsilon_3(t)\). With \(o(d^3)\) elements, the asymptotic
gain over the moment-matched Gaussian vanishes for orientation-blind dictionaries.
Since \(\rho_t\geq c_t\), we have \(\varepsilon_1(t)\geq c_t\norm{\hat
g_t}^2_{L^2(N(0,h_t))}\geq c_t\kappa_t\), so the limit of
Theorem~\ref{thm:height} is never below the limit \(F_\infty(t)\) of
Theorem~\ref{thm:floor}. At finite \(d\), the correction carries \(1/c_t\) and
\(\Psi_t\geq c_t\varepsilon_1(t)\). Both bounds are valid, and the larger one
applies.

The condition \(a_t^2\tau^2<h_t\), equivalently \(t>\frac12\log(1+\tau^2)\), says
that each mixture component of \(u^\top X_t\), of variance \(a_t^2\tau^2+h_t\), is
narrower than \(2h_t\). It holds for \(t>0.058\) in our parameters. By
Proposition~\ref{prop:offsubspace}, the theorem also holds for data off the
subspace, with \(\lambda_t^{-2}\) and the shifted time \(t'\), whenever
\(a_{t'}^2\tau^2<h_{t'}\). The exact-height theorem treats one hidden
direction and the first step. For \(k\geq2\), the weighted residual
\(\rho r\) is a multi-index function of \(U^\top x\), for which
Lemma~\ref{lem:spectral} has no direct analogue.

Table~\ref{tab:height} evaluates the bound. For \(L=3\) bandwidths and width
\(p=1024\), it retains \(0.89\)--\(0.99\) of \(R_G\) at \(d=1024\) at the tabulated
times \(t\in\{0.3,0.6,1\}\), against \(0.15\)--\(0.51\) for Theorem~\ref{thm:floor}.
At \(t=0.1\), where \(c_t=0.03\), it is vacuous at these sizes. At \(d=64\) it
exceeds the cubic floor at the tabulated times \(t\in\{0.6,1\}\) but not at
\(t\le0.3\).

\begin{table}[t]
\centering
\caption{Lower bounds on the orientation-averaged score risk as fractions of the
Gaussian control \(R_G(t)\), for \(\mu=1.5\), \(\tau=0.35\), \(k=1\), \(L=3\)
bandwidths and width \(p=1024\), so \(D=3075+d+1\). \(F_\infty/R_G\): the
large-\(d\) limit of Theorem~\ref{thm:floor}. \emph{Cubic}: the Ky Fan floor of
Theorem~\ref{thm:floor}. \emph{Expl.}\ and \emph{KF}: the explicit and Ky Fan
forms of Theorem~\ref{thm:height}, whose large-\(d\) limit is \(1\). Values are
computed from the bounds stated in the theorems.}
\label{tab:height}

\begin{tabular}{rrrrrrrrrrrr}
\toprule
& & & \multicolumn{3}{c}{$d=64$} & \multicolumn{3}{c}{$d=256$} & \multicolumn{3}{c}{$d=1024$} \\
\cmidrule(lr){4-6}\cmidrule(lr){7-9}\cmidrule(lr){10-12}
$t$ & $c_t$ & $F_\infty/R_G$ & cubic & expl. & KF & cubic & expl. & KF & cubic & expl. & KF \\
\midrule
0.1 & 0.030 & 0.03 & 0.03 & 0.00 & 0.00 & 0.03 & 0.00 & 0.00 & 0.03 & 0.00 & 0.00 \\
0.3 & 0.284 & 0.15 & 0.14 & 0.00 & 0.00 & 0.15 & 0.00 & 0.51 & 0.15 & 0.23 & 0.89 \\
0.6 & 0.615 & 0.31 & 0.28 & 0.09 & 0.33 & 0.31 & 0.90 & 0.92 & 0.31 & 0.98 & 0.98 \\
1.0 & 0.833 & 0.51 & 0.46 & 0.72 & 0.74 & 0.50 & 0.97 & 0.97 & 0.51 & 0.99 & 0.99 \\
\bottomrule
\end{tabular}

\end{table}

\subsubsection{Data off the subspace}

The exact low-dimensional support is not what drives the barrier. Suppose the
data are only near the subspace.

\begin{proposition}[Data off the subspace]\label{prop:offsubspace}
Replace \(X_0=UZ\) by \(X_0^{\sigma}=UZ+\sigma_0G_0\), with \(\sigma_0\geq0\)
and \(G_0\sim N(0,\Id_d)\) independent of everything else, and let
\(s^\sigma_t\) be the score of the diffused law. Set
\(\lambda_t=(1+a_t^2\sigma_0^2)^{1/2}\) and \(t'=t+\log\lambda_t\). Then
\(a_tX_0^\sigma+\sqrt{h_t}G\) has the law of \(\lambda_tX_{t'}\), where
\(X_{t'}\) follows the noiseless model of Assumption~\ref{ass:model}, and
\(s^\sigma_t(x)=\lambda_t^{-1}s_{t'}(x/\lambda_t)\). For every estimator
supported by an orientation-independent dictionary of dimension \(M\), and every
orientation, the risk for \(s^\sigma_t\) equals \(\lambda_t^{-2}\) times the
noiseless risk at time \(t'\) of an estimator supported by an
orientation-independent dictionary of the same dimension. Consequently
Theorem~\ref{thm:floor} holds with \(F_d(M;t)\) replaced by
\(\lambda_t^{-2}F_d(M;t')\) (and likewise for \(F^{(J)}_d\) and
\(F^{\mathrm{KF}}_d\)); Theorem~\ref{thm:typical} holds with
\(\lambda_t^{-2}[F_d(M;t')-k(a_{t'}/h_{t'})^2c_{t'}^ks]\) and exceptional mass
\(k\exp\{-(d-2)s^2/(8\kappa_{t'}J_{t'})\}\); and Proposition~\ref{prop:fixed}
holds verbatim for the ambient random-feature estimator trained on
\(X_0^\sigma\).
\end{proposition}

Since \(t'>t\), the noisy problem inherits the larger density constant
\(c_{t'}\) and the smaller nonlinear energy \(\kappa_{t'}\). The barrier varies
continuously with \(\sigma_0\), reduces to Theorem~\ref{thm:floor} at
\(\sigma_0=0\), and tends to zero as \(\sigma_0\to\infty\); it need not decrease
monotonically, because at early times the gain in \(c_{t'}\) can outweigh the
loss in \(\kappa_{t'}\).

\begin{table}[t]
\centering
\caption{Exact constants for \(\mu=1.5\), \(\tau=0.35\) and \(k=1\). Top:
\(F_\infty\) is the large-\(d\) limit of the explicit barrier for dictionaries of
size \(o(d^3)\); the Gaussian control is the score risk of the population
moment-matched Gaussian denoiser. Middle: the staircase levels of
Corollary~\ref{cor:staircase} in score-risk units, i.e.\ multiplied by
\((a_t/h_t)^2\); lower levels \(c_t\kappa_{J,t}\) apply below \(d^J/J!\), upper
levels \(\varepsilon_J\) are attained at \(\binom{d+J}J\), and
\(\bar\jmath_t=J_t/\kappa_t\). Bottom: the reference width at which the explicit
barrier vanishes with \(L=3\) bandwidths. Values use the exact formulas of
the preceding results.}
\label{tab:constants}

\begin{tabular}{rrrrrr}
\toprule
$t$ & $c_t$ & $\kappa_t$ & $F_\infty(t)$ & Gaussian control & ratio \\
\midrule
0.1 & 0.0305 & 0.0995 & 0.0755 & 2.7069 & 0.028 \\
0.3 & 0.2835 & 0.1042 & 0.0797 & 0.5241 & 0.152 \\
0.6 & 0.6149 & 0.0467 & 0.0177 & 0.0565 & 0.313 \\
1.0 & 0.8334 & 0.0108 & 0.0016 & 0.0032 & 0.510 \\
\bottomrule
\end{tabular}
\par\medskip
\begin{tabular}{rrrrrrrr}
\toprule
$t$ & $c_t\kappa_{3,t}$ & $\varepsilon_3$ & $c_t\kappa_{5,t}$ & $\varepsilon_5$ & $c_t\kappa_{7,t}$ & $\varepsilon_7$ & $\bar\jmath_t$ \\
\midrule
0.1 & 0.0755 & 1.03 & 0.03 & 0.54 & 0.0151 & 0.331 & 4.84 \\
0.3 & 0.0797 & 0.189 & 0.0238 & 0.0878 & 0.00936 & 0.0459 & 4.06 \\
0.6 & 0.0177 & 0.0127 & 0.00316 & 0.00378 & 0.000796 & 0.00132 & 3.49 \\
1.0 & 0.00163 & 0.0003 & 0.000126 & 4.2e-05 & 1.52e-05 & 7.45e-06 & 3.18 \\
\bottomrule
\end{tabular}
\par\medskip
\begin{tabular}{lrrrr}
\toprule
reference width & $d=16$ & $d=32$ & $d=64$ & $d=128$ \\
\midrule
$p$ with $3(p+1)+d+1\geq N_{d,3}$ & 260 & 1972 & 15210 & 119167 \\
\bottomrule
\end{tabular}

\end{table}

\subsection{Geometry adaptation is statistically dimension-free}\label{sec:adaptive}

\begin{theorem}[Projector dependence and dimension invariance]
\label{thm:invariance}
Under Assumption~\ref{ass:model} with \(n\geq k+1\), fix \(U\). Then almost
surely \(\widehat\Pi=UU^\top\), and the following hold.
\begin{enumerate}
\item The PCA estimator depends on \(\widehat U\) only through
\(\widehat\Pi\); with shared \(W,b\) and corruption noise it equals the oracle
estimator as a function, together with its selected ridge and bandwidth.
\item There is a map \(\widehat G_t:\R^k\to\R^k\) with
\(\widehat m_{\rm PCA}(x,t)=U\widehat G_t(U^\top x)\). The joint law of
\(\{\widehat G_t\}\) over the fitted times, the selected tuning parameters and
the validation errors is the same for every \(d\geq k\) and every \(U\). It
equals the law of the procedure run at \(d=k\) with \(U=\Id_k\).
\item Consequently \(\risk_t(\widehat s_{\rm PCA};U)\) depends on neither \(d\)
nor \(U\), for every \(t\) in the fitted interval, including interpolated times.
\end{enumerate}
\end{theorem}

\begin{theorem}[Generation invariance]\label{thm:generation}
For any estimator of the form~\eqref{eq:tweedie} whose denoiser takes values in
\(\spanop(\widehat U)\), the sampler output satisfies
\(\widehat x=\widehat m(x_{t_0},t_0)\in\spanop(\widehat U)\); in particular it
has no off-subspace component when \(\widehat\Pi=UU^\top\). For the PCA
estimator with \(n\geq k+1\), the law of the latent output \(U^\top\widehat x\)
is the same for every \(d\geq k\). Hence any distance between the generated law
and \(\mathrm{Law}(X_0)\) that is invariant under the isometry
\(z\mapsto Uz\), such as \(W_2\), does not depend on \(d\).
\end{theorem}

The invariance covers the whole implemented pipeline. It includes finite
corruption counts, held-out tuning, time interpolation, the finite terminal time,
the Euler--Maruyama discretization and the final Tweedie step. These
equalities hold at every ambient dimension. Forming ambient inputs, the
principal subspace and the projected weights still costs time and memory
that grow with \(d\) (linearly, for fixed
\(n\), \(p\) and \(k\)). Any \(k\)-dimensional statistical guarantee for the
oracle-subspace problem transfers to every ambient dimension without change.

With off-subspace noise in the training data
(Proposition~\ref{prop:offsubspace}), \(\widehat\Pi\neq UU^\top\) and the
invariance is no longer exact. If the subspace is estimated on an independent
split, the fitted latent model depends on \((d,U,\widehat U)\) only through the
\(k\times k\) overlap \(\widehat U^\top U\), because \(\widehat U^\top G\) and
\(\widehat U^\top G_0\) are then standard Gaussian in \(\R^k\). Classical
spiked-covariance asymptotics, stated for Gaussian spikes
\citep{baik2005phase,paul2007asymptotics}, predict
\(\norm{\sin\Theta(\widehat U,U)}_F^2\to
k\{1-(1-\gamma/\varrho^2)/(1+\gamma/\varrho)\}\approx k\gamma(1+\varrho)/\varrho^2\)
when \(d/n\to\gamma\) and the signal-to-noise ratio
\(\varrho=\mathrm{Var}(Z_1)/\sigma_0^2\) exceeds \(\sqrt\gamma\), and no
consistent recovery below that threshold. We use this only as a reference curve.
The Gaussian-spike formula suggests degradation governed by \(d/n\) and
\(\sigma_0\) at a fixed feature budget. Section~\ref{sec:experiments}
measures this dependence.

\subsection{A checkable separation}\label{sec:separation}

Fix \(k,t\), the sample sizes, \(p\), and the grids. Let
\(r_k(t)=\risk_t(\widehat s_{\rm PCA})\). By Theorem~\ref{thm:invariance} it
can be computed at \(d=k\).

\begin{lemma}[Monotonicity in the ambient dimension]\label{lem:monotone}
For fixed \(M\geq0\) and \(t>0\), the explicit floor \(d\mapsto F_d(M;t)\) is
nondecreasing on \(d\geq3\) and converges to
\(F_\infty(t)=k(a_t/h_t)^2c_t^k\kappa_t\). Hence, for any \(r<F_\infty(t)\), the
set \(\{d\geq3:F_d(M;t)>r\}\) equals \(\{d\geq d_0\}\) with
\(d_0=\min\{d\geq3:F_d(M;t)>r\}<\infty\); for \(r\geq F_\infty(t)\) it is empty.
\end{lemma}

\begin{corollary}[Separation at matched budgets]\label{cor:separation}
For every \(d\geq\max(3,k)\) with \(r_k(t)<F_d(L(p+1);t)\), and every
orientation \(U\), the PCA estimator has strictly smaller score risk than the
ambient estimator with the same samples, features, readout size, ridge grid and
bandwidth grid. If \(r_k(t)<F_\infty(t)\), this holds for every
\(d\geq\max(d_0,k)\), where \(d_0\) is the first crossing of
Lemma~\ref{lem:monotone} with \(M=L(p+1)\); if \(r_k(t)\geq F_\infty(t)\), the
explicit floor certifies no \(d\). The first statement also holds with the Ky
Fan floor \(F^{\mathrm{KF}}_d\).
\end{corollary}

In the experiments, \(r_k(t)\) is replaced by an upper confidence limit from
independent replicates at \(d=k\). The corresponding \(d_0\) is then the exact
first crossing of the explicit floor, and by Lemma~\ref{lem:monotone} the
comparison holds for all larger \(d\), with no upper limit. The confidence
limits are approximate (Section~\ref{sec:experiments}), so the resulting
certificate is statistical, while its deterministic part is
Corollary~\ref{cor:separation}.

\begin{remark}[Fixed-dimensional rate inputs]\label{rem:rates}
Corollary~\ref{cor:separation} needs \(r_k(t)\) to be small, not a rate. For
a rate, Theorem~\ref{thm:invariance} reduces the question to random-feature
ridge regression in fixed dimension \(k\), where existing analyses apply under
their own source and capacity conditions \citep{rudi2017generalization}.
Theorem~\ref{thm:generation} likewise reduces sampling error to a
\(k\)-dimensional reverse process, for which score-to-sample bounds of
\citet{chen2023sampling} and \citet{benton2024dlinear} apply with their
dimension replaced by \(k\). We have not verified those conditions for the
tanh dictionary and the mixture target, so we state no rate.
\end{remark}

\subsection{The covariance-matched regime: the sharp frozen exponent}\label{sec:frozen-cm}

In the covariance-matched regime the density comparison of
Lemma~\ref{lem:density} is unavailable: the hidden marginal has narrower
mixture components than the matched Gaussian, so its ratio to \(N(0,\bar\sigma_t^2)\) has
infimum zero. The one-sided change of measure (Lemma~\ref{lem:onesided}) needs
the comparison only on the dictionary, where hypercontractivity supplies it.

\begin{lemma}[Norm comparison]\label{lem:comparison}
Let \(Q=N(0,\bar\sigma^2\Id)\) and \(dP/dQ(x)=\rho(u^\top x/\bar\sigma)\) with \(\rho\) continuous and
positive. (a) If \(\rho\ge c\), then \(\|f\|_P^2\ge c\|f\|_Q^2\) for every \(f\).
(b) If every element \(\psi\) of a linear space \(V\) of scalar functions satisfies
\(\|\psi\|_{L^4(Q)}\le\kappa\|\psi\|_{L^2(Q)}\), then for every \(R>0\) with
\(\kappa^2\pi_R^{1/2}<1\), \(\pi_R=\Pr(|N(0,1)|>R)\),
\[
 \|f\|^2_{L^2(P)}\ge c_\kappa\|f\|^2_{L^2(Q)}\quad(f\in V^d),\qquad
 c_\kappa=\min_{|y|\le R}\rho(y)\,\bigl(1-\kappa^2\pi_R^{1/2}\bigr).
\]
Polynomials of degree at most \(D\) satisfy the hypothesis with
\(\kappa=3^{D/2}\), for every \(d\) \citep{odonnell2014analysis}.
\end{lemma}

Under Assumption~\ref{ass:cm} with \(k=1\), let \(r_t\) be the density ratio of
\(u^\top X_t/\bar\sigma_t\) to \(N(0,1)\) and put
\(\hat\psi=(h_t/(a_t\bar\sigma_t))\,r_t'\), with Hermite coefficients \(c_j\) in
\(L^2(N(0,1))\) and \(\Psi=\sum_jc_j^2\). Let
\(B=(h_t/(a_t\bar\sigma_t))^2\E_{N(0,1)}[r_t'^2/r_t]\), which equals the excess denoiser risk
of the Gaussian baseline, and \(W_J(d)=\sum_jc_j^2w_{j,J}(d)/\Psi\) with the
harmonic weights of~\eqref{eq:weights}. The dictionaries and the risk are as in
Definition~\ref{def:dictionary} and~\eqref{eq:haar}, with
\(D=\dim(\mathcal S+\Aff)\le M+d+1\).

\begin{theorem}[Sharp frozen exponent]\label{thm:frozen}
Under Assumption~\ref{ass:cm} with \(k=1\), let \(\nu\) be symmetric with \(L^2\)
likelihood ratio and even generative exponent \(k^\ast\ge4\), and let
\(J=k^\ast-1\). Then \(c_j=(h_t/(a_t\bar\sigma_t))\sqrt{j+1}\,e_{j+1}(t)=0\) for \(j<J\),
\(W_J(d)\to1\), and:
\begin{enumerate}
\item For \(u\) uniform on the sphere and every orientation-independent dictionary
whose span \(V\) has \(\dim V\le D\) and \(L^4/L^2\) constant \(\kappa\) under
\(Q_t\),
\[
 \bar\risk_t(\widehat s)\;\geq\;\Bigl(\frac{a_t}{h_t}\Bigr)^2\Bigl[B-\frac{\Psi}{c_\kappa}
 \Bigl(1-W_J(d)+\frac D{N_{d,J}}\Bigr)\Bigr],
\]
which is \((1-o(1))\) times the Gaussian baseline's risk \((a_t/h_t)^2B\) when
\(D=o(d^J)\) and \(\kappa\) is bounded, with \(\nu\) and \(t\) fixed as
\(d\to\infty\).
\item The span of all polynomials of degree at most \(J\), of dimension
\((1+o(1))d^J/J!\), contains at every orientation an estimator with score risk at
most \((a_t/h_t)^2(B-G)\), where \(G=c_J^2/\E[h_J(u^\top X_t/\bar\sigma_t)^2]>0\) does not
depend on \(d\).
\end{enumerate}
\end{theorem}

The dictionary size at which hypercontractive orientation-blind dictionaries
start to improve on the Gaussian baseline is therefore \(d^{k^\ast-1}\) up to
constants. For polynomial dictionaries of every fixed degree the answer is
complete: degree below \(J\) gains nothing at any size
(Corollary~\ref{cor:polybarrier}), and degree at least \(J\) gains nothing
asymptotically with \(o(d^J)\) elements. For \(\mu=0.8\), \(\sigma_0=1\)
(\(k^\ast=4\)) and cubic polynomial dictionaries (\(\kappa=3^{3/2}\)), \(G/B\) lies
between \(0.73\) and \(0.94\) for \(t\in\{0.144,0.3,0.6\}\), while for \(D\le10d\)
the lower bound is at least \(0.83B\) at \(d=128\), \(0.96B\) at \(d=512\) and
\(0.99B\) at \(d=2048\). A uniform \(L^4/L^2\) constant for spans of
\(M\) random tanh features has not been established. Without such control,
a dictionary could place its mass where the data density is small. The
exponent \(k^\ast-1\) counts dictionary elements. The
\(d^{k^\ast/2}\) scale of Section~\ref{sec:training} counts clean points
for an explicit adaptive learner; the statistical-query result there is
a tolerance bound with a heuristic sample interpretation.
\section{Kernels on the noisy inputs pay the same power in samples}\label{sec:kernels}

Kernel ridge regression on the noisy training inputs is not covered by
Section~\ref{sec:frozen}: its span \(\{K(\cdot,x_i)\}\) depends on the orientation
through the inputs. The Hermite structure still controls it, now through the
sample size. We treat positive-semidefinite dot-product kernels
\(K(x,x')=\varphi(\inner x{x'}/(d\bar\sigma_t^2))\) with \(\varphi\in C^J(\R)\), under
Assumption~\ref{ass:cm} with \(k=1\) and the notation of Theorem~\ref{thm:frozen}:
\(P=P_t\), \(Q=Q_t=N(0,\bar\sigma_t^2\Id)\), \(r=m_t-m^Q_t=u\,\bar r(u^\top x)\),
\(B=\|r\|_P^2\), and \(\rho\,(u^\top r)=\hat\psi(u^\top x/\bar\sigma_t)\) with Hermite
coefficients \(c_j=0\) for \(j<J=k^\ast-1\) and \(\Psi=\sum_jc_j^2\). For every
\(\sigma\ge0\) we assume that \(\varphi(\sigma Z)\) and \(\varphi^{(J)}(\sigma Z)\) lie
in Gaussian \(L^2\), and that
\[
 \sup_{d\ge d_0}\E\Bigl[|u^\top X_t/\bar\sigma_t|^{2J}
 \bigl\|\varphi^{(J)}\bigl(\tfrac{|X_t|}{d\bar\sigma_t}Z\bigr)\bigr\|_{L^2(Z)}^2\Bigr]<\infty,
 \qquad Z\sim N(0,1).
\]
Polynomial growth of \(\varphi^{(J)}\) suffices. For \(\varphi(z)=e^z\) the inner
squared norm is \(\exp(2|X_t|^2/(d^2\bar\sigma_t^2))\), so the condition holds for all
large \(d\) in the mixture model.

\begin{lemma}[Kernel leakage]\label{lem:leakage}
For \(y\in\R^d\setminus\{0\}\) let \(\tau_u(y)=\E_P[(u^\top r(X))\,K(X,y)]\) and
\(\sigma_y=|y|/(d\bar\sigma_t)\). Then
\[
 \tau_u(y)=\sum_{j\ge J}c_j\Bigl(\frac{u^\top y}{|y|}\Bigr)^{j}
 \E\bigl[h_j(Z)\varphi(\sigma_yZ)\bigr],\qquad
 |\tau_u(y)|\le\sqrt{\frac{\Psi}{J!}}\;\Bigl|\frac{u^\top y}{\bar\sigma_t}\Bigr|^{J}
 d^{-J}\,\bigl\|\varphi^{(J)}(\sigma_yZ)\bigr\|_{L^2}.
\]
\end{lemma}

A kernel centred at a noisy input sees the hidden residual only through the
overlap \((u^\top y/|y|)^J\), of order \(d^{-J/2}\) for a typical input, and the
\(J\)-th derivative of its profile contributes another \(d^{-J/2}\).

\begin{theorem}[Kernels on the noisy inputs pay \(d^{k^\ast-1}\)]\label{thm:kernel}
Let \(x_1,\dots,x_n\in\R^d\), \(\mathbf K=(K(x_i,x_j))_{ij}\), and
\(\widehat f(x)=Ax+b+\sum_i\alpha_iK(x,x_i)\) with \(A\), \(b\) and the rows
\(\alpha_i^\top\) of \(\alpha\in\R^{n\times d}\) arbitrary functions of the data and
of \(u\); the denoiser is \(m^Q_t+\widehat f\). Then, with \(\tau=(\tau_u(x_i))_{i\le n}\),
\[
 \|r-\widehat f\|^2_{L^2(P)}\;\geq\;B-2\,|\inner{\alpha u}{\tau}|\;\geq\;
 B-2\,\|\alpha u\|_2\,\|\tau\|_2 .
\]
Let the \(x_i\) be the noisy inputs of \(n\) independent denoising pairs, let
\(Y\in\R^{n\times d}\) be any targets with \(\E(u^\top y_i)^2\le\sigma_Y^2\), let
\(\alpha=(\mathbf K+n\lambda\Id)^{-1}Y\), and let
\(\Delta=(B-\|r-\widehat f\|_P^2)_+\le B\) be the gain over the Gaussian baseline.
Write \(T_d^2=\E\,\tau_u(X_t)^2=O(d^{-2J})\).
\begin{enumerate}
\item \emph{Any ridge.} For \(\lambda>0\), \(\E\Delta\le2\sigma_YT_d/\lambda\).
\item \emph{Self-regularized, including ridgeless.} Suppose \(\varphi\) is
real-analytic with \(\varphi_j=\varphi^{(j)}(0)/j!\ge0\) for all \(j\), let
\(\mathbf K_{\ge J}=(\varphi_{\ge J}(\inner{x_i}{x_j}/(d\bar\sigma_t^2)))_{ij}\) with
\(\varphi_{\ge J}(z)=\sum_{j\ge J}\varphi_jz^j\), and
\(E_\gamma=\{\lambda_{\min}(\mathbf K_{\ge J})\ge\gamma\}\). For every \(\lambda\ge0\)
(for \(\lambda=0\), the interpolant, defined on \(E_\gamma\)),
\(\E[\Delta\,\mathbf 1_{E_\gamma}]\le2n\,\sigma_YT_d/\gamma\), hence
\(\E\Delta\le2n\,\sigma_YT_d/\gamma+B\Pr(E_\gamma^c)\).
\end{enumerate}
Consequently, with \(\sigma_Y\) bounded and a ridge \(\lambda\gg d^{-(k^\ast-1)}\),
kernel ridge regression gains \(o(1)\) at every sample size; and if
\(\Pr(E_\gamma)\to1\) for some \(\gamma\) bounded below, it gains at most
\(O(nd^{-(k^\ast-1)})+o(1)\) at every ridge, so it needs \(n=\Omega(d^{k^\ast-1})\)
samples to gain a constant fraction of \(B\).
\end{theorem}

The diagonal of \(\mathbf K_{\ge J}\) concentrates near
\(\gamma_0=\varphi_{\ge J}(1)\) and each off-diagonal entry is \(O(d^{-J/2})\), but the
raw Taylor tail is not spectrally flat: terms of degree above \(J\) contain
lower-degree components that produce positive low-rank spikes. Such spikes do
not lower \(\lambda_{\min}\). For spherical or Gaussian inputs, kernel matrices in
polynomial regimes are a low-degree part plus \(\gamma_0\Id\) up to vanishing
operator-norm error \citep{misiakiewicz2022spectrum,lu2022equivalence}. For the
noisy inputs with one non-Gaussian direction we prove a weaker, constant lower
edge, which removes the hypothesis of part~2 for the exponential kernel.

\begin{lemma}[Self-regularization from a high-degree term]\label{lem:selfreg}
In the setting of Theorem~\ref{thm:kernel}, suppose the noisy inputs have an
independent \(N(0,\bar\sigma_t^2\Id_{u^\perp})\) component on \(u^\perp\), that \(\varphi\) has
an everywhere convergent power series with nonnegative coefficients, and that
\(\varphi_L>0\) for some integer \(L>2J\). Put \(q_i=(\Id-uu^\top)x_i/\bar\sigma_t\) and
\(G^\perp_{ij}=\inner{q_i}{q_j}/d\). For every realization,
\[
 \lambda_{\min}(\mathbf K_{\ge J})\ \ge\
 \varphi_L\Bigl\{\min_i(G^\perp_{ii})^L-(n-1)\max_{i\ne j}|G^\perp_{ij}|^L\Bigr\}.
\]
If \(n=O(d^J)\), then \(\Pr\{\lambda_{\min}(\mathbf K_{\ge J})\ge\varphi_L/2\}\to1\) as
\(d\to\infty\). For \(\varphi(z)=e^z\) one may take \(L=2J+1\) and
\(\gamma=1/[2(2J+1)!]\) in Theorem~\ref{thm:kernel}(2).
\end{lemma}

For the exponential kernel and \(J=3\), the certified constant is
\(1/(2\cdot7!)\), far below \(\gamma_0=e-5/2\). The bound is asymptotic
and does not specify a finite-dimensional threshold. Combined with
Theorem~\ref{thm:kernel}(2) it shows \(\E\Delta=o(1)\) for \(n=o(d^J)\) at every
ridge, including interpolation. Kernel self-regularization for spherical or
hypercubic inputs is established in the cited works; the lemma transfers it to
the noisy inputs of this model.
\section{Gradient learners: from \texorpdfstring{\(d^{k^\ast-1}\)}{d^(k*-1)} to the statistical-query scale}\label{sec:training}

Throughout this section Assumption~\ref{ass:cm} holds with \(k=1\), \(y=x/\bar\sigma_t\),
the residual is \(r=x_0-m^Q_t(x_t)\) against the isotropic baseline, and
\(\rho^2=\sigma_0^2h_t/\bar\sigma_t^2\) is its null variance per coordinate.
Online training of one unit takes \(d^{k^\ast-1}\) samples up to logarithms.
Repeated views and smoothing give an explicit learner at
\(d^{k^\ast/2}\) clean points up to logarithms, the scale heuristically
suggested by the statistical-query tolerance bound.

\subsection{One unit trained online pays \texorpdfstring{\(d^{k^\ast-1}\)}{d^(k*-1)}}\label{sec:online}

\begin{proposition}[One unit sees an overlap polynomial]\label{prop:neuron}
Let \(J=k^\ast-1\), \(\beta\in\R\) and, for unit \(w\),
\(f_w(x)=\beta\,h_J(w^\top x/\bar\sigma_t)\,w\). The population denoising loss of
\(m^Q_t+f_w\) is \(L(w)=L_0+\phi(\inner wu)\) with
\[
 \phi(m)=-2\beta\frac{h_t}{a_t\bar\sigma_t}\sqrt{k^\ast}\,e_{k^\ast}(t)\,m^{k^\ast}
 +\beta^2\sum_{j=k^\ast}^{2J}b_j\,e_j(t)\,m^j,\qquad b_j=\E[h_J^2h_j],
\]
a polynomial in the overlap \(m=\inner wu\) with no nonconstant terms below degree
\(k^\ast\).
\end{proposition}

\begin{corollary}[Online SGD pays \(d^{k^\ast-1}\)]\label{cor:online}
Let \(k^\ast\ge3\) and choose \(\beta\) with \(\beta e_{k^\ast}(t)>0\) and \(|\beta|\) small
enough that \(\phi'<0\) on \((0,1)\). Run online spherical SGD on the per-sample
denoising loss, one fresh pair per step, with learning rate \(\delta_d/d\), from
\(w_0\) uniform on \(\{w\in\Sph^{d-1}:\inner wu\ge0\}\), for \(n=\alpha_dd\) steps
with \(\alpha_d\) at most polynomial. If \(\alpha_d\gg d^{k^\ast-2}\log^2d\) and
\(\alpha_d^{-1}\ll\delta_d\ll\alpha_d^{-1/2}\), then \(\inner{w_n}u\to1\) in
probability; if \(\alpha_d\ll d^{k^\ast-2}\) and \(\delta_d=O(\alpha_d^{-1/2})\), then
\(\sup_{j\le n}|\inner{w_j}u|\to0\) in probability. The sample cost is
\(d^{k^\ast-1}\) up to the factor \(\log^2d\).
\end{corollary}

The corollary applies \citet[Theorems~1.3--1.4]{bgj2021} to the loss of
Proposition~\ref{prop:neuron}, whose information exponent is \(k^\ast\); the proof
checks their moment conditions. It matches the cubic cost found by
\citet{bardone2026easyhard} for the fourth cumulant in a related model.

\subsection{The statistical-query limit}\label{sec:sq}

An \(L^2\) statistical query of tolerance \(\tau\) on the denoising pair is a
function \(\psi(x_0,x_t)\) with \(\E_Q\psi^2\le1\) under the Gaussian null
(\(X_0\sim Q_0\)); the oracle may return any \(v\) with \(|v-\E_P\psi|\le\tau\),
adversarially. Such queries are simulable from about \(\tau^{-2}\) samples only
when \(\psi\) has controlled variance and tails under the data law, so the result
below is a bound on the tolerance; a sample bound follows only heuristically.

\begin{proposition}[Statistical-query cost of score estimation]\label{prop:sq}
Assume the \(L^2\) hypothesis of Theorem~\ref{thm:noiseop}, let \(\chi^2=\chi^2_0\),
and let \(S\) be a set of \(N\) unit vectors with \(|\inner uv|\le\beta\) for
\(u\ne v\). Consider a deterministic algorithm that makes \(q\) queries of tolerance
\(\tau\) and, for every \(u\in S\) and every answer sequence consistent with
\(P_u\), outputs \(f\) with \(\|r_u-f\|^2_{L^2(P_t)}\le(1-\delta)B\) and
\(\|f\|_{L^2(Q_t)}\le\Lambda\sqrt B\). If
\(2\beta^{k^\ast}\chi^2<\tau^2\) and
\(\tau<\tau_0:=\delta \bar\sigma_t\sqrt B/(2\sigma_0\sqrt{h_t}\,\Lambda)\), then
\(q+1\ge N\tau^2/(2\chi^2)\). Packings with \(N=\exp(d^c)\) and
\(\beta=Cd^{-(1-c)/2}\) exist for \(0<c<1\); with them, an algorithm making
polynomially many queries needs \(\tau\le\sqrt{2\chi^2}\,C^{k^\ast/2}d^{-k^\ast(1-c)/4}\)
when \(\tau_0\) stays above this level. Letting \(c\to0\) slowly, the tolerance
must be \(d^{-k^\ast/4+o(1)}\), which corresponds heuristically to
\(n\gtrsim d^{k^\ast/2-o(1)}\) samples.
\end{proposition}

Score estimation for such distributions is also hard under cryptographic
assumptions \citep{song2024crypto}; the proposition needs none, within the oracle
model and under the norm assumption on the output. Restricting the oracle to the
noisy marginal multiplies the \(j\)-th correlation by
\(\lambda_t^{2j}\le\lambda_t^{2k^\ast}\), so such learners are weaker.

\paragraph{Terminology.} Proposition~\ref{prop:sq} bounds a query
\emph{tolerance}. Simulating a query of tolerance \(\tau\) from samples takes about
\(\tau^{-2}\) samples when the query has controlled variance and tails, which
gives the heuristic count \(n\approx d^{k^\ast/2}\). We call this the
\emph{statistical-query scale}. When a section or theorem below is said to be
``at the statistical-query scale'', it means an upper bound of order
\(d^{k^\ast/2}\) up to logarithmic factors; no matching lower bound for
gradient learners on the denoising loss is claimed.

\subsection{One gradient step and reused views}\label{sec:views}

For \(k^\ast=4\) let
\(D(x_0,x_t)=\sum_cr_c\,\partial_{y_c}\mathcal H(y)=2(r^\top y)(yy^\top-\Id)+(|y|^2-d-4)(ry^\top+yr^\top)\).

\begin{theorem}[Denoising partial trace, \(k^\ast=4\)]\label{thm:partialtrace}
Assume \(k^\ast=4\) and, in part~3, that \(Z\) is sub-Gaussian. Let
\(\bar\kappa_t=\lambda_t^4\tilde\kappa\), \(X=(a_t\bar\sigma_t/(4h_t))D(X_0,X_t)\) and
\(v_t=a_t^2\sigma_0^2/(2h_t)\). Then:
\begin{enumerate}
\item \(\E X=\bar\kappa_t\,uu^\top\) exactly, for every \(d\).
\item \(\frac1n\sum_iX_i=-\frac{a_t\bar\sigma_t}{8h_t}\nabla_B\widehat L(0)\), where
\(\widehat L\) is the empirical denoising loss of the score model
\(m^Q_t+\nabla_y\inner B{\mathcal H(y)}\), \(B\) symmetric: the statistic is one
full-batch gradient step from zero.
\item \(\|\E X^2\|=v_td^2(1+O(1/d))\), and for \(\delta,\varepsilon\in(0,1)\) the
eigenvector \(\widehat u\) of \(\frac1n\sum_iX_i\) whose eigenvalue has the largest
modulus satisfies \(\sin\angle(\widehat u,u)\le\varepsilon\) with probability at least
\(1-\delta\) whenever
\[
 n\ge C\Bigl(\frac{v_t}{\bar\kappa_t^2}\,\frac{d^2\log(4d/\delta)}{\varepsilon^2}
 +\frac{d^{3/2}\log^{5/2}(nd/\delta)}{\varepsilon|\bar\kappa_t|}\Bigr),
\]
with \(C\) depending only on \(t\), \(\sigma_0\) and \(\nu\).
\item The leading constant
\(v_t/\bar\kappa_t^2=\bar\sigma_t^8/(2h_ta_t^6\sigma_0^6\tilde\kappa^2)\) is minimized at
\(t^\ast=\frac12\log\{(3+\sigma_0^2)/3\}\), the time of Theorem~\ref{thm:time}.
\end{enumerate}
\end{theorem}

A \(d^2\) rate for \(k^\ast=4\) is expected from fourth-moment tensor unfolding
\citep{montanari2014tensor} and spectral methods for non-Gaussian component
analysis \citep{tan2018ngca}. One gradient step of the denoising loss
computes the unfolding, with an explicit leading constant optimized at
\(t^\ast\). Theorem~\ref{thm:multiview} extends the calculation to every
even \(k^\ast\) and to \(r\) views of each clean point.

\begin{theorem}[Reused views of the denoising partial trace]\label{thm:multiview}
Let \(k^\ast=2p+2\ge4\) be even, assume the \(L^2\) hypothesis of
Theorem~\ref{thm:noiseop}, and define for a view
\[
 X(x_0,x_t)=\frac{a_t\bar\sigma_t^2}{h_tk^\ast}\Bigl[\inner{x_0-m^Q_t(x_t)}{\nabla_x\bigl(M_p(x_t/\bar\sigma_t)\bigr)_{bc}}\Bigr]_{b,c\le d}.
\]
For \(r\in\{1,2,\dots\}\) let \(Y_i\) be the average of \(X\) over \(r\) independent
views of the clean point \(x_0^{(i)}\); for \(r=\infty\) let \(Y_i=\E[X\mid x_0^{(i)}]\).
Put \(\bar Y_n=\frac1n\sum_{i\le n}Y_i\). Then:
\begin{enumerate}
\item \(\E[Y_i\mid x_0^{(i)}]=\lambda_t^{k^\ast}M_p(x_0^{(i)}/\sigma_0)\) and
\(\E Y_i=\lambda_t^{k^\ast}\tilde\kappa\,uu^\top\).
\item \(\bar Y_n=-\frac{a_t\bar\sigma_t^2}{2h_tk^\ast}\nabla_B\widehat L_r(0)\) for the score
model \(m^Q_t(x)+\nabla_x\inner B{M_p(x/\bar\sigma_t)}\), with the loss averaged over the
\(r\) views of each of the \(n\) clean points.
\item Under the null \(P_0=Q_0\),
\[
 \begin{aligned}
 \E Y_i^2
 &=\lambda_t^{2k^\ast}\Bigl(1+\frac{\iota_t-1}r\Bigr)c_p\,d^{p+1}
 \bigl(1+O(1/d)\bigr)\Id,\\
 c_p&=2^pp!,\qquad
 \iota_t=\frac{a_t^2\sigma_0^2}{h_tk^\ast\lambda_t^{2k^\ast}}
 =\frac{\tilde\kappa^2}{k^\ast!\,S_t^2}\ge1 .
 \end{aligned}
\]
and under the data law \(\|\E Y_i^2\|\) is at most \(3^{k^\ast}(1+\chi^2)^{1/2}\)
times the null value.
\item There are constants \(C,C'\) depending only on \(k^\ast\) and \(\chi^2\) such that,
with probability at least \(1-\delta\), the eigenvector \(\widehat w\) of
\(\bar Y_n\) whose eigenvalue has the largest modulus satisfies
\(\sin\angle(\widehat w,u)\le\varepsilon\) whenever
\[
 \begin{aligned}
 n\ge{}& C\Bigl(1+\frac{\iota_t-1}r\Bigr)
 \frac{d^{k^\ast/2}\log(4d/\delta)}{\tilde\kappa^2\varepsilon^2}\\
 &+C'\Bigl(1+\frac{\iota_t-1}r\Bigr)^{1/2}
 \frac{d^{(k^\ast+2)/4}\log^{k^\ast/2+1}(nd/(\delta\varepsilon))}
 {|\tilde\kappa|\,\varepsilon}.
 \end{aligned}
\]
\item \(\iota_t\) is minimized at the time \(t^\ast\) of Theorem~\ref{thm:time}; for
\(r=\infty\) the leading constant depends on neither \(t\) nor \(\sigma_0\).
\end{enumerate}
\end{theorem}

The factor \(\lambda_t^{k^\ast}\) multiplies signal and noise alike and
cancels. With full reuse, the statistical constant is the clean-data value
\(c_p/\tilde\kappa^2\), independent of diffusion time. A single view
inflates the noise by \(\iota_t\), the ratio of forward-process variance to
clean variance. This factor is at least one and is smallest at \(t^\ast\).
The exact null variance constant \(c_p=2^pp!\) counts the permutations
that preserve the pairing of the partial trace.

\paragraph{Presentations under reuse.} The case \(r=\infty\) is not a number of
presentations. By part~1, \(Y_i=\lambda_t^{k^\ast}M_p(x_0^{(i)}/\sigma_0)\) is an explicit
polynomial of the clean point, so a learner that has the clean points can compute
the full-reuse statistic directly, with no noise and at the arithmetic cost of one
view, \(O(d^2)\) per point; \(\sigma_0\) is the data standard deviation, and the factor
\(\lambda_t^{k^\ast}\) does not change the eigenvectors. A learner that sees only noisy
views approaches this statistic as \(r\) grows, at \(r\) presentations per clean point.
Minibatch SGD that cycles over \(n\) stored clean points with fresh noise uses, on
average, (number of presentations)\(/n\) views per point.

\begin{remark}[Sharp constants and the spectral transition for \(k^\ast=4\)]\label{rem:bbp}
For \(k^\ast=4\), \(c_1=2\), the null moment is exactly
\(\E\mathcal H^2=(2d^2+10d+12)\Id\), and the argument of
Theorem~\ref{thm:partialtrace}(3) shows that the data law changes \(\|\E Y^2\|\) only
by a factor \(1+O(1/d)\). With \(r=1\), \(\lambda_t^8\iota_tc_1=v_t\), recovering
Theorem~\ref{thm:partialtrace}. At \(t^\ast\) and \(\sigma_0=1\),
\(\iota_{t^\ast}=64/27\); for \(\mu=0.8\) the leading constants are \(7.06\) with one
view and \(2/\tilde\kappa^2=2.98\) with full reuse. A computation of the four-index
covariance of the noise matrix \(\bar Y_n-\E\bar Y_n\) under the null, by the same
permutation count as in the proof of Theorem~\ref{thm:multiview}(3), shows that its
leading terms are those of a Wigner matrix with entry variance
\(c_p\lambda_t^{2k^\ast}(1+(\iota_t-1)/r)d^{p}/n\) plus a random multiple of the
identity; we have not proved eigenvector universality for this ensemble. Heuristically, if its extreme eigenvectors behave as for Wigner matrices
\citep{benaych2011eigenvalues}, the leading eigenvector acquires a nontrivial
overlap at
\begin{equation}\label{eq:bbp}
 n_c=\Bigl(1+\frac{\iota_t-1}r\Bigr)\frac{c_p\,d^{k^\ast/2}}{\tilde\kappa^2},
 \qquad \sin^2\angle(\widehat w,u)\to\frac{n_c}{n}\quad(n>n_c).
\end{equation}
This is a heuristic prediction, not a theorem. Section~\ref{sec:exp-cm} finds the
simulated overlaps consistent with it, within Monte Carlo error, for
\(n\ge3n_c\) at \(d\in\{64,128,256\}\); a finite grid cannot establish the asymptotic
formula.
\end{remark}

\subsection{Smoothing creates the curvature that reuse estimates}\label{sec:smooth}

The matrix of Theorem~\ref{thm:multiview} belongs to a matrix-valued score
model. A standard network has vector-valued units whose landscape is flat
at the origin. Gaussian smoothing of the single-unit landscape creates
the curvature measured by the partial trace.

\begin{proposition}[Smoothing exposes the partial trace]\label{prop:smoothing}
Let \(k^\ast=2p+2\) and, for \(w\in\R^d\),
\(u_w(x)=(\bar\sigma_t/k^\ast)\,\He_{k^\ast}(x/\bar\sigma_t)[w^{\otimes k^\ast}]\); for unit \(w\),
\(\nabla u_w(x)=\He_{k^\ast-1}(w^\top x/\bar\sigma_t)\,w\) is the field of the unit of
Proposition~\ref{prop:neuron}. Let \(\widehat C_r(w)\) be the average of
\(\inner{x_0-m^Q_t(x_t)}{\nabla u_w(x_t)}\) over \(r\) views of each of \(n\) clean
points (\(r=\infty\): the conditional mean), and
\(\widehat C_{r,\sigma}(w)=\E_z\widehat C_r(w+\sigma z)\) with \(z\sim N(0,\Id)\). Then
\(\widehat C_\infty(w)=(h_t\lambda_t^{k^\ast}/(a_t\bar\sigma_t))\,\widehat T[w^{\otimes k^\ast}]\)
with \(\widehat T=\frac1n\sum_i\He_{k^\ast}(x_0^{(i)}/\sigma_0)\),
\[
 \widehat C_{\infty,\sigma}(w)=\frac{h_t\lambda_t^{k^\ast}}{a_t\bar\sigma_t}\sum_{j=0}^{p+1}
 \binom{k^\ast}{2j}(2j-1)!!\,\sigma^{2j}\,
 \widehat T\bigl[w^{\otimes(k^\ast-2j)},\Id^{\otimes j}\bigr],
\]
and for every \(r\) and \(\sigma\ge0\)
\[
 \nabla^2_w\widehat C_{r,\sigma}(0)=\frac{h_t}{a_t\bar\sigma_t}\,k^\ast(k^\ast-1)(k^\ast-3)!!\,\sigma^{k^\ast-2}\,\bar Y_n .
\]
The unsmoothed landscape is flat to order \(k^\ast-1\) at the origin. As
\(\sigma\to\infty\), at every fixed \(w\) with \(\bar Y_nw\ne0\), the normalized smoothed
gradient \(\nabla_w\widehat C_{r,\sigma}(w)/|\nabla_w\widehat C_{r,\sigma}(w)|\) converges to
\(\bar Y_nw/|\bar Y_nw|\), since the constant multiplying \(\bar Y_n\) above is positive;
iterating the normalized smoothed gradient map is
therefore, in the limit, the power method on \(\bar Y_n\), which converges from almost
every start to the eigenvector of largest-modulus eigenvalue (when that eigenvalue
is simple in modulus) whatever the sign of
\(\tilde\kappa\). (Plain gradient \emph{ascent} with a small step would instead favour
the most positive eigenvalue, which is the wrong end when \(\tilde\kappa<0\).)
\end{proposition}

\begin{figure}[t]
\centering
\includegraphics[width=\linewidth]{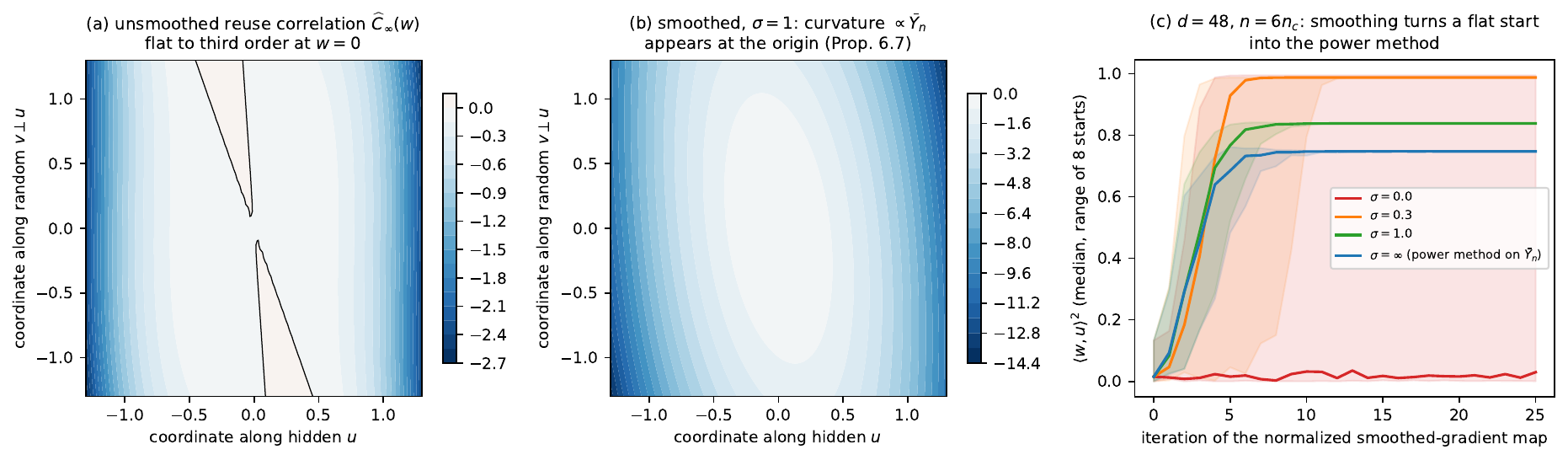}
\caption{Smoothing, computed on simulated clean data (\(d=48\), \(n=6n_c\) points,
full reuse). (a)~The reuse correlation \(\widehat C_\infty(w)\) of one unit on the
plane spanned by the hidden direction \(u\) and a random \(v\perp u\): it is a
quartic, flat to third order at the origin, so a small random start sees no
gradient toward \(u\). (b)~After Gaussian smoothing with \(\sigma=1\), the quadratic
term \(6\sigma^2w^\top\widehat Hw\), proportional to the partial-trace statistic
\(\bar Y_n\), creates curvature at the origin aligned with \(u\)
(Proposition~\ref{prop:smoothing}; the constant \(3\sigma^4\tr\widehat H\) is omitted).
(c)~Squared overlap under the normalized smoothed-gradient map from eight random
starts. Without smoothing the iteration stays near the equator. With
\(\sigma=\infty\) it is the power method on \(\bar Y_n\) and stops at the spectral
overlap, close to \(1-n_c/n\) (Remark~\ref{rem:bbp}). At moderate \(\sigma\) the quartic
term also acts once the iterate is warm, which is the sharpening step of the
two-stage learner.}
\label{fig:smoothing}
\end{figure}

The matrix computed by one gradient step of the matrix model is, up to a
constant, the curvature at the origin of the smoothed single-unit landscape
(Figure~\ref{fig:smoothing}).

The partial-trace matrix is, up to a constant, the derivative of
\(\E_{w\sim N(0,\Sigma)}\widehat C_r(w)\) with respect to the covariance \(\Sigma\) at
\(\Sigma=\sigma^2\Id\), the covariance of a Gaussian ensemble of first-layer weights.
Smoothing as a route from the information exponent to \(d^{k/2}\) is due to
\citet{damian2023smoothing} for online single-index learning, and Gaussian
homotopy reaches the unfolding threshold for tensor PCA
\citep{anandkumar2017homotopy}. Here both smoothing and reuse act on the
denoising loss, reuse removes the forward-process noise exactly, and the
curvature created by smoothing is the partial-trace statistic.

\subsection{A two-stage learner at the statistical-query scale}\label{sec:twostage}

\begin{figure}[t]
\centering
\includegraphics[width=0.9\linewidth]{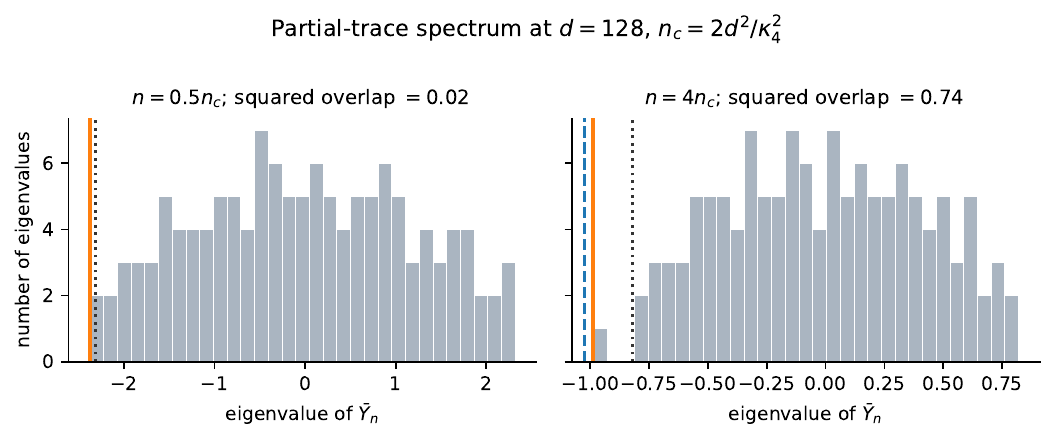}
\caption{The first stage of the two-stage learner:
eigenvalues of the full-reuse partial-trace matrix \(\bar Y_n\) for a simulated
covariance-matched mixture with \(\mu=0.8\), \(d=128\), \(u=e_1\) and
\(k^\ast=4\). The two panels use \(n=0.5n_c\) and \(n=4n_c\), where
\(n_c=2d^2/\kappa_4^2\). The dotted line is a null bulk-edge heuristic, and
the dashed line in the right panel is the corresponding outlier prediction.
The orange line marks the observed leading eigenvalue. Its eigenvector
has little squared overlap with \(u\) below the transition and substantial
overlap above it. Theorem~\ref{thm:multiview} supplies a finite-sample recovery
bound; these edge and outlier locations are heuristic.}
\label{fig:spectrum}
\end{figure}

The learner uses three independent sets of clean points: \(S_1\) of size \(n_1\)
with \(r\) views each, \(S_2\) of size \(n_2\) with \(r_2\) views each, and \(S_3\) of size
\(n_3\) with one view each.
\begin{enumerate}[leftmargin=*]
\item \emph{Geometry.} \(\widehat w_1\) is the eigenvector of the largest-modulus
eigenvalue of \(\bar Y_{n_1}\) computed on \(S_1\): the limit of the normalized
smoothed-gradient iteration (Proposition~\ref{prop:smoothing}), or the leading
direction of one gradient step of the matrix model (Theorem~\ref{thm:multiview}).
\item \emph{Sharpening.} \(\widehat v\) is the average over the views of \(S_2\) of
\((x_0-m^Q_t(x_t))\He_{k^\ast-1}(\widehat w_1^\top x_t/\bar\sigma_t)\), that is, \(-\frac12\) times
the gradient of the denoising loss of the untied unit
\(\He_{k^\ast-1}(\widehat w_1^\top x/\bar\sigma_t)\,v\) at \(v=0\); set
\(\widehat w=\widehat v/|\widehat v|\).
\item \emph{Readout.} \(\widehat\beta\) minimizes the empirical denoising loss on
\(S_3\) of \(m^Q_t+\beta\,h_{k^\ast-1}(\widehat w^\top x/\bar\sigma_t)\,\widehat w\).
\end{enumerate}
No step uses the hidden direction or the sign of \(\tilde\kappa\).
Figure~\ref{fig:spectrum} shows the spectral behaviour of the first stage
on the covariance-matched mixture.

 Let
\(G_\ast=(k^\ast h_t^2/(a_t^2\bar\sigma_t^2))\,e_{k^\ast}(t)^2/\E_Ph_{k^\ast-1}(u^\top X_t/\bar\sigma_t)^2\), the
denoiser gain of the best tied unit at the true direction; \(0<G_\ast\le B\) and
neither depends on \(d\).

\begin{theorem}[A two-stage denoising learner at the statistical-query scale]\label{thm:twostage}
Let \(k^\ast\ge4\) be even, assume the \(L^2\) hypothesis of
Theorem~\ref{thm:noiseop}, and assume \(\E|Z|^{4(k^\ast-1)}<\infty\) (implied by
Assumption~\ref{ass:hidden}). There are constants \(C_1,\dots,C_4\), not depending
on \(d\), such that for
\(\varepsilon,\delta\in(0,\frac12)\), if
\[
 n_1\ge C_1\Bigl(1+\frac{\iota_t-1}r\Bigr)d^{k^\ast/2}\log\frac{4d}\delta
 +C_2\,d^{(k^\ast+2)/4}\log^{k^\ast/2+1}\frac{n_1d}\delta,\qquad
 n_2\ge\frac{C_3\,d}{\delta\varepsilon},\qquad n_3\ge\frac{C_4}{\delta\varepsilon},
\]
then with probability at least \(1-3\delta\) over \(S_1,S_2,S_3\) and their views:
\(\sin^2\angle(\widehat w_1,u)\le\frac14\), \(\sin^2\angle(\widehat w,u)\le\varepsilon\), and
the conditional score risk of the learned denoiser,
\((a_t/h_t)^2\,\E[\|\widehat m(X_t)-m_t(X_t)\|^2\mid\mathcal F]\) for an independent test point
\(X_t\), where \(\mathcal F\) is generated by all training clean points and all their
views, is at most \((a_t/h_t)^2[B-(1-\varepsilon)G_\ast]\). The constants depend on
\(k^\ast\), \(t\) and \(\sigma_0\), and on \(\nu\) only through \(\tilde\kappa\), \(\chi^2\), the
Hermite moments \(e_j(t)\) for \(j\le2k^\ast-2\), \(G_\ast\) and \(\E|Z|^{4(k^\ast-1)}\).
\end{theorem}

Finding the geometry costs
\(d^{k^\ast/2}\log d\) clean points and needs only constant accuracy. Sharpening
to accuracy \(\varepsilon\) costs \(O(d/\varepsilon)\), because one power step from a
warm start divides the noise by a power of the overlap rather than of
\(d^{-1/2}\). The scalar readout costs \(O(1/\varepsilon)\), independent of
\(d\). The total cost in noisy presentations is \(rn_1+r_2n_2+n_3\), and the
arithmetic cost is \(O(rn_1d^2+d^3+r_2n_2d+n_3d)\).

\begin{corollary}[A dimension separation for \(k^\ast=4\)]\label{cor:dimsep}
Under Assumption~\ref{ass:cm} with \(k=1\), \(k^\ast=4\) and the two-component
mixture, to improve the score at time \(t\) on the Gaussian baseline by a constant
fraction of \(B\): the two-stage learner needs \(O(d^2\log d)\) clean points, with
any number \(r\ge1\) of views per point, with probability at least \(1-3\delta\); every
deterministic \(L^2\) statistical-query algorithm that makes polynomially many queries
and satisfies the output-norm assumption of Proposition~\ref{prop:sq} with \(\Lambda\)
bounded independently of \(d\) needs tolerance
\(d^{-1+o(1)}\); whereas hypercontractive orientation-blind dictionaries need
\(\Omega(d^3)\) elements (Theorem~\ref{thm:frozen}), kernel ridge regression with the
exponential kernel on the noisy inputs needs \(\Omega(d^3)\) samples at every ridge
under the target-moment assumption of Theorem~\ref{thm:kernel}
(Lemma~\ref{lem:selfreg}), and spherical online SGD on one denoiser unit, in the
initialization and learning-rate regime of Corollary~\ref{cor:online}, needs \(d^3\)
samples up to \(\log^2d\).
\end{corollary}

The corollary compares the explicit two-stage learner with the restricted
classes in its statement. Section~\ref{sec:exp-cm} separately measures a
network trained by SGD.

\begin{conjecture}[Implicit smoothing]\label{conj:sgd}
Fix an even \(k^\ast\ge4\), a width \(m\), a batch size \(b\), an initial output scale
\(c_0\ne0\) and an interval \([\eta_0,\eta_1]\subset(0,\infty)\), none depending on \(d\).
Train a two-layer network \(\sum_{j\le m}\He_{k^\ast-1}(w_j^\top x/\bar\sigma_t)\,v_j\), with each
\(w_j\) initialized independently and uniformly on the sphere and renormalized after
every step and \(v_j=c_0w_j\) at initialization, by minibatch SGD on the denoising loss
with a constant learning rate \(\eta\in[\eta_0,\eta_1]\) for at most \(\mathrm{poly}(d)\)
steps, each batch revisiting the \(n\) stored clean points with fresh noise. Replace the
output layer of the final iterate by the tied readout
\(\sum_j\beta_j\He_{k^\ast-1}(w_j^\top x/\bar\sigma_t)\,w_j\), with \((\beta_j)\) fitted by least
squares on \(O(1)\) independent calibration pairs. Then there is \(C<\infty\) such that,
if \(n\ge Cd^{k^\ast/2}\log^Cd\), the resulting denoiser gains at least \(G_\ast/2\) over the
Gaussian baseline with probability at least \(2/3\).
\end{conjecture}

The positive lower bound on \(\eta_0\) encodes the requirement that SGD
fluctuations be large enough to smooth. A proof
would identify the implicit smoothing of SGD \citep{kleinberg2018alternative} with
the curvature of Proposition~\ref{prop:smoothing}, and would need to control the
noise of a nonstationary, data-dependent smoothing, which the explicit learner
avoids.
\section{Experiments}\label{sec:experiments}

The experiments measure risk barriers, dimension invariance, threshold
locations and leading constants. For each study, we fixed the prediction,
statistic and decision rule in a tracked protocol before collecting data.
We report every pre-stated outcome, including the failed predictions in
Section~\ref{sec:exp-failures}.
Appendix~\ref{app:experiments} gives the protocols, hardware, seeds and
record definitions.

\newcommand{\NcReuse}{2.98}
\newcommand{\NcSingle}{7.06}
\newcommand{\KappaSq}{0.671}
\newcommand{\BBPMaxErr}{0.019}
\newcommand{\BBPMaxSE}{0.029}
\newcommand{\BBPMaxZ}{1.8}
\newcommand{\XTotA}{18.7}
\newcommand{\XTotB}{6.8}
\newcommand{\XTotC}{4.0}

\subsection{Covariance-visible regime}\label{sec:exp-cv}

All runs use Assumption~\ref{ass:model} with \(\mu=1.5\), \(\tau=0.35\), \(r=8\)
corruptions per example, ridge grid \(\{10^{-6},\dots,10^{-1}\}\) and \(L=3\)
bandwidths, with selection on held-out denoising error only. Ambient, PCA and
oracle methods share weights, data and corruption noise within a replicate, and
each replicate draws a fresh Haar orientation. Score risks are Monte Carlo
averages against exact population scores (Lemma~\ref{lem:posterior}). The main
suite has 168 units (barrier, feature budget, certificate, generation) and ran
on one Tesla~T4 in \(4.9\) GPU-hours; the revision suite has 60 units (staircase,
data off the subspace, generation off the subspace) and ran on a 16-core CPU.
The pre-stated checks were: (i) no detected barrier violation, (ii) PCA and
oracle denoisers equal to floating-point precision with noiseless data, (iii)
PCA risk and \(W_2\) invariant in \(d\) up to Monte Carlo error, (iv) zero
off-subspace mass, and (v) the full level-\(J\) Hermite dictionary at the
dimension-free level \(k(a_t/h_t)^2\varepsilon_J(t)\) at every \(d\). All five
hold in the tested configurations.

\paragraph{Barrier and invariance.} No cell of the barrier study (288 cells)
falls more than three standard errors below the Ky Fan floor. PCA and oracle
fitted denoisers differ by at most \(3.8\times10^{-12}\), and the PCA risk varies
across \(d\in\{8,\dots,256\}\) with maximal \(|z|=3.12\) over all configurations.
At \(t=0.3\) the PCA risk stays near \(0.003\) for every \(d\), while the ambient
risk rises to the Gaussian control's value, about \(0.52\), once \(d\ge64\), even
with \(p=1024\) features.

\begin{figure}[t]
\centering
\includegraphics[width=\linewidth]{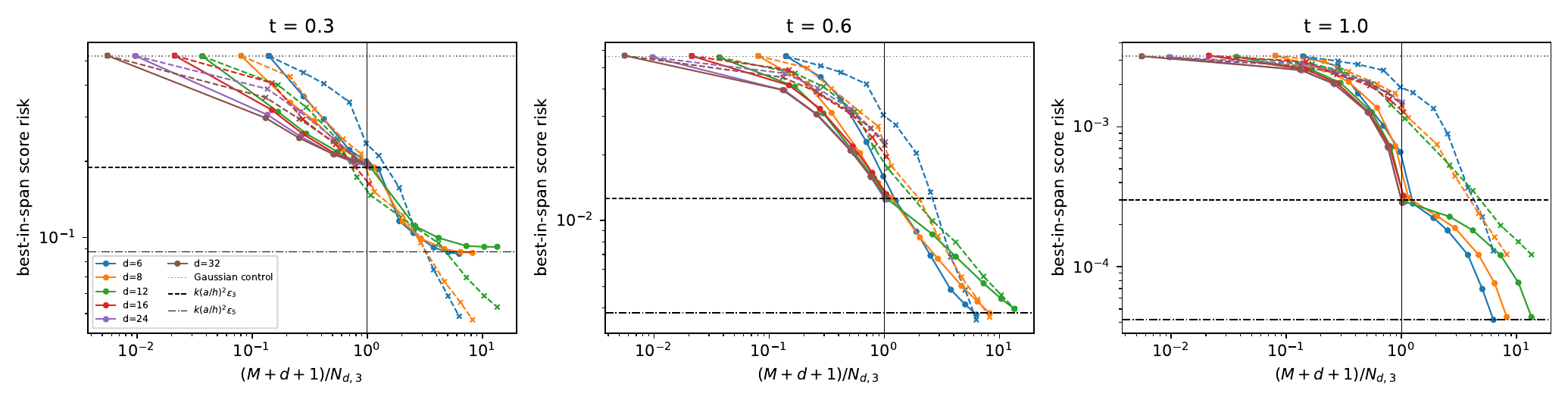}
\caption{Best-in-span score risk of orientation-blind dictionaries against the
normalized dimension \((M+d+1)/N_{d,3}\) (covariance-visible regime). Circles:
affine functions, full lower levels and a random fraction of the cubic Hermite
level; crosses: tanh random features of equal size; dashed: the explicit
barrier. The curves for \(d=6\)--\(32\) collapse, and the full cubic level reaches
the dimension-free level of Theorem~\ref{thm:sharp}.}
\label{fig:staircase-main}
\end{figure}

\paragraph{Staircase.} In the best-in-span study, none of 324 cells falls below
the floor. The full cubic Hermite dictionary reaches the predicted dimension-free
level within \(8\%\) at every \(d\in\{6,\dots,32\}\) (dimensions \(63\) to \(6017\)),
and the quintic dictionary reaches \(\varepsilon_5\) within \(5\%\).
Normalized by \(N_{d,3}\), the curves at \(t\in\{0.6,1\}\) coincide across \(d\)
(Figure~\ref{fig:staircase-main}); tanh random features of equal size stay four to
six times above the cubic level at \(t=1\). The approach to the cubic level is
convex in \(M/N_{d,3}\), so the explicit barrier is conservative in shape but
exact in location.

\paragraph{Exact height.} Theorem~\ref{thm:height}, in its Ky Fan form and with
the time change of Proposition~\ref{prop:offsubspace} off the subspace, was
checked against all \HeightCells{} cells with \(k=1\). The number of cells more
than three standard errors below the bound is \HeightViolations; the bound
exceeds the cubic floor of Theorem~\ref{thm:floor} in \HeightImproved{} cells and
reaches up to \HeightMaxRatio{} of the measured risk. For the ambient estimator
at \(d\ge64\) and \(t\ge0.6\) (\HeightAmbientCells{} cells) it lies between
\HeightAmbientMin{} and \HeightAmbientMax{} of the measured risk, so the failure
of geometry-blind features to improve on the Gaussian control is largely
accounted for by the bound.

\paragraph{Certificate and generation.} With simultaneous Bonferroni--\(t\)
confidence limits, the separation of Corollary~\ref{cor:separation} is certified
for every \(d\ge49\) in all 75 cells in which the \(d=k\) PCA estimator has enough
samples, and by Lemma~\ref{lem:monotone} with no upper limit on \(d\). At
\(d=512\), ambient random features reproduce the Gaussian control's latent
\(W_2\) to three digits for all three latent laws, while PCA random features reach
\(W_2\) between \(0.18\) and \(0.31\), close to the exact-score sampler, including
for a four-component law outside the theorem's hypotheses. Off-subspace mass is at
most \(4.5\times10^{-13}\).

\paragraph{Data off the subspace.} With training data \(UZ+\sigma_0G_0\), no
Gaussian-control or ambient cell (384 cells) falls below the time-changed floor.
PCA and oracle-projection risks differ by a median of \(1.3\%\). The PCA
improvement over the Gaussian control is nearly independent of \(d\) until the
subspace error \(\|\sin\Theta\|_F^2\) reaches about \(0.23\)
(\(d\sigma_0^2/n\ge\frac12\)), where it collapses while the oracle retains
\(30\)--\(80\%\); this degradation is measured, not bounded.

\subsection{Covariance-matched regime}\label{sec:exp-cm}

All runs use Assumption~\ref{ass:cm} with \(k=1\), \(\sigma_0=1\), the
two-component latent law with \(\mu=0.8\) (\(k^\ast=4\), \(\kappa_4^2=\KappaSq\)) and
\(t=t^\ast=\frac12\log\frac43\) unless stated.

\paragraph{Polynomial barrier (X3).} Nested quadratic Hermite dictionaries fitted
by ridge regression on \(4{,}096\) observable pairs, for \(d\in\{8,16,32\}\) and
four sizes each, eight seeds: all 96 held-out gains over the Gaussian baseline are
negative (the largest is \(-0.0049\)), and no condition has a significant positive
mean after Holm correction. Corollary~\ref{cor:polybarrier} predicts zero
population gain; the negative finite-sample gains are estimation noise.

\paragraph{Kernels (X5).} Kernel ridge regression with the exponential kernel on
noisy inputs, \(d\in\{8,11,16\}\), \(n\le2d^3\), three seeds and ridges
\(10^{-8}\)--\(10^2\): the largest gain over the Gaussian baseline is \(10^{-3}\) of
\(B\), even when the hidden direction is used to pick the best ridge and to
discard off-direction outputs. The self-regularization event holds only slowly:
\(\lambda_{\min}(\mathbf K_{\ge3})/\gamma_0\) rises from \(0.02\)--\(0.10\) at
\(d\in\{32,64\}\) to \(0.41\)--\(0.53\) at \(d=512\). The certificate of
Lemma~\ref{lem:selfreg} was positive and below the measured eigenvalue in all
nine seeded checks.

\begin{figure}[t]
\centering
\includegraphics[width=\linewidth]{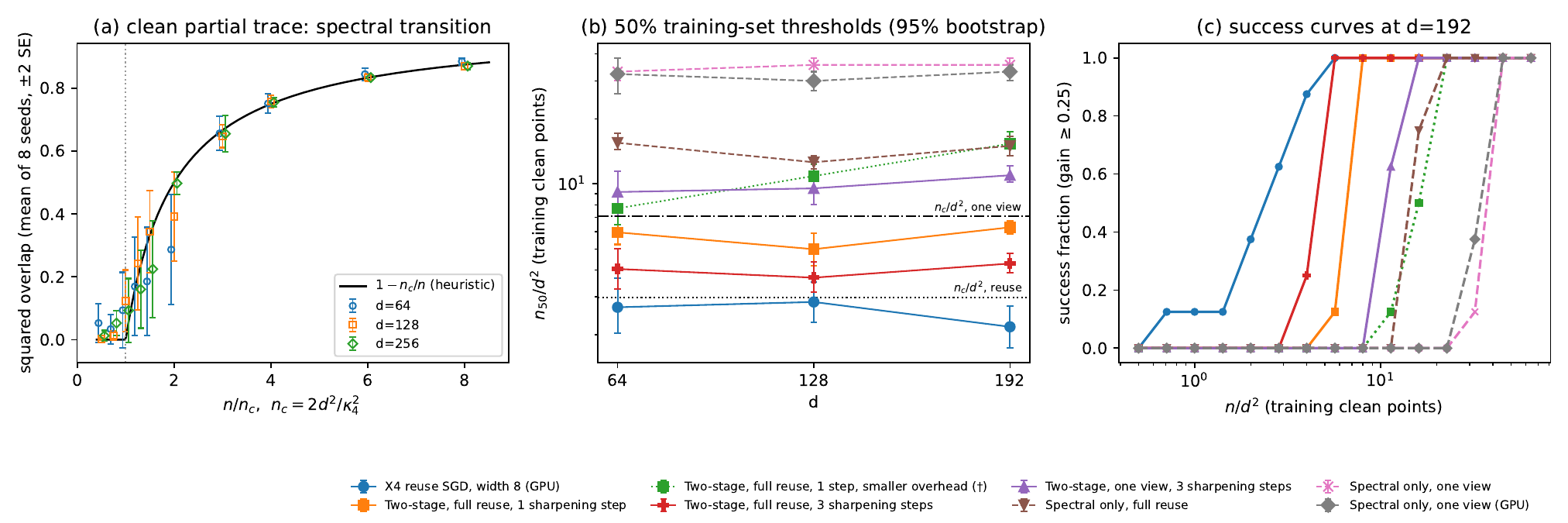}
\caption{Covariance-matched regime, \(k^\ast=4\). (a) Squared overlap of the
leading eigenvector of the clean partial trace against \(n/n_c\) with
\(n_c=2d^2/\kappa_4^2\): means of eight seeds with \(\pm2\) standard errors; the curve is
the heuristic prediction~\eqref{eq:bbp}. (b) Training-set thresholds \(n_{50}/d^2\) of
Table~\ref{tab:cm} with 95\% cell-bootstrap intervals; horizontal lines mark the reuse
and single-view transitions. (c) Success fractions against \(n/d^2\) (training points)
at \(d=192\).}
\label{fig:cm}
\end{figure}

\paragraph{The spectral transition.} For the single-view statistic of
Theorem~\ref{thm:partialtrace}, recovery curves collapse in units of
\(v_td^2/\bar\kappa_t^2\) for \(d\in\{32,\dots,256\}\) at \(t^\ast\), \(d\le128\) at \(t=0.6\)
and \(d\le64\) at \(t=1\), although raw sample sizes differ by a factor of 64 and
the constant is \(5.5\) and \(49\) times larger at \(t=0.6\) and \(t=1\): the squared
overlap is \(0.32\)--\(0.55\) at two units, \(0.64\)--\(0.77\) at four and
\(0.85\)--\(0.88\) at eight. For the full-reuse statistic, the clean partial trace,
Figure~\ref{fig:cm}(a) compares the squared overlap of its leading eigenvector at
\(n=x\,n_c\), \(n_c=2d^2/\kappa_4^2=\NcReuse\,d^2\), with the prediction \(1-1/x\) of
Remark~\ref{rem:bbp}. For \(x\ge3\) the largest deviation of the eight-seed mean
from \(1-1/x\) over \(d\in\{64,128,256\}\) is \(\BBPMaxErr\); the standard errors of these
means are at most \(\BBPMaxSE\), and no deviation exceeds \(\BBPMaxZ\) standard errors.
At these dimensions the data agree with the heuristic within Monte Carlo error.
The asymptotic prediction uses an unproved eigenvector-universality argument.
Below \(x=1\) the overlap is near zero, and between
\(1\) and \(3\) the finite-\(d\) transition is smoothed as expected.

\paragraph{Network sample scaling (X4).} The X4 study asks how many
distinct clean points a gradient-trained network needs. Its learner, frozen in
advance, is a width-8 network of units \(\He_3(w_j^\top x)\,v_j\) with normalized
first-layer weights, trained by minibatch SGD (batch 512, \(16d^2\) steps, learning
rate \(0.1\) or \(0.3\) selected on validation) that cycles over \(n\) clean points
with fresh noise. At 32 checkpoints a tied readout, one coefficient per unit, is
fitted on \(32{,}768\) calibration pairs and the checkpoint is selected by
readout-fitted loss on \(32{,}768\) validation pairs. A run succeeds if its test gain
is at least \(0.25\) of \(B\); \(n_{50}(d)\) is the 50\% point of a penalized
logistic fit over a 15-point grid \(n/d^2\in[0.5,64]\) with eight seeds, and the
pre-stated decision was H2 (\(n_{50}\propto d^2\)) if the slope of \(\log n_{50}\) on
\(\log d\) was at most \(2.4\) with upper 95\% limit below \(2.7\), and H3
(\(\propto d^3\)) if at least \(2.6\) with lower limit above \(2.3\). The
pre-registered heuristic favoured H3: small-step reuse descends a fixed landscape
whose correlation part is tensor PCA, for which local methods need \(d^3\)
\citep{bgj2020tensor}. The GPU study (1,080 cells, \(d\in\{64,128,192\}\)) returned
H2: slope \(1.84\) (the frozen protocol's seed bootstrap gives a 95\% interval
\(1.50\)--\(2.17\), the cell bootstrap below \(1.48\)--\(2.18\)), with no failure to
escape at the top of the grid. Over this factor-of-three range in dimension,
the threshold follows a \(d^2\)-like trend. The asymptotic exponent remains
unresolved.

\paragraph{What is counted.} Here \(n\) is the size of the training set. Each run
also uses \(32{,}768\) validation and \(32{,}768\) calibration pairs, drawn once for each
\((d,\text{seed})\) and shared by every \(n\) and both learning rates; under the
accounting of Section~\ref{sec:resources} these are clean points too. They amount to
\(16.0\), \(4.0\) and \(1.78\,d^2\) at \(d=64,128,192\) (Table~\ref{tab:cm}), so the
network's total budget at its threshold is \(\XTotA\), \(\XTotB\) and \(\XTotC\,d^2\): in
this range the fixed overhead dominates at small \(d\), and the total budget per \(d^2\)
falls with \(d\). All \(d^2\) statements of this section therefore concern the
training-set threshold at a fixed overhead, and all comparisons are between learners
given the same overhead and the same success rule. Asymptotically the overhead is
of lower order (Theorem~\ref{thm:twostage} needs \(n_2=O(d)\) and \(n_3=O(1)\)), but these
experiments do not reach that regime. Thresholds carry 95\% intervals from a cell
bootstrap: the eight outcomes of each \((d,n)\) cell are resampled, 2,000 times, and the
threshold fit is repeated. The cell bootstrap is exact for the explicit learners,
whose runs use independent randomness; for the GPU arm it ignores the coupling of
runs that share a seed across \(n\). Contrasts (Table~\ref{tab:contrasts}) are ratios of
independently resampled thresholds.

\paragraph{The explicit learners.} We ran the learner of
Theorem~\ref{thm:twostage} on the identical protocol and
overhead: stage~1 on the \(n\) training points, sharpening on the \(32{,}768\)
validation points, and the same tied readout on the \(32{,}768\) calibration points,
with the same success rule and threshold fit. Full reuse is computed directly as the
clean partial trace \(M_p(x_0)\) (Section~\ref{sec:training},
\emph{Presentations under reuse}), at the arithmetic cost of one view and with
no noisy presentations; the
sharpening step with full reuse uses the corresponding clean statistic. Table~\ref{tab:cm} and
Figure~\ref{fig:cm}(b,c) report the thresholds. Every learner run with the standard
overhead has slope between \(1.84\) and \(2.15\), each interval within \(1.48\)--\(2.39\). The
constants order as the theory predicts. Learners that use the clean statistic sit
within a factor of about two of the reuse transition \(n_c=\NcReuse\,d^2\), and the
ratio of one-view to full-reuse thresholds is \(2.27\)--\(2.59\) for the two-stage
learners and \(2.13\)--\(2.82\) for the spectral-only learners, against the predicted
inflation \(\iota_{t^\ast}=64/27\approx2.37\) of Theorem~\ref{thm:multiview}; five of
the six intervals contain it. The CPU replication of the single-view spectral learner
(\(33\)--\(36\,d^2\)) agrees with the frozen GPU arm (\(30\)--\(33\,d^2\)), with overlapping
intervals. Spectral-only learners need an overlap high enough for the readout to
clear the gain threshold, about \(1-n_c/n\ge0.8\), which multiplies their thresholds
by about five. One sharpening step, the learner analysed in
Theorem~\ref{thm:twostage}, lowers the full-reuse threshold from \(12.6\)--\(15.5\,d^2\)
to \(5.0\)--\(6.3\,d^2\), and three steps to \(3.7\)--\(4.3\,d^2\), as the theorem's
separation of geometry from accuracy predicts.

\paragraph{Sensitivity to the overhead.} Sharpening spends the validation set. With
\(8{,}192\) validation and \(8{,}192\) calibration points (rows marked \(\dagger\)), the
one-step learner needs \(7.7\), \(10.9\) and \(15.3\,d^2\) training points, \(1.3\)--\(2.4\)
times as many as with the standard overhead, and the ratio grows with \(d\) (slope
\(2.61\), interval \(2.43\)--\(2.80\)). This is the trade-off of the proof of
Theorem~\ref{thm:twostage}: the sharpening budget scales as \(d\) divided by a power of
the stage-1 overlap, so a sharpening set that shrinks relative to \(d\), from \(128d\) to
\(43d\), must be paid for with a better stage-1 direction and hence more training
points. At these dimensions the sample cost of the explicit learner is its total
budget \(n+n_2+n_3\), not \(n\) alone, and its \(d^2\) slope is a statement at a fixed,
ample overhead.

\paragraph{The trained network.} Under the common success rule and
validation overhead, the SGD-trained network needs \(2.2\)--\(2.8\,d^2\)
training points. Its thresholds are \(0.35\)--\(0.57\) of those of the
one-step two-stage learner, with bootstrap intervals excluding one at every
\(d\), and \(0.51\)--\(0.77\) of those of the three-step learner; the latter
interval reaches one at \(d=128\) (Table~\ref{tab:contrasts}). The network
also needs roughly an order of magnitude fewer training points than the
single-view spectral step.

For the learners tested here, the network thresholds
lie at \(0.73\)--\(0.95\) of \(n_c\), but \(n_c\) marks the overlap transition
of the clean partial-trace eigenvector, rather than a denoising-gain
threshold. The methods also use their fixed validation and calibration
sets differently: the network selects among 64 checkpoints and learning
rates and fits an eight-coefficient readout; the two-stage learner uses
these points to sharpen its direction. The network's sensitivity to a
smaller overhead has not been measured.

At its threshold, the network processes
\(16d^2\times512=8{,}192\,d^2\) noisy presentations, or
\(2{,}900\)--\(3{,}800\) views per clean point. Its clean-sample result is
therefore measured in a high-reuse regime. Conjecture~\ref{conj:sgd}
asks whether SGD forms the smoothed curvature of
Proposition~\ref{prop:smoothing}.

\begin{table}[t]
\centering\small
\caption{Fifty-percent \emph{training-set} thresholds \(n_{50}/d^2\) with 95\%
cell-bootstrap intervals, and slopes of \(\log n_{50}\) on \(\log d\) with intervals, all
on the frozen X4 protocol: eight seeds, gain threshold \(0.25\). Unless marked \(\dagger\),
every learner also uses \(32{,}768\) validation and \(32{,}768\) calibration clean points,
which the thresholds do not include; the overhead rows give these fixed costs in the
same units, to be added for the total clean-point budget. The \(\dagger\) row uses
\(8{,}192+8{,}192\) points. GPU rows are the frozen X4 study; the other rows are the
explicit learners of Section~\ref{sec:twostage} with full reuse or one view and with
or without the sharpening step. Reference transitions: \(n_c/d^2=\NcReuse\) (full
reuse) and \(\NcSingle\) (one view).}
\label{tab:cm}

\begin{tabular}{lcccc}
\toprule
Learner & \(d=64\) & \(d=128\) & \(d=192\) & slope \\
\midrule
X4 reuse SGD, width 8 (GPU) & \begin{tabular}[t]{@{}c@{}}2.68\\[-2pt]{\scriptsize[2.03, 3.66]}\end{tabular} & \begin{tabular}[t]{@{}c@{}}2.84\\[-2pt]{\scriptsize[2.29, 3.52]}\end{tabular} & \begin{tabular}[t]{@{}c@{}}2.18\\[-2pt]{\scriptsize[1.74, 2.71]}\end{tabular} & \begin{tabular}[t]{@{}c@{}}1.84\\[-2pt]{\scriptsize[1.48, 2.18]}\end{tabular} \\
Two-stage, full reuse, 1 sharpening step & \begin{tabular}[t]{@{}c@{}}5.96\\[-2pt]{\scriptsize[5.24, 7.07]}\end{tabular} & \begin{tabular}[t]{@{}c@{}}4.98\\[-2pt]{\scriptsize[4.18, 5.92]}\end{tabular} & \begin{tabular}[t]{@{}c@{}}6.29\\[-2pt]{\scriptsize[5.85, 6.73]}\end{tabular} & \begin{tabular}[t]{@{}c@{}}2.02\\[-2pt]{\scriptsize[1.85, 2.18]}\end{tabular} \\
Two-stage, full reuse, 1 step, smaller overhead$^\dagger$ & \begin{tabular}[t]{@{}c@{}}7.70\\[-2pt]{\scriptsize[6.49, 8.75]}\end{tabular} & \begin{tabular}[t]{@{}c@{}}10.87\\[-2pt]{\scriptsize[9.53, 12.34]}\end{tabular} & \begin{tabular}[t]{@{}c@{}}15.34\\[-2pt]{\scriptsize[13.46, 17.42]}\end{tabular} & \begin{tabular}[t]{@{}c@{}}2.61\\[-2pt]{\scriptsize[2.43, 2.80]}\end{tabular} \\
Two-stage, full reuse, 3 sharpening steps & \begin{tabular}[t]{@{}c@{}}4.04\\[-2pt]{\scriptsize[3.26, 5.01]}\end{tabular} & \begin{tabular}[t]{@{}c@{}}3.68\\[-2pt]{\scriptsize[3.15, 4.36]}\end{tabular} & \begin{tabular}[t]{@{}c@{}}4.28\\[-2pt]{\scriptsize[3.87, 4.76]}\end{tabular} & \begin{tabular}[t]{@{}c@{}}2.03\\[-2pt]{\scriptsize[1.83, 2.24]}\end{tabular} \\
Two-stage, one view, 3 sharpening steps & \begin{tabular}[t]{@{}c@{}}9.17\\[-2pt]{\scriptsize[7.08, 11.41]}\end{tabular} & \begin{tabular}[t]{@{}c@{}}9.53\\[-2pt]{\scriptsize[8.01, 11.33]}\end{tabular} & \begin{tabular}[t]{@{}c@{}}10.96\\[-2pt]{\scriptsize[10.19, 12.09]}\end{tabular} & \begin{tabular}[t]{@{}c@{}}2.15\\[-2pt]{\scriptsize[1.91, 2.39]}\end{tabular} \\
Spectral only, full reuse & \begin{tabular}[t]{@{}c@{}}15.49\\[-2pt]{\scriptsize[14.43, 17.11]}\end{tabular} & \begin{tabular}[t]{@{}c@{}}12.59\\[-2pt]{\scriptsize[11.72, 13.47]}\end{tabular} & \begin{tabular}[t]{@{}c@{}}14.97\\[-2pt]{\scriptsize[13.46, 16.53]}\end{tabular} & \begin{tabular}[t]{@{}c@{}}1.94\\[-2pt]{\scriptsize[1.81, 2.06]}\end{tabular} \\
Spectral only, one view & \begin{tabular}[t]{@{}c@{}}33.07\\[-2pt]{\scriptsize[29.97, 38.11]}\end{tabular} & \begin{tabular}[t]{@{}c@{}}35.54\\[-2pt]{\scriptsize[33.06, 38.07]}\end{tabular} & \begin{tabular}[t]{@{}c@{}}35.53\\[-2pt]{\scriptsize[33.06, 38.06]}\end{tabular} & \begin{tabular}[t]{@{}c@{}}2.07\\[-2pt]{\scriptsize[1.96, 2.19]}\end{tabular} \\
Spectral only, one view (GPU) & \begin{tabular}[t]{@{}c@{}}32.24\\[-2pt]{\scriptsize[26.03, 38.28]}\end{tabular} & \begin{tabular}[t]{@{}c@{}}29.95\\[-2pt]{\scriptsize[26.93, 33.06]}\end{tabular} & \begin{tabular}[t]{@{}c@{}}33.06\\[-2pt]{\scriptsize[29.95, 35.60]}\end{tabular} & \begin{tabular}[t]{@{}c@{}}2.01\\[-2pt]{\scriptsize[1.82, 2.22]}\end{tabular} \\
\midrule
Overhead: validation + calibration, \(65{,}536\) points & 16.00 & 4.00 & 1.78 & \\
Overhead$^\dagger$: \(16{,}384\) points & 4.00 & 1.00 & 0.44 & \\
\bottomrule
\end{tabular}

\end{table}

\begin{table}[t]
\centering\small
\caption{Contrasts between training-set thresholds: ratios of \(n_{50}\) with 95\%
intervals from independent cell-bootstrap draws (unpaired). Learners in each ratio
share the overhead, except in the last row, which compares the one-step learner
with \(8{,}192+8{,}192\) and with \(32{,}768+32{,}768\) validation and calibration points.}
\label{tab:contrasts}

\begin{tabular}{lccc}
\toprule
Ratio of training-set thresholds & \(d=64\) & \(d=128\) & \(d=192\) \\
\midrule
X4 network / two-stage (1 step) & 0.45 \,[0.32, 0.64] & 0.57 \,[0.44, 0.74] & 0.35 \,[0.27, 0.45] \\
X4 network / two-stage (3 steps) & 0.66 \,[0.47, 0.95] & 0.77 \,[0.60, 1.00] & 0.51 \,[0.39, 0.64] \\
one view / full reuse, two-stage (3 steps) & 2.27 \,[1.68, 3.07] & 2.59 \,[2.08, 3.22] & 2.56 \,[2.22, 2.92] \\
one view / full reuse, spectral only & 2.13 \,[1.87, 2.46] & 2.82 \,[2.54, 3.25] & 2.37 \,[2.13, 2.83] \\
smaller / standard overhead, two-stage (1 step) & 1.29 \,[1.04, 1.59] & 2.18 \,[1.75, 2.71] & 2.44 \,[2.00, 2.77] \\
\bottomrule
\end{tabular}

\end{table}

\paragraph{The earlier X4 screens.} Before the two-stage protocol, an uncalibrated
version that used the network's own output layer passed the gain threshold in
\(2/3\) seeds at \(d=64\) but in \(0/3\) at \(d=96\) and \(128\). A diagnostic that refitted a
tied readout on independent pairs raised the \(d=128\) gains to \(0.71\) in \(3/3\)
seeds: the first layers had aligned, and the output layer was the bottleneck. This
motivated the two-stage protocol, frozen before the rate study.

\subsection{Additional tests}\label{sec:exp-failures}

\paragraph{Diffusion time in fixed-step SGD (X1).} We trained a single Hermite-3
unit by spherical online SGD with step \(0.1/d\), using sixteen paired seeds
per \((d,t)\) over seven times. The pre-stated prediction was that the escape time scales
as \(S_t^{-2}\): fastest time in \([0.10,0.20]\), and a \(t=0.6\) to \(t^\ast\) ratio in
\([3.5,8]\). The fastest time was in the interval at \(d=32\) and \(d=64\), but the
ratio was \(7.77\) at \(d=32\) and \(32.37\) at \(d=64\) (bootstrap interval
\(10.2\)--\(75.9\)), which falsifies the ratio prediction for this optimizer. The fixed
step falls outside the small-step regime of Corollary~\ref{cor:online}.
Theorem~\ref{thm:time} concerns a static signal-to-noise ratio, while
Theorem~\ref{thm:multiview}(5) gives the time result for the analysed
spectral learner.

\paragraph{Residual exponent for fixed-step SGD (X2).} In the covariance-visible
model, we ran paired spherical SGD with the full target and with the
fitted-Gaussian residual. The full arm escaped after about \(d\log d\) pairs, as
predicted. The residual arm was predicted to need \(d^3\); its fitted exponent over
\(d\in\{16,\dots,128\}\) was \(1.57\) (interval \(1.37\)--\(2.26\)), below the cutoff
\(2.5\). Corollary~\ref{cor:shift} concerns the degree of the first nonzero
population correlation, not a finite-step escape exponent, and the residual arm
was calibrated from a clean point that reveals the span exactly, so the experiment
tests the gradient signal of the residual parametrization rather than a
best-algorithm sample complexity.

\paragraph{Gradient descent on the unsmoothed reuse landscape.} Deterministic
gradient descent on the noise-averaged single-unit landscape, with the output
coefficient profiled out, from eight random starts on each of three data sets per
\((d,n)\), did not reach 50\% escape by \(n=64d^2\) at any \(d\in\{32,64,128\}\)
(\(11/24\), \(5/24\) and \(10/24\) starts escaped at the top of the grid), while the
single-view spectral statistic on the same clean points reached squared overlap
\(0.88\)--\(0.89\). This is consistent with Proposition~\ref{prop:smoothing}, which
shows that the unsmoothed landscape is flat to third order at the origin,
and with the \(d^3\) threshold of local methods for tensor PCA
\citep{bgj2020tensor}.
\section{Discussion}\label{sec:discussion}

The gradient of the denoising loss exposes moment differences
between the data and a baseline
(Theorem~\ref{thm:bijection}). A polynomial field of degree below
\(k^\ast-1\) cannot use the first unmatched moment
(Corollary~\ref{cor:polybarrier}). Under the hypotheses of
Theorems~\ref{thm:frozen} and~\ref{thm:kernel} and
Corollary~\ref{cor:online}, frozen dictionaries, kernel ridge regression and
small-step online training of one unit pay \(d^{k^\ast-1}\) in their respective
resources. In the covariance-visible regime the frozen threshold is exactly
at \(d^3/6\); for one hidden direction, its first-step height is the Gaussian
control (Theorems~\ref{thm:sharp} and~\ref{thm:height}).

The two-stage learner forms the curvature of a smoothed denoising landscape
and averages over noisy views of each clean point. It needs
\(O(d^{k^\ast/2}\log d)\) clean points (Theorem~\ref{thm:twostage}).
Proposition~\ref{prop:sq} gives a statistical-query tolerance bound of
\(d^{-k^\ast/4}\), but does not prove a matching sample lower bound for
gradient learners. When covariance reveals the subspace, a fitted Gaussian
baseline and principal components give a projected estimator whose risk law
is independent of ambient dimension (Corollary~\ref{cor:shift},
Theorem~\ref{thm:invariance}).

Averaging fresh noisy views of one clean point removes the forward-process
variance in the analysed learner. The learner uses more presentations of
each point without requiring more clean points
(Lemma~\ref{lem:condstein}, Theorem~\ref{thm:multiview}). For one view per
point, the leading spectral constant is minimized at \(t^\ast\); full reuse
removes this time-dependent inflation. Once the learner has found the
direction, fitting a scalar readout on independent pairs turns alignment into
a score gain at dimension-free sample cost (Theorem~\ref{thm:twostage}).
In the X4 study, output-layer calibration affected the measured gain even
when the first layer was aligned.

The two-stage theorem covers one hidden direction and even \(k^\ast\).
Its learner uses strong smoothing, equivalently a matrix parameter, and
independent data splits. The clean-sample bound includes a logarithm and,
for \(k^\ast>4\), a non-sharp constant. The proof does not cover standard
network training.

The covariance-matched frozen-feature theorem assumes an \(L^4/L^2\)
constant that has not been established for random-feature spans. The
kernel theorem assumes a high-degree Taylor coefficient. The exact-height
result covers the first staircase step for one hidden direction. PCA
invariance uses noiseless training points; off-subspace degradation is
measured here but has no bound in the current analysis. We do not derive
a lower bound on generated-sample error. Spectral experiments end
at \(d=256\), and the trained-network study ends at \(d=192\).

\clearpage
\appendix
\etocdepthtag.toc{mtappendix}
\phantomsection
\section*{Guide to the appendices}\label{app:guide}
\addcontentsline{toc}{section}{Guide to the appendices}

The appendices are organized by proof technique. This guide lists the
results, their dependencies and the notation used in the proofs. Roadmaps at
the start of each appendix and long proof indicate the order of the
arguments.

\begingroup
\etocsettagdepth{mtmain}{none}
\etocsettagdepth{mtappendix}{subsection}
\etocsettocstyle{\subsection*{Contents of the appendices}\small}{}
\tableofcontents
\endgroup

\subsection*{How the results depend on each other}
Two lemmas carry most of the weight. Lemma~\ref{lem:spectral} (an
orientation-averaged Ky Fan bound) drives
every frozen-feature lower bound in both regimes, and Lemma~\ref{lem:condstein}
(the conditional Stein identity) drives every positive training result. The
one-sided change of measure (Lemma~\ref{lem:onesided}) is the bridge that turns the
covariance-visible argument into the covariance-matched one.

\subsection*{Where each result is stated and proved}
\newcommand{\resrow}[4]{#1~\ref{#2} & #3 & \pageref{#2} & \ref{pf:#2}, p.~\pageref{pf:#2} & #4\\}
\begingroup\small
\setlength{\tabcolsep}{4pt}
\begin{longtable}{p{1.7cm}p{3.6cm}p{0.8cm}p{2.0cm}p{6.2cm}}
\toprule
Result & Content & Page & Proof & Main tools \\
\midrule
\endfirsthead
\toprule
Result & Content & Page & Proof & Main tools \\
\midrule
\endhead
\bottomrule
\endfoot
\multicolumn{5}{l}{\emph{Appendix~\ref{app:moments}: moment calculus (Section~\ref{sec:moments})}}\\
\resrow{Thm}{thm:stein}{Stein--Tweedie identity}{Gaussian integration by parts; Tweedie's formula}
\resrow{Thm}{thm:bijection}{polynomial fields \(\Leftrightarrow\) moments}{Ornstein--Uhlenbeck generator in the Hermite basis}
\resrow{Lem}{lem:condstein}{conditional Stein identity}{Stein's lemma at fixed \(x_0\); Mehler's formula; Hermite eigen-equation}
\resrow{Thm}{thm:baseline}{baseline hides matched moments}{binomial expansion of moment tensors; Theorem~\ref{thm:bijection}}
\resrow{Cor}{cor:polybarrier}{exact polynomial barrier}{tower property; Theorem~\ref{thm:baseline}}
\resrow{Cor}{cor:shift}{fitted Gaussian shifts the exponent}{symmetry of the latent law; fourth cumulant}
\resrow{Thm}{thm:noiseop}{diffusion as a noise operator}{Mehler's formula; Parseval}
\resrow{Thm}{thm:time}{optimal geometry time}{Hermite expansion of the test field; Cauchy--Schwarz; one-variable calculus}
\midrule
\multicolumn{5}{l}{\emph{Appendix~\ref{app:proofs}: covariance-visible regime and one-sided change of measure (Section~\ref{sec:frozen})}}\\
\resrow{Lem}{lem:reduction}{denoiser reduction}{conditional independence of the normal coordinates; Tweedie}
\resrow{Lem}{lem:posterior}{posterior mean of a mixture coordinate}{Gaussian conditioning; Bayes' rule}
\resrow{Lem}{lem:density}{density floor \(c_t\)}{symmetry of \(Z\); \(\cosh\ge1\)}
\resrow{Lem}{lem:spectral}{orientation-averaged Ky Fan bound}{Funk--Hecke; Rodrigues' formula; Ky Fan's maximum principle}
\resrow{Thm}{thm:floor}{cubic barrier, orientation average}{Lemmas~\ref{lem:reduction}, \ref{lem:density}, \ref{lem:spectral}}
\resrow{Cor}{cor:ambient}{geometry-blind random features}{the dictionary is a function of rotation-invariant data}
\resrow{Prop}{prop:fixed}{every orientation for isotropic features}{rotation equivariance; transitivity on the Stiefel manifold}
\resrow{Thm}{thm:typical}{typical orientations}{Lipschitz floor; log-Sobolev concentration on the sphere}
\resrow{Thm}{thm:sharp}{matching polynomial dictionaries}{orthogonal projection in \(L^2(q_t)\); Hermite completeness}
\resrow{Cor}{cor:staircase}{polynomial staircase}{Theorems~\ref{thm:floor} and~\ref{thm:sharp}; dominated convergence}
\resrow{Cor}{cor:price}{cubic feature price}{rearranging Theorems~\ref{thm:floor} and~\ref{thm:typical}}
\resrow{Lem}{lem:onesided}{one-sided change of measure}{Rayleigh quotient; change of measure in the cross term only}
\resrow{Thm}{thm:height}{exact height of the first step}{Lemmas~\ref{lem:onesided} and~\ref{lem:spectral} applied to \(\psi_t=\rho_t\hat g_t\)}
\resrow{Prop}{prop:offsubspace}{data off the subspace}{exact time change and dilation}
\resrow{Lem}{lem:monotone}{monotonicity in \(d\)}{explicit difference of \(N_{d,3}\)}
\resrow{Thm}{thm:invariance}{PCA dimension invariance}{exact recovery; projector dependence; Gaussian rotation invariance}
\resrow{Thm}{thm:generation}{generation invariance}{closed latent recursion of the sampler}
\resrow{Cor}{cor:separation}{checkable separation}{Theorems~\ref{thm:floor}, \ref{thm:invariance}; Lemma~\ref{lem:monotone}}
\midrule
\multicolumn{5}{l}{\emph{Appendix~\ref{app:cm}: covariance-matched frozen and kernel barriers (Sections~\ref{sec:frozen-cm} and~\ref{sec:kernels})}}\\
\resrow{Lem}{lem:comparison}{norm comparison}{hypercontractivity; truncation of the hidden coordinate}
\resrow{Thm}{thm:frozen}{sharp frozen exponent \(d^{k^\ast-1}\)}{Lemmas~\ref{lem:onesided}, \ref{lem:comparison}, \ref{lem:spectral}}
\resrow{Lem}{lem:leakage}{kernel leakage}{Mehler's formula; integration by parts}
\resrow{Thm}{thm:kernel}{kernels pay \(d^{k^\ast-1}\)}{cross-term bound; Schur product theorem}
\resrow{Lem}{lem:selfreg}{self-regularization}{positive-semidefinite binomial decomposition; Gershgorin; Gaussian concentration}
\midrule
\multicolumn{5}{l}{\emph{Appendix~\ref{app:training}: gradient learners (Section~\ref{sec:training})}}\\
\resrow{Prop}{prop:neuron}{one unit sees an overlap polynomial}{Theorem~\ref{thm:stein}; Mehler's formula}
\resrow{Cor}{cor:online}{online SGD pays \(d^{k^\ast-1}\)}{verification of the hypotheses of \citet{bgj2021}}
\resrow{Prop}{prop:sq}{statistical-query lower bound}{pairwise correlations; fooling argument; random packing}
\resrow{Thm}{thm:partialtrace}{one gradient step, \(k^\ast=4\)}{Theorem~\ref{thm:bijection}; chi-square moments; matrix Bernstein; Davis--Kahan}
\resrow{Thm}{thm:multiview}{reused views, every even \(k^\ast\)}{Lemma~\ref{lem:condstein}; pairing count; hypercontractivity; matrix Bernstein}
\resrow{Prop}{prop:smoothing}{smoothing exposes the partial trace}{Gaussian moments of the smoothing variable; homogeneity}
\resrow{Thm}{thm:twostage}{two-stage learner}{Theorem~\ref{thm:multiview}; Lemma~\ref{lem:condstein}(2); Chebyshev}
\resrow{Cor}{cor:dimsep}{dimension separation, \(k^\ast=4\)}{combination of the four lower bounds and Theorem~\ref{thm:twostage}}
\end{longtable}
\endgroup

\subsection*{Notation}
Symbols are listed in order of first use. Scores are always \(s_t\) (and
\(\widehat s\) for estimates); the coordinate scale of the covariance-matched regime
is \(\bar\sigma_t\). A few symbols (\(p\), \(B\), \(\rho\)) have a second meaning in one
part of the paper, as stated in the table.
\begingroup\small
\setlength{\tabcolsep}{4pt}
\begin{longtable}{p{2.9cm}p{9.6cm}p{2.0cm}}
\toprule
Symbol & Meaning & Defined \\
\midrule
\endfirsthead
\toprule
Symbol & Meaning & Defined \\
\midrule
\endhead
\bottomrule
\endfoot
\(a_t,\;h_t\) & \(e^{-t}\) and \(1-e^{-2t}\), signal and noise variances of the forward process & §\ref{sec:pairs}\\
\(X_0,\;X_t,\;P_t,\;p_t\) & clean point, noisy view \(a_tX_0+\sqrt{h_t}G\), law and density of \(X_t\) & §\ref{sec:pairs}\\
\(m_t\) & denoiser \(\E[X_0\mid X_t=\cdot]\) & §\ref{sec:pairs}\\
\(s_t,\;\widehat s\) & score \(\nabla\log p_t\) and a score estimate \((a_t\widehat m-x)/h_t\). In App.~\ref{app:proofs} and Lemma~\ref{lem:posterior}, \(s\) without index is the mixture-component scale \((a_t^2\tau^2+h_t)^{1/2}\) & §\ref{sec:pairs}\\
\(\bar\sigma_t\) & coordinate scale \((a_t^2\sigma_0^2+h_t)^{1/2}\) of \(X_t\) in the covariance-matched regime (\(\bar\sigma\) when \(t\) is fixed) & Ass.~\ref{ass:cm}\\
\(Q_0,\;Q_t,\;q_t,\;m^Q_t\) & baseline law, its diffused law and density, its denoiser & §\ref{sec:pairs}\\
\(\Stein_{Q_t}\) & Langevin--Stein operator \(\dvg\varphi+\inner{\nabla\log q_t}{\varphi}\) & §\ref{sec:pairs}\\
\(\He_j,\;h_j,\;H_\alpha\) & probabilists' Hermite polynomials, their orthonormal versions, multivariate products & §\ref{sec:pairs}\\
\(\He_j(x)\), \(M_p\), \(\mathcal H\) & Hermite tensor of \(x\in\R^d\); its partial trace over \(p\) index pairs; \(\mathcal H=M_1\) & §\ref{sec:pairs}\\
\(U,\;u,\;k,\;Z,\;\nu\) & hidden orientation (\(u\) when \(k=1\)), its rank, latent vector and its law & Ass.~\ref{ass:hidden}\\
\(\sigma_0\) & scale of the Gaussian component off the subspace (covariance-matched); scale of added isotropic noise in Proposition~\ref{prop:offsubspace} & Ass.~\ref{ass:hidden}\\
\(\mu,\;\tau\) & mixture parameters: \(Z=\epsilon\mu+\tau\xi\) & Ass.~\ref{ass:model}\\
\(k^\ast\) & generative exponent: first moment order at which \(\nu\) differs from \(N(0,\sigma_0^2\Id)\) & Ass.~\ref{ass:cm}\\
\(\lambda_t\) & \(a_t\sigma_0/\bar\sigma_t\in(0,1)\), contraction of the hidden coordinate by diffusion & Ass.~\ref{ass:cm}\\
\(\tilde\kappa\) & normalized \(k^\ast\)-th cumulant \(\kappa_{k^\ast}(\nu)/\sigma_0^{k^\ast}\) (\(\kappa_4\) for \(k^\ast=4\)) & Ass.~\ref{ass:cm}\\
\(B\) & excess denoiser risk \(\|m_t-m^Q_t\|^2_{L^2(P_t)}\) of the Gaussian baseline (covariance-matched); in Theorems~\ref{thm:partialtrace} and~\ref{thm:multiview} also a symmetric matrix parameter & §\ref{sec:resources}\\
\(M,\;p,\;L\) & dictionary size; random-feature width (§\ref{sec:model}); number of bandwidths. In the covariance-matched training results \(p\) is instead defined by \(k^\ast=2p+2\) & §\ref{sec:resources}\\
\(e_j(t),\;\chi^2\) & Hermite moments \(\E h_j(u^\top X_t/\bar\sigma_t)\) of the hidden coordinate; chi-square divergence \(\chi^2_0\) & Thm.~\ref{thm:noiseop}\\
\(S_t,\;t^\ast\) & signal-to-noise ratio of the best low-degree correlation; its maximizer & Thm.~\ref{thm:time}\\
\(g_t\) & posterior mean of a latent coordinate (covariance-visible) & Lem.~\ref{lem:posterior}\\
\(c_j(t),\;\kappa_t,\;\kappa_{J,t},\;J_t\) & Hermite coefficients of \(g_t\), its energy beyond degree one, beyond degree \(J\), and its derivative energy (covariance-visible only) & §\ref{sec:reductions}\\
\(c_t\) & density floor \(\min\rho_t\) & Lem.~\ref{lem:density}\\
\(N_{d,\ell},\;D_{d,J}\) & dimension of degree-\(\ell\) spherical harmonics; of polynomials of degree \(\le J\) & §\ref{sec:floor}, Thm.~\ref{thm:sharp}\\
\(w_{j,J}(d),\;\eta_{j,\ell}\) & harmonic weights and Funk--Hecke coefficients & Lem.~\ref{lem:spectral}\\
\(F_d\), \(F^{(J)}_d\), \(F^{\mathrm{KF}}_d\), \(F_\infty\), \(B_d\) & explicit, level-\(J\), Ky Fan and limiting floors; normalizer of Corollary~\ref{cor:price} & Thm.~\ref{thm:floor}\\
\(\varepsilon_J(t),\;R_G\) & best degree-\(J\) polynomial error of \(g_t\); score risk of the Gaussian control & Thm.~\ref{thm:sharp}, §\ref{sec:height}\\
\(\rho\), \(\rho_t\), \(\psi_t\), \(\Psi_t\), \(\alpha_t\) & density ratio to the Gaussian reference; weighted residual \(\rho_t\hat g_t\) and its energy; linear coefficient of \(g_t\). In Section~\ref{sec:training} and App.~\ref{app:training}, \(\rho^2\) is instead the null residual variance \(\sigma_0^2h_t/\bar\sigma_t^2\) & Lem.~\ref{lem:density}, §\ref{sec:height}\\
\(r_t,\;\hat\psi,\;\Psi,\;c_\kappa,\;G\) & covariance-matched density ratio, weighted residual and its energy, comparison constant, gain of the degree-\(J\) dictionary & Thm.~\ref{thm:frozen}\\
\(\tau_u(y)\) & leakage of the hidden residual onto the kernel section at \(y\) & Lem.~\ref{lem:leakage}\\
\(\bar\kappa_t,\;v_t\) & spike \(\lambda_t^4\tilde\kappa\) and single-view variance constant \(a_t^2\sigma_0^2/(2h_t)\) for \(k^\ast=4\) & Thm.~\ref{thm:partialtrace}\\
\(X,\;Y_i,\;\bar Y_n,\;r\) & statistic of one view; average over the \(r\) views of clean point \(i\); its mean over \(n\) points; number of views & Thm.~\ref{thm:multiview}\\
\(c_p,\;\iota_t\) & null variance constant \(2^pp!\); single-view noise inflation & Thm.~\ref{thm:multiview}\\
\(n_c\) & predicted spectral transition \((1+(\iota_t-1)/r)c_pd^{k^\ast/2}/\tilde\kappa^2\) & Rem.~\ref{rem:bbp}\\
\(\widehat C_r,\;\widehat C_{r,\sigma},\;\widehat T\) & reuse correlation of one unit; its Gaussian smoothing at scale \(\sigma\); empirical clean Hermite tensor & Prop.~\ref{prop:smoothing}\\
\(n_1,n_2,n_3,\;G_\ast\) & sample sizes of the three stages; gain of the best tied unit & §\ref{sec:twostage}\\
\(C,\;C',\;C_i,\;K_i\) & constants that depend only on the quantities stated in each result, never on \(d\) & ---\\
\end{longtable}
\endgroup
\section{Proofs for Section~\ref{sec:moments}}\label{app:moments}

\begin{roadmap}[: how Appendix~\ref{app:moments} is organized]
Every result here is an exact identity, proved with three tools only: Gaussian
integration by parts (Stein's lemma), Mehler's formula
\(\E\,h_j(\lambda y+\sqrt{1-\lambda^2}\,\xi)=\lambda^jh_j(y)\), and the spectral
decomposition of the Ornstein--Uhlenbeck generator in the Hermite basis. The order
is: the unconditional identity (\ref{pf:thm:stein}), its polynomial inverse
(\ref{pf:thm:bijection}), the conditional version used by every training result
(\ref{pf:lem:condstein}), the moment-matching consequences
(\ref{pf:thm:baseline}), and the action of diffusion time
(\ref{pf:thm:noiseop}--\ref{pf:thm:time}).
\end{roadmap}

\subsection{Proof of Theorem~\ref{thm:stein} (Stein--Tweedie identity)}\label{pf:thm:stein}
Condition on \(X_0=x_0\). Gaussian integration by parts for
\(g\mapsto\varphi(ax_0+\sqrt hg)\) gives
\(\E[\inner G{\varphi(X_t)}\mid X_0]=\sqrt h\,\E[\dvg\varphi(X_t)\mid X_0]\); the
growth conditions and the moments of \(X_0\) make all terms integrable. Since
\(X_0=(X_t-\sqrt hG)/a\),
\(\E\inner{X_0}{\varphi(X_t)}=a^{-1}(\E\inner{X_t}{\varphi(X_t)}-h\,\E\dvg\varphi(X_t))\).
Tweedie's formula for \(Q\) gives
\(\E\inner{m^Q_t(X_t)}{\varphi(X_t)}=a^{-1}(\E\inner{X_t}{\varphi}+h\,\E\inner{\nabla\log q_t}{\varphi})\).
Subtracting yields~\eqref{eq:stein}. \qed

\subsection{Proof of Theorem~\ref{thm:bijection} (polynomial fields \(\Leftrightarrow\) moments)}\label{pf:thm:bijection}
An orthogonal change of coordinates and a translation map gradient fields to
gradient fields and preserve degrees, so we may take \(\bar m=0\) and
\(C=\mathrm{diag}(c_1,\dots,c_d)\). (1)
\(\Stein_{Q_t}\varphi=\dvg\varphi-\inner{C^{-1}x}\varphi\) has degree at most \(D\),
and \(\E_{Q_t}\Stein_{Q_t}\varphi=0\) by Stein's identity for \(Q_t\). (2)
\(\Stein_{Q_t}\nabla u=\sum_i(\partial_i^2u-c_i^{-1}x_i\partial_iu)=Lu\), the
generator of the Ornstein--Uhlenbeck process with invariant law \(Q_t\). The
products \(H_\alpha(x)=\prod_ih_{\alpha_i}(x_i/\sqrt{c_i})\) are orthonormal in
\(L^2(Q_t)\) and \(LH_\alpha=-(\sum_i\alpha_i/c_i)H_\alpha\), so \(L\) maps
\(\mathcal P_D^0\) to itself diagonally with nonzero eigenvalues
\(\Lambda_\alpha=\sum_i\alpha_i/c_i\). Its inverse is
\(q=\sum_{\alpha\ne0}q_\alpha H_\alpha\mapsto u_q=-\sum_{\alpha\ne0}q_\alpha H_\alpha/\Lambda_\alpha\).
Theorem~\ref{thm:stein} with \(\varphi=\nabla u_q\) gives~\eqref{eq:sqcsq} for
\(q\in\mathcal P_D^0\), and constants contribute nothing to either side. (3)
\(\|\nabla u\|^2_{L^2(Q_t)}=\E_{Q_t}[u(-Lu)]=\sum_{\alpha\ne0}q_\alpha^2/\Lambda_\alpha\),
and \(1/\lambda_{\max}(C)\le\Lambda_\alpha\le D/\lambda_{\min}(C)\) for
\(1\le|\alpha|\le D\). \qed

\subsection{Proof of Lemma~\ref{lem:condstein} (conditional Stein identity)}\label{pf:lem:condstein}
\step{Common step: Stein's lemma at a fixed clean point} Fix \(x_0\), let \(G\sim N(0,\Id)\), \(X_t=ax_0+\sqrt hG\), and write
\(\mathsf Pf(x_0)=\E f(ax_0+\sqrt hG)\). Since \(m^Q_t(x)=cx\) with
\(c=a\sigma_0^2/\bar\sigma^2\) and \(1-ca=h/\bar\sigma^2\), Gaussian integration by parts,
\(\E\inner G{\varphi(X_t)}=\sqrt h\,\E\dvg\varphi(X_t)\), gives for every polynomial
field \(\varphi\)
\[
 \E\inner{x_0-cX_t}{\varphi(X_t)}
 =\frac h{\bar\sigma^2}\inner{x_0}{\mathsf P\varphi(x_0)}-ch\,\mathsf P(\dvg\varphi)(x_0).
\]
\step{Part 1: gradient fields} Take \(\varphi=\nabla u\). Since \(\mathsf P\nabla u=a^{-1}\nabla\mathsf Pu\) and
\(\mathsf P\Delta u=a^{-2}\Delta\mathsf Pu\), the right side is
\((h/(a\bar\sigma^2))[\inner{x_0}{\nabla\mathsf Pu}-\sigma_0^2\Delta\mathsf Pu](x_0)\).
Coordinatewise \((ax_0+\sqrt hG)/\bar\sigma=\lambda(x_0/\sigma_0)+\sqrt{1-\lambda^2}\,G\),
because \(\lambda^2+h/\bar\sigma^2=1\), so Mehler's formula gives \(\mathsf Pu=\lambda^j\tilde u\).
Finally \(\inner x{\nabla f}-\sigma_0^2\Delta f=jf\) for every \(f\) in the degree-\(j\)
chaos of \(N(0,\sigma_0^2\Id)\), which contains \(\tilde u\): with \(y=x/\sigma_0\) this is
the Hermite eigen-equation \(\inner y{\nabla_y}-\Delta_y=j\).
\step{Part 2: output-layer fields} Take
\(\varphi=\He_{j-1}(w^\top x/\bar\sigma)\,e\) for a fixed vector \(e\) and put
\(z=w^\top x_0/\sigma_0\). By Mehler's formula
\(\mathsf P\varphi(x_0)=\lambda^{j-1}\He_{j-1}(z)\,e\) and
\(\mathsf P(\dvg\varphi)=(j-1)\lambda^{j-2}\He_{j-2}(z)\,w^\top e/\bar\sigma\), with the convention
that this term is zero for \(j=1\), where \(\varphi\) is constant. Since
\(ch/\bar\sigma=(h/\bar\sigma^2)\sigma_0\lambda\), the right side equals
\((h\sigma_0\lambda^{j-1}/\bar\sigma^2)[(e^\top x_0/\sigma_0)\He_{j-1}(z)-(j-1)(w^\top e)\He_{j-2}(z)]\),
which is \((h\sigma_0\lambda^{j-1}/\bar\sigma^2)\He_j(x_0/\sigma_0)[w^{\otimes(j-1)},e]\) by the
recursion for Hermite tensors at a unit vector. \qed

\subsection{Proofs of Theorem~\ref{thm:baseline} and its corollaries (baseline hides matched moments and its consequences)}\label{pf:thm:baseline}\label{pf:cor:polybarrier}\label{pf:cor:shift}
Since \(G\) is independent of \(X_0\),
\(\E(aX_0+\sqrt hG)^{\otimes j}=\sum_{i\le j}\binom ji a^i
\mathrm{Sym}(\E X_0^{\otimes i}\otimes\E(\sqrt hG)^{\otimes(j-i)})\), with the same
coefficients for \(Q\). Hence the moments of \(P_t\) and \(Q_t\) agree below order
\(k\), and their difference at order \(k\) is the \(i=k\) term. Part~1 follows from
Theorem~\ref{thm:stein} and Theorem~\ref{thm:bijection}(1), since
\(\Stein_{Q_t}\varphi\) is a polynomial of degree at most \(k-1\) on which \(P_t\) and
\(Q_t\) agree. For part~2 choose a degree-\(k\) polynomial \(q\) with
\(\E_{P_t}q\ne\E_{Q_t}q\) and take \(\varphi=\nabla u_q\).

For Corollary~\ref{cor:polybarrier},
\(\inner{m_t-m^Q_t}{\varphi}_{L^2(P_t)}=\E\inner{X_0-m^Q_t(X_t)}{\varphi(X_t)}=0\)
by the tower property and Theorem~\ref{thm:baseline}, so every polynomial field of
degree at most \(k^\ast-2\) is orthogonal to the residual, and the score residual is
\((a/h)(m_t-m^Q_t)\).

For Corollary~\ref{cor:shift}: against the isotropic Gaussian with the data's
trace, the second moments differ by a multiple of \(UU^\top\) minus a multiple of
\(\Id\), so linear fields correlate with the residual through \(U\). The fitted
Gaussian matches the first two moments, so by Theorem~\ref{thm:baseline} fields of
degree at most one have zero correlation. Central symmetry makes the third
moments match (both vanish), so fields of degree two have zero correlation. The
fourth moment tensor of the data minus that of the fitted Gaussian is the
fourth cumulant tensor, \(\sum_j\kappa_4(Z_j)u_j^{\otimes4}\) for independent
coordinates, which is nonzero; Theorem~\ref{thm:baseline}(2) gives a field of
degree three with nonzero correlation. \qed

\subsection{Proof of Theorem~\ref{thm:noiseop} (diffusion as a noise operator)}\label{pf:thm:noiseop}
\(Y_t/\bar\sigma_t=\lambda_t(Z/\sigma_0)+\sqrt{1-\lambda_t^2}\,\xi\) with \(\xi\sim N(0,1)\)
independent, since \(\lambda_t^2+h/\bar\sigma_t^2=1\). Mehler's formula
\(\E h_j(\lambda y+\sqrt{1-\lambda^2}\xi)=\lambda^jh_j(y)\) gives the first claim;
the moments of \(Z\) are finite, so all expectations exist. Under the \(L^2\)
assumption the Ornstein--Uhlenbeck semigroup maps \(r_0\) to
\(r_t=\sum_j\lambda_t^je_j(0)h_j\), a contraction, and Parseval gives \(\chi^2_t\). The
orthogonal coordinates of \(X_t\) are exactly \(N(0,\bar\sigma_t^2)\) and independent of \(Y_t\),
which gives the ambient ratio, and
\(\E_{Q_t}h_i(u^\top x/\bar\sigma_t)h_j(v^\top x/\bar\sigma_t)=\delta_{ij}\inner uv^j\) gives the
correlation formula; \(e_j=0\) for \(1\le j<k^\ast\). For the mixture with
\(\tilde\tau=\tau/\sigma_0\), \(\int\phi_{\tilde\tau}(x-\tilde\mu)^2/\phi(x)\,dx<\infty\)
iff \(\tilde\tau^2<2\). \qed

\subsection{Proof of Theorem~\ref{thm:time} (optimal geometry time)}\label{pf:thm:time}
Under \(Q\) the residual \(X_0-m^Q_t(X_t)\) is independent of \(X_t\) with covariance
\((\sigma_0^2h/\bar\sigma_t^2)\Id\), so the null standard deviation is
\(\sigma_0\sqrt h\,\|\varphi\|_{L^2(Q_t)}/\bar\sigma_t\). By Theorem~\ref{thm:stein} the mean is
\((h/a)|\E_{P_t}\Stein_{Q_t}\varphi|\). Rotate so that \(u=e_1\) and expand
\(\varphi=\sum_i\sum_\beta\varphi_{i\beta}H_\beta e_i\) in the Hermite basis of
\(Q_t=N(0,\bar\sigma_t^2\Id)\). Since
\(\Stein_{Q_t}(H_\beta e_i)=-\bar\sigma_t^{-1}\sqrt{\beta_i+1}\,H_{\beta+e_i}\) and
\(\E_{P_t}H_\gamma=e_{\gamma_1}(t)\) if \(\gamma\) is supported on the first coordinate
and \(0\) otherwise,
\(\E_{P_t}\Stein_{Q_t}\varphi=-\bar\sigma_t^{-1}\sum_j\varphi_{1,je_1}\sqrt{j+1}\,e_{j+1}(t)\). By
Cauchy--Schwarz the supremum over \(\|\varphi\|=1\) of degree at most \(k^\ast-1\) is
\(\bar\sigma_t^{-1}(\sum_{i\le k^\ast}i\,e_i(t)^2)^{1/2}=\bar\sigma_t^{-1}\sqrt{k^\ast}|e_{k^\ast}(t)|\).
When the moments below order \(k^\ast\) are Gaussian, the cumulants of orders
\(3,\dots,k^\ast-1\) vanish and \(e_{k^\ast}(0)=\kappa/(\sigma_0^{k^\ast}\sqrt{k^\ast!})\);
by Theorem~\ref{thm:noiseop}, \(e_{k^\ast}(t)=\lambda_t^{k^\ast}e_{k^\ast}(0)\). Collecting
factors gives the formula. With \(x=a^2\in(0,1)\),
\(\log S_t=\mathrm{const}+\frac12\log(1-x)+\frac{k^\ast-1}2\log x-\frac{k^\ast}2\log(1+(\sigma_0^2-1)x)\),
whose derivative is a positive multiple of \((k^\ast-1)-(k^\ast+\sigma_0^2-1)x\),
decreasing and changing sign once in \((0,1)\). \qed
\section{Proofs for the covariance-visible regime and the one-sided change of measure}\label{app:proofs}

This appendix proves the results of Sections~\ref{sec:reductions}--\ref{sec:separation}
under Assumption~\ref{ass:model}, together with Lemma~\ref{lem:onesided}, which is
also used in Appendix~\ref{app:cm}. Throughout, \(t>0\) is fixed and we write
\(a=a_t\), \(h=h_t\), \(s^2=a^2\tau^2+h\), and \(\gamma=\gamma_t=N(0,h\Id_d)\); in this
appendix \(s\) is the standard deviation of a mixture component of \(u^\top X_t\), and
\(s_t\) is the score.

\begin{roadmap}[: how Appendix~\ref{app:proofs} is organized]
\begin{enumerate}[leftmargin=*,itemsep=1pt,topsep=1pt]
\item \emph{Reductions} (\ref{pf:lem:reduction}--\ref{pf:lem:density}): the score
problem becomes a one-dimensional regression of the posterior mean \(g_t\), and the
data density is compared with the orientation-free reference \(\gamma\).
\item \emph{Spectral core} (\ref{pf:lem:spectral}): the orientation-averaged
captured energy of any \(M\)-dimensional span is a Ky Fan sum of a Funk--Hecke
spectrum; this is the engine of every lower bound.
\item \emph{Cubic barrier and its refinements} (\ref{pf:thm:floor}--\ref{pf:cor:price}):
average over orientations, then every orientation for isotropic features,
typical orientations by concentration on the sphere, and sharpness by explicit
polynomial dictionaries.
\item \emph{Exact height} (\ref{pf:lem:onesided}--\ref{pf:thm:height}): the
one-sided change of measure removes the density factor from the main term.
\item \emph{Robustness and the adaptive side}
(\ref{pf:prop:offsubspace}--\ref{pf:cor:separation}): time change for data off
the subspace, monotonicity in \(d\), and the exact dimension invariance of the
PCA estimator and its sampler.
\end{enumerate}
\end{roadmap}

\subsection{Proof of Lemma~\ref{lem:reduction} (denoiser reduction)}\label{pf:lem:reduction}

Complete \(U\) to an orthogonal matrix \([U,V]\). Then
\(U^\top X_t=aZ+\sqrt h\,U^\top G\) and \(V^\top X_t=\sqrt h\,V^\top G\), where
\(U^\top G\) and \(V^\top G\) are independent standard Gaussian vectors that are
independent of \(Z\). Hence \(V^\top X_t\) is independent of
\((Z,U^\top X_t)\), and
\[
 \E[X_0\mid X_t]=U\,\E[Z\mid U^\top X_t,V^\top X_t]
 =U\,\E[Z\mid U^\top X_t]=Ug_t(U^\top X_t).
\]
The density of \(X_t\) is \(p_t(x)=\E\,\phi_{h,d}(x-aX_0)\), where
\(\phi_{h,d}\) is the \(N(0,h\Id_d)\) density. Since \(X_0\) has finite second
moments, differentiation under the expectation is justified by dominated
convergence and gives
\(\nabla p_t(x)=\E[\phi_{h,d}(x-aX_0)(aX_0-x)/h]\). Dividing by \(p_t(x)>0\)
yields Tweedie's identity \(s_t(x)=(a\,m_t(x)-x)/h\)
\citep{robbins1956empirical,efron2011tweedie}. Because \(m_t(x)\in\spanop(U)\),
\((\Id-UU^\top)s_t(x)=-(\Id-UU^\top)x/h\). Finally
\(\widehat s-s_t=a(\widehat m-m_t)/h\) pointwise, which gives item~3; if
\(\widehat m\) also takes values in \(\spanop(U)\), the difference is
tangential. \qed

\subsection{Proof of Lemma~\ref{lem:posterior} (posterior mean of a mixture coordinate)}\label{pf:lem:posterior}

Condition on the sign \(\epsilon\). Then \((Z,Y)\), with \(Y=aZ+\sqrt h\,\xi\), is
Gaussian with \(\E[Z\mid\epsilon]=\epsilon\mu\),
\(\E[Y\mid\epsilon]=a\epsilon\mu\), \(\mathrm{Var}(Y\mid\epsilon)=s^2\) and
\(\mathrm{Cov}(Z,Y\mid\epsilon)=a\tau^2\). Therefore
\[
 \E[Z\mid Y,\epsilon]=\epsilon\mu+\frac{a\tau^2}{s^2}(Y-a\epsilon\mu)
 =\epsilon\mu\frac{h}{s^2}+\frac{a\tau^2}{s^2}Y .
\]
Bayes' rule gives
\(\Prob(\epsilon=1\mid Y=y)\propto\exp(-(y-a\mu)^2/(2s^2))\) and the analogous
expression for \(\epsilon=-1\), so
\(\E[\epsilon\mid Y=y]=\tanh(a\mu y/s^2)\). Averaging over \(\epsilon\) proves the
formula, which is odd in \(y\). With independent coordinates, the posterior
mean of \(Z\) given \(aZ+\sqrt h\,\xi\) factorizes coordinatewise. \qed

\subsection{Proof of Lemma~\ref{lem:density} (density floor \(c_t\))}\label{pf:lem:density}

In the coordinates \((U^\top x,V^\top x)\) of the proof of
Lemma~\ref{lem:reduction}, \(X_t\) has density
\(\prod_{j=1}^kq_t(y_j)\cdot\phi_{h,d-k}(V^\top x)\) with \(y=U^\top x\), while
\(\gamma\) has density \(\prod_{j=1}^k\phi_h(y_j)\cdot\phi_{h,d-k}(V^\top x)\).
Both coordinate maps are orthogonal, so the ratio is
\(\prod_j\rho_t(u_j^\top x)\). For one coordinate,
\[
 \rho_t(y)=\frac{\E\,\phi_h(y-aZ_1)}{\phi_h(y)}
 =\E\exp\Bigl(\frac{2ayZ_1-a^2Z_1^2}{2h}\Bigr)
 =\E\Bigl[\cosh\Bigl(\frac{ayZ_1}{h}\Bigr)
 \exp\Bigl(-\frac{a^2Z_1^2}{2h}\Bigr)\Bigr],
\]
where the last step uses the symmetry \(Z_1\overset{d}{=}-Z_1\). Since
\(\cosh\geq1\), with equality at \(y=0\), we get \(\rho_t\geq\rho_t(0)=c_t\).
For \(Z_1\sim N(\pm\mu,\tau^2)\) and \(\beta=a^2/(2h)\), the Gaussian integral
\(\E e^{-\beta Z_1^2}=(1+2\beta\tau^2)^{-1/2}\exp(-\beta\mu^2/(1+2\beta\tau^2))\)
gives the closed form, because \(1+2\beta\tau^2=s^2/h\). \qed

\subsection{Proof of Lemma~\ref{lem:spectral} (orientation-averaged Ky Fan bound)}\label{pf:lem:spectral}

\begin{roadmap}
The captured energy \(\E_u\|\Pi_{\mathcal S}f_u\|^2\) is a trace against the
covariance operator \(T\) of the family \(\{f_u\}\) (Step~4). \(T\) has the same
spectrum as an integral operator on the sphere whose kernel depends only on
\(u^\top v\) (Steps~1--2), so its eigenfunctions are spherical harmonics and its
eigenvalues are explicit Funk--Hecke integrals (Step~3). Counting how many
eigenvalues an \(M\)-dimensional span can capture in each harmonic block gives the
explicit bound and the weights (Step~5).
\end{roadmap}

Write \(\zeta=x/\sqrt h\sim N(0,\Id_d)\) under \(\gamma\), and let \(\sigma\) be
the uniform probability measure on \(\Sph^{d-1}\).

\step{Step 1: Gram kernel} For unit vectors \(u,v\), the pair
\((u^\top\zeta,v^\top\zeta)\) is standard bivariate Gaussian with correlation
\(u^\top v\). Hence
\(\E[h_i(u^\top\zeta)h_j(v^\top\zeta)]=\delta_{ij}(u^\top v)^j\)
\citep[Fact~11.7 and Proposition~11.31]{odonnell2014analysis}, and
\[
 \ip{f_u}{f_v}_{L^2(\gamma)}=\kappa(u^\top v),\qquad
 \kappa(\varsigma)=\sum_{j}c_j^2\varsigma^j .
\]
Since \(\sum_jc_j^2=\norm{g}^2<\infty\), the series converges uniformly on
\([-1,1]\), and \(\kappa\) is continuous. Because
\(\norm{f_u-f_v}^2=2(\kappa(1)-\kappa(u^\top v))\), the map \(u\mapsto f_u\) is
continuous into \(L^2(\gamma)\).

\step{Step 2: two operators with the same spectrum} Define
\(A:L^2(\sigma)\to L^2(\gamma)\) by \(A\psi=\int\psi(u)f_u\,d\sigma(u)\), a
Bochner integral. Its adjoint is \((A^*F)(u)=\ip{F}{f_u}\). Then
\(T=AA^*=\int f_u\otimes f_u\,d\sigma(u)\), and \(K=A^*A\) is the integral
operator on \(L^2(\sigma)\) with continuous positive semidefinite kernel
\(\kappa(u^\top v)\). Both are trace class with
\(\tr T=\int\norm{f_u}^2d\sigma=\sum_jc_j^2\), and their nonzero eigenvalues
coincide with multiplicities, by the singular value decomposition of \(A\).

\step{Step 3: Funk--Hecke} By the Funk--Hecke formula
\citep[Chapter~2]{atkinson2012spherical}, every degree-\(\ell\) spherical
harmonic is an eigenfunction of \(K\). Its eigenvalue is
\[
 \Lambda_\ell=\frac{|\Sph^{d-2}|}{|\Sph^{d-1}|}
 \int_{-1}^1\kappa(\varsigma)P_{\ell,d}(\varsigma)
 (1-\varsigma^2)^{(d-3)/2}\,d\varsigma ,
\]
where \(P_{\ell,d}\) is the Legendre polynomial of degree \(\ell\) in dimension
\(d\), normalized by \(P_{\ell,d}(1)=1\). Spherical harmonics span
\(L^2(\sigma)\), so the eigenvalues of \(K\) are the \(\Lambda_\ell\), with
multiplicities \(N_{d,\ell}\). By uniform convergence,
\(\Lambda_\ell=\sum_jc_j^2\eta_{j,\ell}\), where \(\eta_{j,\ell}\) is the same
integral with \(\kappa(\varsigma)=\varsigma^j\). Rodrigues' formula
\citep[Chapter~2]{atkinson2012spherical} reads
\[
 P_{\ell,d}(\varsigma)(1-\varsigma^2)^{(d-3)/2}
 =(-1)^\ell R_{\ell,d}\frac{d^\ell}{d\varsigma^\ell}
 (1-\varsigma^2)^{\ell+(d-3)/2},\qquad
 R_{\ell,d}=\frac{\Gamma((d-1)/2)}{2^\ell\Gamma(\ell+(d-1)/2)} .
\]
For \(d\geq3\), integration by parts \(\ell\) times has no boundary terms and
gives
\[
 \eta_{j,\ell}=\frac{|\Sph^{d-2}|}{|\Sph^{d-1}|}R_{\ell,d}
 \frac{j!}{(j-\ell)!}\int_{-1}^1\varsigma^{j-\ell}
 (1-\varsigma^2)^{\ell+(d-3)/2}d\varsigma
\]
for \(\ell\leq j\), and \(\eta_{j,\ell}=0\) for \(\ell>j\). The integral
vanishes when \(j-\ell\) is odd. For \(j-\ell=2r\), it equals the beta integral
\(\Gamma(r+\tfrac12)\Gamma(\ell+\tfrac{d-1}2)/\Gamma(r+\ell+\tfrac d2)\).
Combining it with
\(|\Sph^{d-2}|/|\Sph^{d-1}|=\Gamma(d/2)/(\sqrt\pi\,\Gamma((d-1)/2))\),
\(\Gamma(r+\tfrac12)=\sqrt\pi(2r)!/(4^rr!)\) and
\(\Gamma(r+\ell+\tfrac d2)/\Gamma(\tfrac d2)
=2^{-(r+\ell)}\prod_{q=0}^{r+\ell-1}(d+2q)\) gives~\eqref{eq:eta}, since
\(r+\ell=(j+\ell)/2\). We record four consequences:
\begin{enumerate}
\item[(a)] \(\eta_{j,\ell}\geq0\);
\item[(b)] \(\eta_{j,\ell}=0\) unless \(\ell\leq j\) and \(\ell\equiv j\pmod 2\).
Since \(g\) is odd, only odd \(j\) occur, so \(\Lambda_\ell=0\) for even
\(\ell\);
\item[(c)] \(\sum_\ell N_{d,\ell}\eta_{j,\ell}=1\). This is the trace of the
operator with kernel \((u^\top v)^j\), which equals
\(\int(u^\top u)^jd\sigma=1\) by Mercer's theorem;
\item[(d)] \(N_{d,1}\eta_{j,1}=\prod_{q=1}^{m}\frac{2q+1}{d+2q}\) for
\(j=2m+1\). Indeed \(\eta_{j,1}=\E[\varsigma^{j+1}]\) for \(\varsigma\) the
first coordinate of a uniform point. For \(\zeta\sim N(0,\Id_d)\), independence
of \(\norm\zeta\) and \(\zeta/\norm\zeta\) gives
\(\E\varsigma^{2m+2}=\E\zeta_1^{2m+2}/\E\norm\zeta^{2m+2}
=\prod_{q=0}^m(2q+1)/(d+2q)\). Multiplying by \(N_{d,1}=d\) cancels the
\(q=0\) factor.
\end{enumerate}

\step{Step 4: Ky Fan} Condition on \(\mathcal S\), which is independent of
\(u\). Let \(e_1,\dots,e_{M'}\), \(M'\leq M\), be an orthonormal basis of
\(\mathcal S\). Then
\(\E_u\norm{\Pi_{\mathcal S}f_u}^2=\sum_i\E_u\ip{e_i}{f_u}^2
=\sum_i\ip{e_i}{Te_i}=\tr(\Pi_{\mathcal S}T)\), and Ky Fan's maximum principle
\citep{fan1949theorem} bounds it by \(\sum_{i\leq M}\nu_i(T)\). Since
\(\E_u\norm{f_u}^2=\tr T\) and \(f_u-\Pi_{\mathcal S}f_u\perp\Pi_{\mathcal S}f_u\),
the first inequality follows after taking the expectation over \(\mathcal S\).

\step{Step 5: explicit form} The \(M\) largest eigenvalues take at most
\(\min\{N_{d,\ell},M\}\) values from block \(\ell\), so
\(\tr T-\sum_{i\leq M}\nu_i\geq\sum_\ell\Lambda_\ell(N_{d,\ell}-M)_+\). Pascal's
rule gives \(N_{d,\ell}=\binom{\ell+d-2}{d-2}+\binom{\ell+d-3}{d-2}\), which is
nondecreasing in \(\ell\). Hence \((N_{d,\ell}-M)_+\geq
N_{d,\ell}(1-M/N_{d,J})_+\) for \(\ell\geq J\), and
\[
 \tr T-\sum_{i\leq M}\nu_i\;\geq\;\Bigl(1-\frac M{N_{d,J}}\Bigr)_+
 \sum_{\ell\geq J}N_{d,\ell}\Lambda_\ell
 =\Bigl(1-\frac M{N_{d,J}}\Bigr)_+\sum_jc_j^2\sum_{\ell\geq J}N_{d,\ell}\eta_{j,\ell}.
\]
By~(b) the inner sum vanishes for \(j<J\); for \(j\geq J\) it is \(w_{j,J}(d)\),
which lies in \([0,1]\) by~(a) and~(c). This proves the displayed inequality.

For the weights, let \(j\geq J\). The inner sum contains the term \(\ell=j\),
so \(w_{j,J}(d)\geq N_{d,j}\eta_{j,j}\). With \(r=0\) in~\eqref{eq:eta},
\(\eta_{j,j}=j!/\prod_{q=0}^{j-1}(d+2q)\), and
\(N_{d,j}=\frac{2j+d-2}{j}\binom{j+d-3}{j-1}=(d+2j-2)\prod_{i=0}^{j-2}(d-1+i)/j!\).
Cancelling the factor \(d+2j-2\) gives
\(N_{d,j}\eta_{j,j}=\prod_{i=0}^{j-2}(d-1+i)/(d+2i)\). Each factor is
\(1-x_i\) with \(x_i=(i+1)/(d+2i)\in[0,(i+1)/d]\), and
\(\prod_i(1-x_i)\geq1-\sum_ix_i\geq1-j(j-1)/(2d)\). Finally, for \(J=3\) and odd
\(j=2m+1\), (b) leaves only \(\ell=1\) below level three, so
\(w_{j,3}(d)=1-N_{d,1}\eta_{j,1}=1-\prod_{q=1}^m\frac{2q+1}{d+2q}\) by~(c)
and~(d). The product is at most its first factor \(3/(d+2)\), because each later
factor is at most one. Summing, \(\sum_{j\geq3}c_j^2w_{j,3}(d)\geq
(1-\frac3{d+2})\sum_{j\geq3}c_j^2\). \qed

\subsection{Proof of Theorem~\ref{thm:floor} (cubic barrier, orientation average)}\label{pf:thm:floor}

\begin{roadmap}
(i) Reduce the score risk to \(k\) scalar regressions of \(g_t(u_j^\top x)\)
(Lemma~\ref{lem:reduction}); (ii) replace \(P^U_t\) by \(\gamma\) at the cost of
\(c_t^k\) (Lemma~\ref{lem:density}); (iii) average over the Haar orientation and
apply Lemma~\ref{lem:spectral} to the enlarged span \(\mathcal S+\Aff\).
\end{roadmap}

\step{Step 1: scalar reduction} Fix \(t\). Let \(\mathcal S'=\mathcal S+\Aff\), a random subspace of dimension
at most \(M+d+1\) that is independent of \(U\), since \(\Aff\) is deterministic.
Let \(X\sim P_t^U\) be independent of the training data. By
Lemma~\ref{lem:reduction}, Pythagoras in the orthonormal directions
\(u_1,\dots,u_k\), and Lemma~\ref{lem:posterior},
\[
 \risk_t(\widehat s;U)=\Bigl(\frac ah\Bigr)^2\E\norm{\widehat m(X)-m_t(X)}^2
 \geq\Bigl(\frac ah\Bigr)^2\sum_{j=1}^k
 \E\bigl(u_j^\top\widehat m(X)-g_t(u_j^\top X)\bigr)^2 ,
\]
where \(g_t\) now denotes the scalar map of Lemma~\ref{lem:posterior}.
\step{Step 2: change of reference measure} Condition on the training data and \(\omega\). Almost surely,
\(x\mapsto u_j^\top\widehat m(x)\) lies in \(\mathcal S'\). If the conditional
expectation is infinite there is nothing to prove. Otherwise
Lemma~\ref{lem:density} gives
\begin{equation}\label{eq:pointwise}
 \E\bigl[(u_j^\top\widehat m(X)-g_t(u_j^\top X))^2\bigm|\cdot\bigr]
 \geq c_t^k\inf_{f\in\mathcal S'}\norm{f-g_t(u_j^\top\cdot)}^2_{L^2(\gamma)}
 =c_t^k\norm{f_{u_j}-\Pi_{V}f_{u_j}}^2_{L^2(\gamma)} ,
\end{equation}
with \(f_u=g_t(u^\top\cdot)\in L^2(\gamma)\), since \(g_t\) grows at most
linearly, and \(V=\mathcal S'\cap L^2(\gamma)\).
\step{Step 3: average over orientations} Now let \(U\) be Haar and take
the expectation over \(U\) as well. Each column \(u_j\) is uniform on
\(\Sph^{d-1}\) and independent of \(\omega\), hence of \(V\).
Lemma~\ref{lem:spectral}, applied with dimension at most \(M+d+1\), level \(J\)
and the normalized Hermite coefficients of \(g_t\) in \(L^2(N(0,h))\), bounds
each term from below by \((1-\frac{M+d+1}{N_{d,J}})_+\sum_{j\ge J}c_j^2w_{j,J}(d)\),
by \((1-\frac3{d+2})(1-\frac{M+d+1}{N_{d,3}})_+\kappa_t\) when \(J=3\), or by
\(\tr T_t-\sum_{i\leq M+d+1}\nu_i(T_t)\) in the Ky Fan form. Summing over the
\(k\) columns proves~\eqref{eq:floor} and the two other bounds. \qed

\begin{remark}[Truncated series]
Let \(T_{t,J}\) be the operator built from \(c_0,\dots,c_J\). Then
\(T_t=T_{t,J}+R\) with \(R\succeq0\), because the Hermite degrees are
orthogonal. Hence
\(\sum_{i\leq M}\nu_i(T_t)\leq\sum_{i\leq M}\nu_i(T_{t,J})+\tr R\), and the
Ky Fan floor computed from a truncated series is still a valid lower bound. Our
numerical evaluations use \(J=31\).
\end{remark}

\subsection{Proofs of Corollary~\ref{cor:ambient} and Proposition~\ref{prop:fixed} (geometry-blind random features; every orientation for isotropic features)}\label{pf:cor:ambient}\label{pf:prop:fixed}

\step{Corollary~\ref{cor:ambient}} For each \(\beta\), the coordinates of
\(\varphi_\beta\) depend on \(W\), \(b\), \(t\), \(\beta\) and
\(\varsigma_t\). The sample covariance of \(X_0^{(i)}=UZ^{(i)}\) is
\(U\widehat C_ZU^\top\), so \(\tr\widehat\Sigma=\tr\widehat C_Z\) is a function
of the latent data alone. With \(\omega=(W,b,Z^{(1)},\dots,Z^{(n)})\), whose law
does not depend on \(U\), the span of all \(L\) feature maps has dimension at
most \(L(p+1)\). Every coordinate of
\(\widehat m=m_G+\widehat UB^\top\varphi_\beta\) is an affine function plus a
combination of the coordinates of the selected \(\varphi_\beta\). Apply
Theorem~\ref{thm:floor}.

\step{Proposition~\ref{prop:fixed}} Let \(Q\) be orthogonal. Replace \(U\) by
\(QU\), every corruption and validation noise \(G\) by \(QG\), and \(W\) by
\(QW\), keeping \(Z\), \(b\) and all grids fixed. The new training data are \(Q\)
times the old ones. Principal subspaces rotate,
\(\widehat\Pi\mapsto Q\widehat\Pi Q^\top\), and \(\tr\widehat\Sigma\) is
unchanged. Each feature satisfies \(\varphi'_\beta(Qx)=\varphi_\beta(x)\),
since \((QW)^\top Qx=W^\top x\). The Gaussian denoiser satisfies
\(m'_G(Qx)=Qm_G(x)\). The ridge targets, expressed through the projector as in
the proof of Theorem~\ref{thm:invariance}, are rotated by \(Q\). Hence every
validation error is unchanged, the same \((\lambda,\beta)\) is selected, and
\(\widehat m'(Qx)=Q\widehat m(x)\). The new test point is \(QX\), and
\(m'_t(Qx)=Qm_t(x)\), so the risk is unchanged. Since \((QG,QW)\) has the same
law as \((G,W)\) and is independent of \(Z\) and \(b\), the risk has the same
law at \(QU\) as at \(U\). The orthogonal group acts transitively on the Stiefel
manifold, so \(\risk_t(\widehat s;U)\) is constant in \(U\). It therefore equals
its Haar average \(\bar\risk_t(\widehat s)\), which is bounded below by
Corollary~\ref{cor:ambient}. \qed

\subsection{Proof of Theorem~\ref{thm:typical} (typical orientations)}\label{pf:thm:typical}

\begin{roadmap}
Define the orientation-wise floor \(\Phi_t(u)\) (the expected distance of the
nonlinear part of \(f_u\) to the span). It lower-bounds the risk at every
orientation (Step~1), is Lipschitz on the sphere (Step~2), has the cubic-barrier
mean (Step~3), and therefore concentrates by the logarithmic Sobolev inequality on
\(\Sph^{d-1}\) (Step~4).
\end{roadmap}

Fix \(t\) and write \(a,h,\kappa,J\) for \(a_t,h_t,\kappa_t,J_t\). Let
\(V(\omega)=(\mathcal S(\omega)+\Aff)\cap L^2(\gamma)\), a subspace of dimension
at most \(M+d+1\) that contains \(\Aff\). For \(u\in\Sph^{d-1}\) let
\(f^{\rm nl}_u=f_u-c_1h_1(u^\top\cdot/\sqrt h)\), the part of \(f_u\) orthogonal
to the linear functions, so that \(\norm{f^{\rm nl}_u}^2=\kappa\). Define
\[
 \Phi_t(u)=\E_\omega\,\mathrm{dist}^2_{L^2(\gamma)}\bigl(f^{\rm nl}_u,V(\omega)\bigr),
\]
where \(\E_\omega\) is taken under the law of \(\omega\), which does not depend
on \(U\). Since \(0\in V\), \(0\le\Phi_t\leq\kappa\).

\step{Step 1: pointwise reduction} Fix \(U\). The display before~\eqref{eq:pointwise}
and~\eqref{eq:pointwise} itself hold for every fixed \(U\), conditionally on the
training data and \(\omega\). The linear part \(c_1h_1(u^\top x/\sqrt h)\) lies in
\(\Aff\subset V\), so \(\mathrm{dist}(f_u,V)=\mathrm{dist}(f^{\rm nl}_u,V)\).
Taking expectations, \(\risk_t(\widehat s;U)\geq(a/h)^2c_t^k\sum_i\Phi_t(u_i)\).

\step{Step 2: Lipschitz bound} By Step~1 of the proof of
Lemma~\ref{lem:spectral},
\(\norm{f^{\rm nl}_u-f^{\rm nl}_v}^2=2\sum_{j\geq3}c_j^2(1-(u^\top v)^j)\). For
\(\varsigma\in[-1,1]\) and odd \(j\),
\(1-\varsigma^j=(1-\varsigma)\sum_{i<j}\varsigma^i\leq j(1-\varsigma)\). Hence
\(\norm{f^{\rm nl}_u-f^{\rm nl}_v}^2\leq2J(1-u^\top v)=J\norm{u-v}^2\). The map
\(f\mapsto\mathrm{dist}(f,V)\) is \(1\)-Lipschitz and bounded by
\(\sqrt\kappa\) on \(\{f^{\rm nl}_u\}\), so
\(|\mathrm{dist}^2(f^{\rm nl}_u,V)-\mathrm{dist}^2(f^{\rm nl}_v,V)|\leq
2\sqrt{\kappa J}\norm{u-v}\); averaging over \(\omega\) preserves the bound.
The identity \(J=h\E[g_t'(\sqrt h\xi)^2]-c_1^2\) follows from Gaussian
integration by parts,
\(\sqrt h\,\E[g_t'(\sqrt h\xi)h_{j-1}(\xi)]=\sqrt j\,c_j\), and Parseval's
identity.

\step{Step 3: mean} Let \(u\) be uniform on \(\Sph^{d-1}\) and independent of
\(\omega\). The function \(g_t-c_1h_1(\cdot/\sqrt h)\) is odd with Hermite
coefficients \(c_j\), \(j\geq3\). Lemma~\ref{lem:spectral} with \(J=3\), applied
to it and to \(V(\omega)\), gives
\(\int\Phi_t\,d\sigma\geq(1-\frac3{d+2})(1-\frac{M+d+1}{N_{d,3}})_+\kappa\).

\step{Step 4: concentration} For \(d\geq3\), the round sphere
\(\Sph^{d-1}\) has Ricci curvature \(d-2\). By the Bakry--\'Emery criterion its
uniform measure \(\sigma\) satisfies a logarithmic Sobolev inequality with
constant \(1/(d-2)\), and Herbst's argument gives
\(\sigma(F\leq\int F\,d\sigma-s)\leq\exp\{-(d-2)s^2/(2L^2)\}\) for every
function \(F\) that is \(L\)-Lipschitz for the geodesic distance
\citep[Chapter~5]{ledoux2001concentration}; see also \citet{bakry2014analysis}.
A function that is \(L\)-Lipschitz for the Euclidean distance is \(L\)-Lipschitz
for the geodesic distance. Apply this to \(F=\Phi_t\) with
\(L=2\sqrt{\kappa J}\). Each column of a Haar-distributed \(U\) is uniform on
\(\Sph^{d-1}\), so by a union bound, outside a set of Haar measure at most
\(k\exp\{-(d-2)s^2/(8\kappa J)\}\), every column satisfies
\(\Phi_t(u_i)\geq\int\Phi_t\,d\sigma-s\). For such \(U\), Steps~1 and~3 give
\(\risk_t(\widehat s;U)\geq k(a/h)^2c_t^k(\int\Phi_t\,d\sigma-s)\geq
F_d(M;t)-k(a/h)^2c_t^k\,s\). \qed

\subsection{Proof of Theorem~\ref{thm:sharp} (matching polynomial dictionaries)}\label{pf:thm:sharp}

\step{Step 1: the matching estimator} Let \(\pi\) be a univariate polynomial of degree at most \(J\), applied
coordinatewise, and \(\widehat m(x)=U\pi(U^\top x)\). Its \(i\)-th coordinate
\(\sum_jU_{ij}\pi(u_j^\top x)\) is a polynomial of degree at most \(J\) in \(x\),
so \(\widehat m\) is supported by \(\mathcal P_{d,J}\). By
Lemma~\ref{lem:reduction}, \(m_t(x)=Ug_t(U^\top x)\), and \(Y=U^\top X_t\) has
independent coordinates with density \(q_t\). Item~3 of
Lemma~\ref{lem:reduction} and orthonormality of the columns of \(U\) give
\(\risk_t(\widehat s;U)=(a_t/h_t)^2\sum_{j=1}^k\E(\pi(Y_j)-g_t(Y_j))^2
=k(a_t/h_t)^2\E(\pi(Y_1)-g_t(Y_1))^2\). The polynomials of degree at most \(J\)
form a finite-dimensional subspace of \(L^2(q_t)\), because \(q_t\) has all
moments; the minimum \(\varepsilon_J(t)\) is attained by the orthogonal
projection, which proves the risk identity. The spaces are nested, so
\(\varepsilon_J\) is nonincreasing.

\step{Step 2: \(\varepsilon_J\to0\)} Fix \(\sigma^2>s^2=a_t^2\tau^2+h_t\). The function
\(y\mapsto-(y\mp a_t\mu)^2/(2s^2)+y^2/(2\sigma^2)\) is a concave quadratic, so
\(q_t\leq C\phi_{\sigma^2}\) for a constant \(C\). Since \(g_t\) grows at most
linearly, \(g_t\in L^2(N(0,\sigma^2))\), where the Hermite polynomials are
complete. If \(\pi_J\) denotes the truncation of its Hermite expansion at degree
\(J\), then \(\varepsilon_J\leq\norm{g_t-\pi_J}_{L^2(q_t)}^2\leq
C\norm{g_t-\pi_J}^2_{L^2(N(0,\sigma^2))}\to0\).

\step{Step 3: the case \(J=1\) and dimension counts} For \(J=1\), the best affine predictor of \(g_t(Y)=\E[Z_1\mid Y]\) under
\(q_t\) is the best affine predictor of \(Z_1\), the linear least-squares
estimator \(\mathrm{Cov}(Z_1,Y)\mathrm{Var}(Y)^{-1}Y\). It coincides with the
posterior mean under the moment-matched Gaussian prior, whose score risk is the
population Gaussian control. Finally, \(D_{d,J}=\binom{d+J}J\) counts monomials
of degree at most \(J\). Pascal's rule gives
\(D_{d,J}=\binom{d+J-1}{J}+D_{d,J-1}\), and
\(N_{d,J}=\binom{d+J-1}J-\binom{d+J-3}{J-2}\), so
\(D_{d,J}-N_{d,J}=D_{d,J-1}+\binom{d+J-3}{J-2}=O(d^{J-1})\). \qed

\subsection{Proofs of Corollaries~\ref{cor:staircase} and~\ref{cor:price} (polynomial staircase; cubic feature price)}\label{pf:cor:staircase}\label{pf:cor:price}

\step{Corollary~\ref{cor:staircase}} Part~1 is Theorem~\ref{thm:floor} at
level \(J\), with \((1-(M+d+1)/N_{d,J})_+\geq\delta\) and the product lower
bound~\eqref{eq:weights} on \(w_{j,J}(d)\). Each product tends to one as
\(d\to\infty\) and lies in \([0,1]\), and \(\sum_jc_j^2<\infty\), so dominated
convergence gives the limit; it is positive because \(\kappa_{J,t}>0\). Part~2
is Theorem~\ref{thm:sharp}, after enlarging \(\mathcal P_{d,J}\) by \(M-D_{d,J}\)
fixed functions. Both \(\binom{d+J-1}{J}\) and \(\binom{d+J}J\) equal
\((1+O(1/d))d^J/J!\), and \(\binom{d+J-3}{J-2}=O(d^{J-2})\). For the last claim,
Theorem~\ref{thm:floor} at level \(J\) and \(\bar\risk_t\leq\varepsilon\) give
\(M+d+1\geq N_{d,J}(1-\varepsilon/\{k(a_t/h_t)^2c_t^k\kappa_{J}(d)\})\), with
\(\kappa_J(d)=\sum_{j\ge J}c_j^2w_{j,J}(d)\to\kappa_{J,t}\); for
\(\varepsilon<k(a_t/h_t)^2c_t^k\kappa_{J,t}\) the bracket is eventually bounded
below by a positive constant.

\step{Corollary~\ref{cor:price}} By Theorem~\ref{thm:floor},
\(B_d(t)\,y_+\leq\varepsilon\) with \(y=1-(M+d+1)/N_{d,3}\). Since \(y\leq y_+\),
we get \(y\leq\varepsilon/B_d(t)\), which rearranges to part~1. Part~2 is the
same argument with Proposition~\ref{prop:fixed}. For part~3, the set where
\(\risk_t\leq\varepsilon\) has Haar measure larger than the exceptional set of
Theorem~\ref{thm:typical}, so it contains an orientation \(U\) outside it; there
\(\varepsilon\geq F_d(M;t)-k(a_t/h_t)^2c_t^ks=B_d(t)y_+-k(a_t/h_t)^2c_t^ks\),
and we conclude as in part~1. If the bracket is at least \(\delta\), then
\(M\geq\delta N_{d,3}-d-1\), and \(N_{d,3}\sim d^3/6\) gives \(\Omega(d^3)\).
For a fixed \(\varepsilon<F_\infty(t)\) and small \(s\), the margin holds for all
large \(d\) because \(B_d(t)\to F_\infty(t)\). The \(O(d^3)\) statement is
Theorem~\ref{thm:sharp} with \(J=3\). \qed

\subsection{Proof of Lemma~\ref{lem:onesided} (one-sided change of measure)}\label{pf:lem:onesided}

Since \(\mathcal V\) is a linear space, minimizing first over multiples of each
\(f\) gives
\(\inf_{f\in\mathcal V}\norm{r-f}_P^2=\norm r_P^2-\sup\ip rf_P^2/\norm f_P^2\),
the supremum over \(f\in\mathcal V\) with \(\norm f_P>0\); such \(f\) have
\(\norm f_Q>0\). Since \(\ip rf_P=\ip{\rho r}f_Q\) and
\(\norm f_P^2\geq c\norm f_Q^2\), the supremum is at most
\(c^{-1}\sup_f\ip{\rho r}f_Q^2/\norm f_Q^2=c^{-1}\norm{\Pi^Q_{\mathcal V}(\rho r)}_Q^2\).
\qed

\subsection{Proof of Theorem~\ref{thm:height} (exact height of the first step)}\label{pf:thm:height}

\begin{roadmap}
The proof of Theorem~\ref{thm:floor} pays \(c_t\) on the whole squared error. Here
the main term \(\|r\|^2_{L^2(P)}=\varepsilon_1(t)\) is kept exact, and the change
of measure is applied only to the cross term through Lemma~\ref{lem:onesided}
(Step~2). What remains is the captured energy of the weighted residual
\(\psi=\rho\hat g\), which has no degree-one Hermite component and is bounded by
Lemma~\ref{lem:spectral} at level three (Step~3). Step~1 checks that \(\psi\) has
finite energy.
\end{roadmap}

Write \(a,h,c,\rho,\hat g,\psi,\Psi\) for \(a_t,h_t,c_t,\rho_t,\hat g_t,\psi_t,\Psi_t\),
\(s^2=a^2\tau^2+h\), \(P=P_t^u\) and \(\gamma=N(0,h\Id_d)\).

\step{Step 1: finiteness of \(\Psi_t\)} By Lemma~\ref{lem:posterior}, \(\hat g\) is a bounded
function plus a linear one. The density \(q_t\) is at most
\((2\pi s^2)^{-1/2}\exp\{-(|y|-a\mu)^2/(2s^2)\}\), so
\(\rho^2\phi_h=q_t^2/\phi_h\) is at most a constant times
\(\exp\{-(|y|-a\mu)^2/s^2+y^2/(2h)\}\). The coefficient of \(y^2\) in the
exponent, \(1/(2h)-1/s^2\), is negative exactly when \(s^2<2h\), that is, when
\(a^2\tau^2<h\). Hence \(\Psi=\int\hat g^2\,q_t^2/\phi_h\,dy<\infty\).

\step{Step 2: one-sided reduction} By Lemma~\ref{lem:density} with \(k=1\),
\(dP/d\gamma(x)=\rho(u^\top x)\geq c\), so
\(\norm f^2_{L^2(P)}\geq c\norm f^2_{L^2(\gamma)}\) for every \(f\), and
\(L^2(P)\subseteq L^2(\gamma)\). Condition on the training data and on
\(\omega\), and let \(\mathcal S'=\mathcal S+\Aff\). If
\(\E\norm{\widehat m(X)-m_t(X)}^2=\infty\) there is nothing to prove. Otherwise
every coordinate of \(\widehat m\) lies in
\(V_u=\mathcal S'\cap L^2(P)\), because \(m_t\in L^2(P)\). By
Lemma~\ref{lem:reduction},
\(m_t(x)=u\,g_t(u^\top x)\). The field \(u\,\alpha_t\,u^\top x\) lies in
\(V_u^d\), since affine functions are in \(L^2(P)\), so
\[
 \E\norm{\widehat m(X)-m_t(X)}^2\;\geq\;\inf_{f\in V_u^d}\norm{m_t-f}^2_{L^2(P)}
 =\inf_{f\in V_u^d}\norm{r-f}^2_{L^2(P)},\qquad r(x)=u\,\hat g(u^\top x).
\]
Here \(\norm r^2_{L^2(P)}=\E_{Y\sim q_t}\hat g(Y)^2=\varepsilon_1(t)\), and
\(\rho r=u\,\psi(u^\top x)\) has \(\norm{\rho r}^2_{L^2(\gamma)}=\Psi\).
Lemma~\ref{lem:onesided} with \(Q=\gamma\) and \(\mathcal V=V_u^d\) gives the
lower bound \(\varepsilon_1(t)-c^{-1}\norm{\Pi^\gamma_{V_u^d}(\rho r)}^2\). The
space \(V_u\) may depend on \(u\), but it is contained in
\(W=\mathcal S'\cap L^2(\gamma)\), which does not. Projecting onto a larger space
does not decrease the norm. Projection onto \(W^d\) acts coordinatewise and
\(|u|=1\), so
\(\norm{\Pi^\gamma_{V_u^d}(\rho r)}^2\leq\norm{\Pi_W f_u}^2_{L^2(\gamma)}\) with
\(f_u=\psi(u^\top\cdot)\).

\step{Step 3: averaging over the orientation} The function \(\psi\) is odd, because
\(\rho\) is even and \(\hat g\) is odd. Its normalized Hermite coefficients
\(b_j=\E[\psi(\sqrt h\xi)h_j(\xi)]\) vanish for even \(j\), and
\[
 b_1=\frac1{\sqrt h}\E_{Y\sim N(0,h)}[\rho(Y)\hat g(Y)Y]
 =\frac1{\sqrt h}\E_{Y\sim q_t}[\hat g(Y)Y]=0
\]
by the definition of \(\alpha_t\). Hence \(\sum_{j\geq3}b_j^2=\Psi\). The space
\(W\) has dimension at most \(D\) and is independent of \(u\).
Lemma~\ref{lem:spectral} applied to \(\psi\), with \(\E\norm{f_u}^2=\tr
T^\psi_t=\Psi\), gives
\[
 \E_u\norm{\Pi_Wf_u}^2\;\leq\;\sum_{i\leq D}\nu_i(T^\psi_t)\;\leq\;
 \Psi\Bigl[1-\Bigl(1-\frac D{N_{d,3}}\Bigr)_+\Bigl(1-\frac3{d+2}\Bigr)\Bigr]
 \;\leq\;\Psi\Bigl(\frac3{d+2}+\frac D{N_{d,3}}\Bigr),
\]
where the last step uses \(1-(1-x)_+(1-y)\leq x+y\) for \(x\geq0\) and
\(y\in[0,1]\). Take expectations over \(u\), the training data and \(\omega\),
and multiply by \((a/h)^2\) (Lemma~\ref{lem:reduction}). If \(M=o(d^3)\), the
bracket tends to zero while \(\Psi\), \(c\) and \(\varepsilon_1(t)\) do not
depend on \(d\). \qed

\begin{remark}[Truncated series for the Ky Fan form]
Let \(\psi_J\) be the Hermite series of \(\psi\) truncated at degree \(J\) and
\(\eta=\Psi-\sum_{j\leq J}b_j^2\). By the triangle inequality in
\(L^2(\gamma)\) and Minkowski's inequality over \(u\),
\(\E_u\norm{\Pi_Wf_u}^2\leq(\{\sum_{i\leq D}\nu_i(T^{\psi_J}_t)\}^{1/2}+\eta^{1/2})^2\).
Table~\ref{tab:height} uses \(J=61\).
\end{remark}

\subsection{Proof of Proposition~\ref{prop:offsubspace} (data off the subspace)}\label{pf:prop:offsubspace}

Write \(a,h\) for \(a_t,h_t\) and \(a',h'\) for \(a_{t'},h_{t'}\), so that
\(a'=a/\lambda\) and \(h'=1-a^2/\lambda^2=(h+a^2\sigma_0^2)/\lambda^2\). Then
\(aX_0^\sigma+\sqrt hG=aUZ+\sqrt{h+a^2\sigma_0^2}\,G''\) with \(G''\) standard
Gaussian and independent of \(Z\), and
\(\lambda X_{t'}=\lambda a'UZ+\lambda\sqrt{h'}\,G'\) has the same law. The
densities satisfy \(p^\sigma_t(x)=\lambda^{-d}p_{t'}(x/\lambda)\), which gives
the score identity. Let \(\widehat s\) have coordinates in \(\mathcal S+\Aff\);
by~\eqref{eq:tweedie} this is equivalent to \(\widehat m\) being supported by
\(\mathcal S\). Define \(\widehat s'(y)=\lambda\widehat s(\lambda y)\). Its
coordinates lie in \(\mathcal S_\lambda+\Aff\), where
\(\mathcal S_\lambda=\{f(\lambda\,\cdot):f\in\mathcal S\}\) is an
orientation-independent dictionary of the same dimension, and
\(\widehat s'=(a'\widehat m'-y)/h'\) for \(\widehat m'=(h'\widehat s'+y)/a'\),
which is supported by \(\mathcal S_\lambda\). For every \(U\), with
\(X\sim P^\sigma_t\) written as \(\lambda X'\),
\[
 \E\norm{\widehat s(X)-s^\sigma_t(X)}^2
 =\E\norm{\widehat s(\lambda X')-\lambda^{-1}s_{t'}(X')}^2
 =\lambda^{-2}\,\E\norm{\widehat s'(X')-s_{t'}(X')}^2 .
\]
The right side is \(\lambda^{-2}\) times the noiseless risk of \(\widehat s'\)
at time \(t'\), to which Theorems~\ref{thm:floor} and~\ref{thm:typical} apply.
The dictionary may depend on the noisy training data through rotation-invariant
statistics, such as \(\tr\widehat\Sigma\), whose law given \(U\) does not depend
on \(U\). The proof of Proposition~\ref{prop:fixed} applies verbatim when the
data noise \(G_0\) is rotated together with \(G\) and \(W\). \qed

\subsection{Proof of Lemma~\ref{lem:monotone} (monotonicity in \(d\))}\label{pf:lem:monotone}

Write \(F_d(M;t)=F_\infty(t)\,\alpha_d\,\beta_d^+\) with
\(\alpha_d=1-3/(d+2)\) and \(\beta_d=1-(M+d+1)/N_{d,3}\). The factor
\(\alpha_d\) increases in \(d\). For \(\beta_d\), note
\(N_{d+1,3}-N_{d,3}=d(d+3)/2\). The ratio \((M+d+1)/N_{d,3}\) does not increase
from \(d\) to \(d+1\) if and only if \(N_{d,3}\leq(M+d+1)(N_{d+1,3}-N_{d,3})\),
that is, \((d-1)(d+4)\leq3(M+d+1)(d+3)\). This holds for every \(M\geq0\)
because \(3(d+1)(d+3)-(d-1)(d+4)=2d^2+9d+13>0\). Hence \(\beta_d\), and so
\(\beta_d^+\) and \(F_d\), are nondecreasing in \(d\). As \(d\to\infty\),
\(\alpha_d\to1\) and \(\beta_d\to1\). If \(r<F_\infty\), some \(d\) has
\(F_d>r\), and monotonicity gives the interval; if \(r\geq F_\infty\), then
\(F_d\leq F_\infty\leq r\) for all \(d\). \qed

\subsection{Proof of Theorem~\ref{thm:invariance} (PCA dimension invariance)}\label{pf:thm:invariance}

\begin{roadmap}
Noiseless data make PCA exact (Step~1). Every quantity the estimator computes,
including tuning, depends on the estimated basis only through the projector
(Step~2). In the latent coordinates the whole procedure is a fixed function of
Gaussian quantities whose joint law is free of \(d\) and \(U\) (Step~3), and the
risk inherits this (Step~4).
\end{roadmap}

\step{Step 1: exact recovery} The centered training matrix is \(\bar Z U^\top\), where
\(\bar Z\in\R^{n\times k}\) holds the centered latent rows. The latent law has a
density on \(\R^k\), and \(n\geq k+1\). Hence \(n-1\geq k\) of the centered
points are in general position almost surely, and \(\bar Z\) has rank \(k\).
The right singular vectors with nonzero singular values span \(\spanop(U)\), so
\(\widehat\Pi=UU^\top\).

\step{Step 2: projector dependence} Two orthonormal bases of the same subspace differ
by an orthogonal \(O\in\R^{k\times k}\). Replacing \(\widehat U\) by
\(\widehat UO\) maps \(\widehat\mu\mapsto O^\top\widehat\mu\) and
\(\widehat\Sigma\mapsto O^\top\widehat\Sigma O\). Since
\((a^2O^\top\widehat\Sigma O+h\Id)^{-1}=O^\top(a^2\widehat\Sigma+h\Id)^{-1}O\),
\(m_G\) is unchanged, and so is \(\tr\widehat\Sigma\). The PCA features use
\(\Pi=\widehat\Pi\) by definition. Let \(\Phi\in\R^{N\times(p+1)}\) be the
training design and \(R\in\R^{N\times d}\) have rows
\((x_0^{(i)}-m_G(x_t^{(i)}))^\top\). The ridge solution is
\(B=(\Phi^\top\Phi+N\lambda\Id)^{-1}\Phi^\top R\widehat U\), so
\[
 \widehat UB^\top\varphi(x)
 =\widehat\Pi R^\top\Phi(\Phi^\top\Phi+N\lambda\Id)^{-1}\varphi(x).
\]
The validation error is
\(\norm{(\Phi_v(\Phi^\top\Phi+N\lambda\Id)^{-1}\Phi^\top R-R_v)\widehat\Pi}_F^2/N_v\),
because \(\norm{A\widehat U}_F=\norm{A\widehat\Pi}_F\). Every quantity, the
selection included, is therefore a function of \(\widehat\Pi\). When
\(\widehat\Pi=UU^\top\) and \(W,b\) and the noise are shared, the PCA and oracle
estimators are the same function. This proves item~1.

\step{Step 3: latent representation} Take \(\widehat U=U\), which is allowed by item~1.
Write \(y=U^\top x\). The features are
\(\varphi(x)=(1,p^{-1/2}\tanh((U^\top W)^\top y/(\sqrt k\beta\varsigma_t)+b))\).
The training inputs have latent coordinates \(aZ^{(i)}+\sqrt h\,U^\top G^{(i)}\).
The ridge targets are \(Z^{(i)}-\widehat g_G(aZ^{(i)}+\sqrt hU^\top G^{(i)})\),
where \(\widehat g_G\) is the affine Gaussian posterior mean built from the
latent sample mean and covariance; validation is analogous. Hence the fitted
\(\widehat G_t\), the selection and the validation errors are one measurable
function of
\[
 \bigl(Z^{(i)},\;U^\top G^{(i)},\;U^\top W,\;b\bigr)
\]
and of their validation counterparts, applied with the corresponding
\(k\)-dimensional procedure. For any orthonormal \(U\), the vectors
\(U^\top G^{(i)}\) are independent \(N(0,\Id_k)\). The matrix \(U^\top W\) has
independent \(N(0,1)\) entries. All are independent of each other, of the latent
data and of \(b\). The joint law of this tuple therefore depends on neither \(d\)
nor \(U\), and coincides with the case \(d=k\), \(U=\Id_k\). Log-time
interpolation forms fixed convex combinations of the fitted \(\widehat G\),
which proves item~2.

\step{Step 4: risk} By Lemma~\ref{lem:reduction}, both \(\widehat m_{\rm PCA}\) and
\(m_t\) lie in \(\spanop(U)\), so
\(\risk_t=(a/h)^2\E\norm{\widehat G_t(Y)-g_t(Y)}^2\) with
\(Y=U^\top X_t=aZ+\sqrt h\,U^\top G\). The law of \(Y\) does not depend on \(d\),
and \(Y\) is independent of \(\widehat G_t\). \qed

\subsection{Proof of Theorem~\ref{thm:generation} (generation invariance)}\label{pf:thm:generation}

At the last step,
\(\widehat x=e^{t_0}(x+h_{t_0}(a_{t_0}\widehat m(x,t_0)-x)/h_{t_0})
=e^{t_0}a_{t_0}\widehat m(x,t_0)=\widehat m(x,t_0)\), because
\(a_{t_0}e^{t_0}=1\). This proves the first claim. For the PCA estimator,
write \(y_i=U^\top x_i\) along the sampler grid. Projecting one Euler--Maruyama
step onto \(U\) gives
\[
 y_{i-1}=y_i+\Bigl(y_i+\frac2{h_{t_i}}\bigl(a_{t_i}\widehat G_{t_i}(y_i)-y_i\bigr)
 \Bigr)\Delta_i+\sqrt{2\Delta_i}\,U^\top\xi_i ,
\]
because \(U^\top\widehat m(x,t)=\widehat G_t(U^\top x)\) depends on \(x\) only
through \(y\). The recursion is closed in \(\R^k\). It starts from
\(y_N=U^\top x_T\sim N(0,\Id_k)\) and is driven by independent
\(U^\top\xi_i\sim N(0,\Id_k)\), which are independent of the fitted model.
Hence \(U^\top\widehat x=\widehat G_{t_0}(y_0)\) is a fixed measurable function
of a tuple whose law does not depend on \(d\), by Theorem~\ref{thm:invariance}.
Finally, \(\widehat x=UU^\top\widehat x\) and \(X_0=UZ\). Since
\(z\mapsto Uz\) is an isometric embedding, couplings of the latent laws and of
the embedded laws correspond one to one, with equal costs. \qed

\subsection{Proof of Corollary~\ref{cor:separation} (checkable separation)}\label{pf:cor:separation}

Fix \(U\). By Proposition~\ref{prop:fixed} and Corollary~\ref{cor:ambient}, the
ambient risk at \(U\) equals its orientation average and is at least
\(F_d(L(p+1);t)\), and also at least \(F^{\mathrm{KF}}_d(L(p+1);t)\) by
Theorem~\ref{thm:floor}. By Theorem~\ref{thm:invariance}, the PCA risk equals
\(r_k(t)\) for every \(d\) and \(U\). This proves the first statement. The
statement for all \(d\geq\max(d_0,k)\) and the case \(r_k(t)\geq F_\infty(t)\)
follow from Lemma~\ref{lem:monotone} with \(M=L(p+1)\). \qed
\section{Proofs for the covariance-matched frozen and kernel results}\label{app:cm}

\begin{roadmap}[: how Appendix~\ref{app:cm} is organized]
Both barriers in the covariance-matched regime bound the correlation between the
hidden residual and what the learner can represent. For frozen dictionaries
(\ref{pf:lem:comparison}--\ref{pf:thm:frozen}) the representation is fixed before
the orientation, so the one-sided change of measure of Appendix~\ref{app:proofs}
applies, with hypercontractivity (Lemma~\ref{lem:comparison}) replacing the density
floor that does not exist here. For kernels (\ref{pf:lem:leakage}--\ref{pf:lem:selfreg})
the span depends on the data, so the argument instead bounds the
\emph{leakage} of the residual onto each kernel section (Lemma~\ref{lem:leakage}) and
the size of the coefficients, through the ridge or through the smallest eigenvalue
of the high-degree kernel matrix (Lemma~\ref{lem:selfreg}).
\end{roadmap}

\subsection{Proof of Lemma~\ref{lem:comparison} (norm comparison)}\label{pf:lem:comparison}
(a) is immediate. (b) For each coordinate \(\psi\) of \(f\),
\(\E_P\psi^2\ge\E_Q[\rho\,\psi^2\mathbf 1\{|u^\top x|\le R\bar\sigma\}]\ge
\min_{|y|\le R}\rho\,(\E_Q\psi^2-\E_Q[\psi^2\mathbf 1\{|u^\top x|>R\bar\sigma\}])\), and by
Cauchy--Schwarz the last expectation is at most
\(\|\psi\|_{L^4(Q)}^2\pi_R^{1/2}\le\kappa^2\pi_R^{1/2}\E_Q\psi^2\). Summing over coordinates
gives the claim. The polynomial case is Gaussian hypercontractivity. \qed

\subsection{Proof of Theorem~\ref{thm:frozen} (sharp frozen exponent \(d^{k^\ast-1}\))}\label{pf:thm:frozen}
By Tweedie's formula for \(P_t\) and \(Q_t\), the residual is
\(r=m_t-m^Q_t=(h/a)\nabla\log(dP_t/dQ_t)=(h/(a\bar\sigma_t))(r_t'/r_t)(u^\top x/\bar\sigma_t)\,u\), so
\(\|r\|_P^2=B\) and \(\rho r=\hat\psi(u^\top x/\bar\sigma_t)\,u\) with \(\rho=dP_t/dQ_t\). Gaussian
integration by parts gives \(\E[r_t'h_j]=\sqrt{j+1}\,\E[r_th_{j+1}]=\sqrt{j+1}\,e_{j+1}\),
which vanishes for \(1\le j+1<k^\ast\); \(\hat\psi\) is odd because \(r_t\) is even.
\step{Part 1: lower bound} Condition on the realized span \(V\supseteq\Aff\), which is independent of
\(u\). Lemma~\ref{lem:onesided} with \(P=P_t\), \(Q=Q_t\) and the comparison constant
\(c_\kappa\) of Lemma~\ref{lem:comparison}(b) gives, for every \(u\),
\(\inf_{f\in V^d}\|r-f\|^2_P\ge B-c_\kappa^{-1}\|\Pi^Q_{V^d}(\rho r)\|^2_Q\). The field
\(\rho r=u\,\hat\psi(u^\top x/\bar\sigma_t)\) has a single nonzero output direction, so
\(\|\Pi^Q_{V^d}(\rho r)\|_Q^2=\|\Pi^Q_V\hat\psi(u^\top\cdot/\bar\sigma_t)\|_Q^2\). Averaging over
\(u\) uniform on the sphere and applying Lemma~\ref{lem:spectral} (with \(h\) replaced
by \(\bar\sigma_t^2\)) at the odd level \(J\) bounds the captured energy by
\(\Psi(1-W_J(d)+D/N_{d,J})\): the Hermite coefficients of \(\hat\psi\) below \(J\)
vanish, the part of degree \(\ge J\) captured by harmonics of degree below \(J\) is
\(\Psi(1-W_J(d))\), and the remaining eigenvalues are at most \(\Psi/N_{d,J}\) each.
The score risk is \((a/h)^2\) times the denoiser risk. \(W_J(d)\to1\) by dominated
convergence, since each \(w_{j,J}(d)\to1\) and \(w_{j,J}\le1\).
\step{Part 2: matching upper bound} The baseline is affine and \(f=\beta\,u\,h_J(u^\top x/\bar\sigma_t)\) lies in the polynomial span;
the optimal \(\beta\) reduces the denoiser risk by
\(\inner r{uh_J}_P^2/\|h_J\|_P^2=\inner{\hat\psi}{h_J}^2_{N(0,1)}/\E_Ph_J^2=G\). \qed

\subsection{Proof of Lemma~\ref{lem:leakage} (kernel leakage)}\label{pf:lem:leakage}
Under \(Q\), \(\xi=X/\bar\sigma_t\sim N(0,\Id)\), and
\(\E_P[(u^\top r)K]=\E_Q[\hat\psi(u^\top\xi)\,\varphi(\sigma_y\,\hat y^\top\xi)]\) with
\(\hat y=y/|y|\). Expand \(\hat\psi=\sum_{j\ge J}c_jh_j\) and
\(\varphi(\sigma_y\cdot)=\sum_ja_jh_j\) in \(L^2(N(0,1))\). The pair
\((u^\top\xi,\hat y^\top\xi)\) is standard Gaussian with correlation
\(\varrho=u^\top\hat y\), so Mehler's formula gives
\(\E[h_i(u^\top\xi)h_j(\hat y^\top\xi)]=\delta_{ij}\varrho^j\) and the series. By
Gaussian integration by parts, \(f=\varphi(\sigma_y\cdot)\) has
\(f^{(J)}=\sigma_y^J\varphi^{(J)}(\sigma_y\cdot)\) with Hermite coefficients
\(a_j\sqrt{j!/(j-J)!}\) at degree \(j-J\), so \(\sum_{j\ge J}a_j^2\le\|f^{(J)}\|^2/J!\).
Cauchy--Schwarz and \(|\varrho|\le1\) give
\(|\tau_u(y)|\le|\varrho|^J\sqrt\Psi\,\sigma_y^J\|\varphi^{(J)}(\sigma_yZ)\|/\sqrt{J!}\), and
\(|\varrho|\sigma_y=|u^\top y|/(d\bar\sigma_t)\). \qed

\subsection{Proof of Theorem~\ref{thm:kernel} (kernels pay \(d^{k^\ast-1}\))}\label{pf:thm:kernel}
\step{Step 1: the cross term is the only gain} \(\|r-\widehat f\|^2\ge B-2\inner r{\widehat f}_P\). Affine fields are orthogonal to \(r\)
in \(L^2(P)\): \(X-u\,u^\top X\) is independent of \(u^\top X\) and centered, and
\(\E_P[\bar r(u^\top X)p(u^\top X)]=0\) for polynomials of degree below \(J\ge2\) because
\(c_0=c_1=0\). Hence \(\inner r{\widehat f}_P=\sum_i(u^\top\alpha_i)\tau_u(x_i)=\inner{\alpha u}\tau\).
\step{Step 2: size of the coefficients} For kernel ridge regression \(\alpha u=(\mathbf K+n\lambda\Id)^{-1}Yu\). In part~1,
\(\mathbf K\succeq0\) gives \(\|\alpha u\|\le\|Yu\|/(n\lambda)\), and Cauchy--Schwarz
gives \(\E[\|Yu\|\|\tau\|]\le(n\sigma_Y^2)^{1/2}(nT_d^2)^{1/2}\). In part~2, Schoenberg's
theorem and the Schur product theorem make each \((\inner{x_i}{x_j})^{\circ j}\)
positive semidefinite, so \(\mathbf K\succeq\mathbf K_{\ge J}\succeq\gamma\Id\) on
\(E_\gamma\), and \(\|\alpha u\|\le\|Yu\|/\gamma\) for every \(\lambda\ge0\). Finally
\(\Delta\le B\) off \(E_\gamma\). The bound \(T_d^2=O(d^{-2J})\) follows from
Lemma~\ref{lem:leakage} and the moment condition. \qed

\subsection{Proof of Lemma~\ref{lem:selfreg} (self-regularization)}\label{pf:lem:selfreg}
\step{Step 1: a positive-semidefinite decomposition} Write \(z_i=u^\top x_i/\bar\sigma_t\), so that \(\inner{x_i}{x_j}/\bar\sigma_t^2=z_iz_j+\inner{q_i}{q_j}\).
For each integer \(m\), the binomial expansion of the matrix with entries
\(((z_iz_j+\inner{q_i}{q_j})/d)^m\) is a sum of positive-semidefinite matrices: its
term with power \(p\) in \(z_iz_j\) is
\(d^{-m}\binom mp\operatorname{diag}(z_i^p)[\inner{q_i}{q_j}^{m-p}]_{ij}\operatorname{diag}(z_i^p)\),
and the Schur product theorem makes the middle matrix positive semidefinite, even
when individual inner products are negative. Nonnegative Taylor coefficients
therefore give \(\mathbf K_{\ge J}\succeq\varphi_L(G^\perp)^{\circ L}\): keep the pure
perpendicular term of degree \(L\) and drop every other positive-semidefinite term.
Gershgorin's bound gives the deterministic inequality.
\step{Step 2: concentration of the Gram matrix} The \(q_i\) are independent
\(N(0,\Id_{d-1})\) vectors. With \(n=O(d^J)\), chi-square concentration and a union
bound give \(\min_iG^\perp_{ii}=1-o_{\Pr}(1)\), and Gaussian inner-product
concentration with a union bound over at most \(n^2\) pairs gives
\(\max_{i\ne j}|G^\perp_{ij}|=O_{\Pr}(\sqrt{\log d/d})\). The off-diagonal term is
therefore \(O_{\Pr}(d^{J-L/2}(\log d)^{L/2})=o_{\Pr}(1)\) since \(L>2J\), and the diagonal
term tends to one. \qed
\section{Proofs for Section~\ref{sec:training}}\label{app:training}

\begin{roadmap}[: how Appendix~\ref{app:training} is organized]
\begin{enumerate}[leftmargin=*,itemsep=1pt,topsep=1pt]
\item \emph{Lower side for one unit} (\ref{pf:prop:neuron}--\ref{pf:cor:online}):
the population loss of one unit is a polynomial in the overlap with information
exponent \(k^\ast\), so the online SGD theory of \citet{bgj2021} applies.
\item \emph{Statistical-query limit} (\ref{pf:prop:sq}): a fooling argument with
the pairwise correlations of Theorem~\ref{thm:noiseop}.
\item \emph{Upper side} (\ref{pf:thm:partialtrace}--\ref{pf:thm:twostage}): an
unbiasedness identity, a variance computation and a matrix concentration
argument for the spectral statistic, first for one view and \(k^\ast=4\), then for
\(r\) views and every even \(k^\ast\); the smoothing identity; and the assembly of
the two-stage learner.
\end{enumerate}
Each upper-side proof follows the same pattern: \emph{mean} (an exact identity
from Lemma~\ref{lem:condstein} or Theorem~\ref{thm:bijection}), \emph{variance}
(a Hermite-isometry computation), \emph{tails} (truncation and hypercontractivity),
then \emph{perturbation} (matrix Bernstein, Weyl and Davis--Kahan).
\end{roadmap}

\subsection{Proof of Proposition~\ref{prop:neuron} (one unit sees an overlap polynomial)}\label{pf:prop:neuron}
\(L(w)=\E\|r\|^2-2\E\inner r{f_w}+\E_P\|f_w\|^2\). For unit \(w\),
\(w^\top y=m\,y_u+\sqrt{1-m^2}\,g\) with \(g\sim N(0,1)\) independent of \(y_u=u^\top y\),
so \(\E_Ph_j(w^\top y)=m^je_j(t)\) by Mehler's formula. By Theorem~\ref{thm:stein},
\(\E\inner r{f_w}=-(h/a)\E_P\Stein_{Q_t}f_w\), and
\(\Stein_{Q_t}(h_J(w^\top x/\bar\sigma)w)=\bar\sigma^{-1}[h_J'(w^\top y)-(w^\top y)h_J(w^\top y)]=-\bar\sigma^{-1}\sqrt{J+1}\,h_{J+1}(w^\top y)\),
which gives the first term. Expanding \(h_J^2=\sum_{j\le2J}b_jh_j\) gives
\(\E_P\|f_w\|^2=\beta^2\sum_jb_je_jm^j\), with \(e_j=0\) for \(1\le j<k^\ast\). \qed

\subsection{Proof of Corollary~\ref{cor:online} (online SGD pays \(d^{k^\ast-1}\))}\label{pf:cor:online}
The derivatives of \(\phi\) below order \(k^\ast\) vanish at \(0\),
\(\phi^{(k^\ast)}(0)<0\) by the sign of \(\beta\), and
\(\phi'(m)=m^{k^\ast-1}(-c_1|\beta|+O(\beta^2))\) on \([0,1]\) with
\(c_1=2k^\ast\sqrt{k^\ast}(h/(a\bar\sigma_t))|e_{k^\ast}(t)|>0\), negative for small \(|\beta|\).
Thus \(\phi\) has information exponent \(k^\ast\) in the sense of
\citet[Definition~1.2]{bgj2021}. The sample is the pair \((X_0,X_t)\); the per-sample
loss \(\ell(w)=\|r-\beta h_J(w^\top y)w\|^2\) has population mean \(L_0+\phi(\inner wu)\);
let \(H_d=\ell-\E\ell\). Its Euclidean gradient is a sum of terms
\(p(w^\top y,w^\top r)\,v\) with \(p\) a polynomial of bounded degree and
\(v\in\{y,r,w\}\), minus its mean. The component along \(u\) is a polynomial in the
one-dimensional marginals \(w^\top y,w^\top r,u^\top y,u^\top r\), each a linear
combination of \(Z\) and independent Gaussians, so its second moment is bounded
uniformly in \(w\) and \(d\). Likewise \(\|y\|\) and \(\|r\|\) have \(q\)-th moments of order
\(d^{q/2}\), which gives \(\E\|\nabla H_d\|^{4+\iota}\le Cd^{(4+\iota)/2}\). These are the
conditions of \citet[Definition~1.1]{bgj2021}, and their Theorems~1.3 and~1.4 give the
two statements. \qed

\subsection{Proof of Theorem~\ref{thm:partialtrace} (one gradient step, \(k^\ast=4\))}\label{pf:thm:partialtrace}

\begin{roadmap}
Part~1 is Theorem~\ref{thm:bijection} applied to the entries of \(\mathcal H\);
part~2 is a first-order expansion of the loss. Part~3 has four steps: the exact
null second moment, a counting argument showing that the data law changes it only
at relative order \(1/d\), truncation to sub-exponential events, and matrix
Bernstein followed by Weyl and Davis--Kahan. Part~4 is one-variable calculus.
\end{roadmap}

\step{Part 1: unbiasedness} Every entry of \(\mathcal H\) is a combination of degree-four products of Hermite
polynomials, so \(L\mathcal H_{bc}=-(4/\bar\sigma^2)\mathcal H_{bc}\) for the generator of
Theorem~\ref{thm:bijection}, and \(u_q=-(\bar\sigma^2/4)\mathcal H_{bc}(x/\bar\sigma)\) for
\(q=\mathcal H_{bc}(x/\bar\sigma)\), whose gradient is \(-(\bar\sigma/4)\nabla_y\mathcal H_{bc}\).
Equation~\eqref{eq:sqcsq} and \(\E_Q\mathcal H=0\) give
\(\E_P\mathcal H=(a\bar\sigma/(4h))\E D\). Since the moments of \(y_u\) up to order three are
Gaussian and the orthogonal coordinates are standard Gaussian,
\(\E_P\He_4(y)=\E\,\He_4(y_u)\,u^{\otimes4}=\bar\kappa_tu^{\otimes4}\), whose partial trace is
\(\bar\kappa_tuu^\top\).
\step{Part 2: one gradient step} The loss is
\(\E\|r\|^2-2\sum_{bc}B_{bc}\E\inner r{\nabla_y\mathcal H_{bc}}+O(\|B\|^2)\), and
\(\inner r{\nabla_y\mathcal H_{bc}}=D_{bc}\).

\step{Part 3, step (a): null second moment} Under the null, by rotation invariance \(\E D^2=\alpha\Id\) with
\(\alpha=\E\|D\|_F^2/d\). With \(A=(\|y\|^2-d-4)(ry^\top+yr^\top)\),
\(B'=2(r^\top y)(yy^\top-\Id)\) and \(Y=\|y\|^2\sim\chi^2_d\) independent of
\(r\sim N(0,\rho^2\Id)\): \(\E\|A\|_F^2=2\rho^2(d+1)(2d^2+8d)\),
\(\E\|B'\|_F^2=4\rho^2[d(d+2)(d+4)-2d(d+2)+d^2]\) and \(\E\inner A{B'}=8\rho^2d\), so
\(\alpha=8\rho^2d^2(1+O(1/d))\) and \(\|\E X^2\|=(a\bar\sigma/(4h))^2\alpha=v_td^2(1+O(1/d))\).
\step{Part 3, step (b): the data law changes the variance only at order \(1/d\)}
Under the data law only the pair \((y_u,r_u)\) changes, and by invariance under
rotations fixing \(u\), \(\E_PD^2-\E_QD^2=\alpha_d(\Id-uu^\top)+\beta_duu^\top\). Write
\(\|y\|^2=Y_\perp+y_u^2\), \(\|r\|^2=R_\perp+r_u^2\), \(r^\top y=c_\perp+r_uy_u\), with
\(Y_\perp\sim\chi^2_{d-1}\), \(R_\perp\) of mean \((d-1)\rho^2\) and \(c_\perp\) centered of
variance \(O(d)\), all independent of \((y_u,r_u)\). Joint moments of \((y_u,r_u)\) of
total degree at most three are Gaussian, so only monomials of degree at least four
contribute to the difference. In \(\E\|D\|_F^2\), whose null value is of order \(d^3\),
such a monomial multiplies a perpendicular factor of expectation \(O(d^2)\), so
\((d-1)\alpha_d+\beta_d=O(d^2)\). In \(\beta_d=\E_P\|Du\|^2-\E_Q\|Du\|^2\), every term of
\(\|Du\|^2\) carries two factors from \(\{\|y\|^2-d-4,\ r^\top y\}\), whose perpendicular
parts are centered of size \(\sqrt d\); a monomial of degree at least four either
replaces one of them by its \(u\)-part or replaces \(Y_\perp\) or \(R_\perp\) by \(y_u^2\) or
\(r_u^2\), and loses a factor \(d\) in both cases. Hence \(\beta_d=O(d)\), \(\alpha_d=O(d)\)
and \(\|\E_PX^2\|=v_td^2(1+O(1/d))\).
\step{Part 3, step (c): truncation} Let \(\widetilde X_i\) be \(X_i\) truncated to the event
that \(|\|y\|^2-d|\), \(|\|r\|^2-d\rho^2|\) and \(|r^\top y|\) are at most
\(C\sqrt{d\log(nd/\delta)}\) and \(|y_u|,|r_u|\le C\sqrt{\log(nd/\delta)}\). By the
sub-Gaussian assumption these variables are sub-exponential, so the event fails with
probability at most \(\delta/(2n)\) per sample, the truncation bias is \(O(d^{-2})\), and
\(\|\widetilde X\|\le R=Cd^{3/2}\log^{3/2}(nd/\delta)\).
\step{Part 3, step (d): concentration and perturbation} Matrix Bernstein
\citep{tropp2012user} for the centered truncated summands gives, with probability at
least \(1-\delta\),
\(\|\frac1n\sum_iX_i-\E X\|\le O(d^{-2})+\sqrt{2\|\E X^2\|\log(4d/\delta)/n}+4R\log(4d/\delta)/(3n)\).
When the right side is below \(|\bar\kappa_t|/2\), Weyl's inequality identifies the
eigenvalue of largest modulus as the perturbed spike, and the Davis--Kahan theorem
gives \(\sin\angle(\widehat u,u)\le2\|\frac1n\sum X_i-\E X\|/|\bar\kappa_t|\le\varepsilon\) under the
displayed condition.
\step{Part 4: optimal time} With \(x=a^2\),
\(v_t/\bar\kappa_t^2\propto(1+(\sigma_0^2-1)x)^4/((1-x)x^3)\), whose logarithmic derivative
vanishes exactly at \(x=3/(3+\sigma_0^2)\). \qed

\subsection{Proof of Proposition~\ref{prop:sq} (statistical-query lower bound)}\label{pf:prop:sq}

\begin{roadmap}
(i) Each query can distinguish few orientations, because the likelihood ratios of
nearly orthogonal orientations are nearly orthogonal (Theorem~\ref{thm:noiseop});
(ii) answering every query with its null value therefore fools most orientations;
(iii) the algorithm's own output defines one more query, which a successful output
must fail to be fooled by, a contradiction; (iv) a random packing supplies many
nearly orthogonal orientations.
\end{roadmap}

\step{Step 1: each query catches few orientations} The pair \((X_0,X_t)\) is obtained from \(X_0\) by the same channel under the data and
under the null, so the pair likelihood ratio is \(D_u(x_0,x_t)=r_0(u^\top x_0/\sigma_0)\),
and by Theorem~\ref{thm:noiseop} at \(t=0\),
\(|\inner{D_u-1}{D_v-1}_Q|\le\sum_{j\ge k^\ast}e_j(0)^2|\inner uv|^j\le\beta^{k^\ast}\chi^2\)
for \(u\ne v\), and \(=\chi^2\) for \(u=v\). Fix a query \(\psi\) with \(\E_Q\psi^2\le1\) and
let \(S'\subset S\) be the set of \(u\) with \(|\E_{P_u}\psi-\E_Q\psi|>\tau\), with signs
\(\epsilon_u\). Then
\(\tau|S'|<\inner{\sum_{S'}\epsilon_u(D_u-1)}{\psi}_Q\le(|S'|\chi^2+|S'|^2\beta^{k^\ast}\chi^2)^{1/2}\),
so \(|S'|<2\chi^2/\tau^2\) whenever \(\tau^2>2\beta^{k^\ast}\chi^2\).
\step{Step 2: the null transcript} Suppose
\((q+1)2\chi^2/\tau^2<N\). Run the algorithm with every answer equal to \(\E_Q\psi\).
Fewer than \(N\) orientations are caught by its \(q\) queries, so the null transcript
is consistent with some \(P_u\); by the guarantee its output \(f_0\) satisfies
\(\|f_0\|_{L^2(Q_t)}\le\Lambda\sqrt B\).
\step{Step 3: the output as a query} Define
\(\psi_0=\bar\sigma_t\inner{x_0-m^Q_t(x_t)}{f_0(x_t)}/(\sigma_0\sqrt h\,\Lambda\sqrt B)\). Under the
null, \(X_0-m^Q_t(X_t)\) is independent of \(X_t\) with covariance
\((\sigma_0^2h/\bar\sigma_t^2)\Id\), so \(\E_Q\psi_0=0\) and \(\E_Q\psi_0^2\le1\). Since fewer than
\(N\) orientations are caught by the \(q+1\) queries \(\psi_1,\dots,\psi_q,\psi_0\), some
\(u\in S\) is fooled by all of them; the null transcript is consistent with \(P_u\), so
the algorithm outputs \(f_0\) at \(u\) and must succeed there. Success gives
\(\inner{r_u}{f_0}_{L^2(P_t)}\ge\delta B/2\), and by the tower property
\(\E_{P_u}\psi_0\ge\tau_0>\tau\), contradicting that \(u\) is fooled by \(\psi_0\).
\step{Step 4: packing} Two independent uniform unit vectors satisfy
\(\Pr(|\inner uv|>\beta)\le2e^{-(d-1)\beta^2/2}\); a union bound over \(N^2\) pairs shows
that \(N=\exp(d^c)\) independent vectors have coherence at most \(\beta=Cd^{-(1-c)/2}\)
with positive probability for a large enough \(C\). \qed

\subsection{Proof of Theorem~\ref{thm:multiview} (reused views, every even \(k^\ast\))}\label{pf:thm:multiview}

\begin{roadmap}
\emph{Mean} (part~1): Lemma~\ref{lem:condstein} turns each view into a clean
Hermite statistic, whose expectation is the cumulant spike. \emph{Variance}
(part~3): the law of total variance separates a clean part and a view part, and
both reduce to \(\E\|M_p\|_F^2\), computed by counting the cycles of a pairing graph;
a change of measure transfers the bound to the data law. \emph{Tails and
perturbation} (part~4): vector hypercontractivity controls \(\|Y_i\|\), and matrix
Bernstein with Weyl and Davis--Kahan gives the angle. Part~2 is a first-order
expansion, and part~5 follows from part~3 and Theorem~\ref{thm:time}.
\end{roadmap}

\step{Part 1: mean} Each entry \((M_p(x/\bar\sigma_t))_{bc}\) lies in the degree-\(k^\ast\) chaos of \(Q_t\), so
Lemma~\ref{lem:condstein}(1) gives \(\E[X\mid x_0]=\lambda^{k^\ast}M_p(x_0/\sigma_0)\), and
averaging over views does not change a conditional mean. Write
\(x_0/\sigma_0=\tilde zu+\xi\) with \(\tilde z=u^\top x_0/\sigma_0\) and \(\xi\) standard
Gaussian on \(u^\perp\), independent of \(\tilde z\). In the expansion of the Hermite tensor
of a sum of independent orthogonal components, every term containing a Hermite
tensor of \(\xi\) of positive degree has mean zero, so
\(\E\He_{k^\ast}(x_0/\sigma_0)=\E\He_{k^\ast}(\tilde z)\,u^{\otimes k^\ast}\). The moments of
\(\tilde z\) below order \(k^\ast\) are Gaussian, so its cumulants of orders
\(3,\dots,k^\ast-1\) vanish and \(\E\He_{k^\ast}(\tilde z)=\tilde\kappa\). The partial trace of
\(u^{\otimes k^\ast}\) over \(p\) pairs is \(uu^\top\).

\step{Part 2: one gradient step} \(\widehat L_r(B)=\widehat L_r(0)-2\sum_{bc}B_{bc}\overline{\inner r{\nabla_x(M_p(x/\bar\sigma_t))_{bc}}}+O(\|B\|^2)\),
and the average is \((h_tk^\ast/(a_t\bar\sigma_t^2))(\bar Y_n)_{bc}\).

\step{Part 3, step (a): law of total variance} Under the null, \(r\) is independent of \(X_t\) with covariance
\((\sigma_0^2h/\bar\sigma^2)\Id\). For \(F,F'\) in the degree-\(k^\ast\) chaos of a standard Gaussian
\(y\), \(\E\inner{\nabla F}{\nabla F'}=k^\ast\E FF'\), since \(-L\) acts as \(k^\ast\) on that
chaos. With \(y=X_t/\bar\sigma\) this gives \(\E X^2=(a^2\sigma_0^2/(hk^\ast))\E M_p(y)^2\), while
\(\E(\E[X\mid X_0])^2=\lambda^{2k^\ast}\E M_p(y)^2\) because \(X_0/\sigma_0\) is standard
Gaussian. The \(r\) views are conditionally independent given \(X_0\), so
\(\E Y^2=\lambda^{2k^\ast}\E M_p^2+r^{-1}(a^2\sigma_0^2/(hk^\ast)-\lambda^{2k^\ast})\E M_p^2\). By
rotation invariance \(\E M_p^2=(\E\|M_p\|_F^2/d)\Id\).
\step{Part 3, step (b): counting pairings} By the Hermite isometry,
\(\E\|M_p\|_F^2=\sum_\pi\sum_{b,c}\inner{A_{bc}}{\pi A_{bc}}\) with
\(A_{bc}=e_b\otimes e_c\otimes\Id^{\otimes p}\), and the inner sum is \(d^{\#(\pi)}\),
where \(\#(\pi)\) counts the connected components of the graph on the \(k^\ast\) slots
with edges \(\{1,\pi(1)\}\), \(\{2,\pi(2)\}\), the \(p\) trace pairs and their images
under \(\pi\). Every vertex has degree two (counting loops), so the graph is a union
of cycles; only slots \(1\) and \(2\) can form one-vertex cycles, and only if \(\pi\) fixes
them. Hence \(\#(\pi)\le p+2\), with equality exactly when \(\pi\) fixes slots \(1\) and \(2\)
and maps the trace pairing onto itself, which \(2^pp!\) permutations do, and
\(\E\|M_p\|_F^2=c_pd^{p+2}+O(d^{p+1})\).
\step{Part 3, step (c): the inflation factor} The inequality \(\iota_t\ge1\) holds because
\(\bar\sigma^{2k^\ast}=(a^2\sigma_0^2+h)^{k^\ast}\ge k^\ast(a^2\sigma_0^2)^{k^\ast-1}h\), the first
nontrivial binomial term. The identity \(\iota_t=\tilde\kappa^2/(k^\ast!S_t^2)\) follows from
Theorem~\ref{thm:time}: both sides equal
\(\bar\sigma^{2k^\ast}/(k^\ast ha^{2k^\ast-2}\sigma_0^{2k^\ast-2})\).
\step{Part 3, step (d): from the null to the data law} The vector
\((X_0,G_1,\dots,G_r)\) has density ratio \(\rho=r_0(u^\top x_0/\sigma_0)\) with respect to
the null, with \(\E_Q\rho^2=1+\chi^2\). For a unit vector \(v\), \(Yv\) is a vector of
polynomials of degree \(k^\ast\) in Gaussian variables, so Cauchy--Schwarz and
hypercontractivity give
\(\E_P|Yv|^2\le(1+\chi^2)^{1/2}(\E_Q|Yv|^4)^{1/2}\le(1+\chi^2)^{1/2}3^{k^\ast}\E_Q|Yv|^2\);
the last step uses \(\E(\sum_iF_i^2)^2\le(\sum_i\|F_i\|_4^2)^2\) and
\(\|F_i\|_4\le3^{k^\ast/2}\|F_i\|_2\).

\step{Part 4: tails, concentration and perturbation} Let \(\sigma^2\) be the bound of part~3 for \(\|\E_PY^2\|\) and
\(\eta=\lambda^{k^\ast}|\tilde\kappa|\varepsilon\). Vector-valued hypercontractivity,
\(\|\,|F|\,\|_q\le(q-1)^{k^\ast/2}\|\,|F|\,\|_2\) for \(q\ge2\) (Minkowski's inequality applied
to \(\sum_iF_i^2\) in \(L^{q/2}\)), together with Markov's inequality at
\(q\asymp\log(nd/(\delta\varepsilon))\), gives
\(\Pr_Q(\|Y\|_F>R)\le(\delta\varepsilon)^2/(4(1+\chi^2)n^2d^{2p+2})\) for
\(R=C_1\lambda^{k^\ast}(1+(\iota_t-1)/r)^{1/2}d^{(p+2)/2}\log^{k^\ast/2}(nd/(\delta\varepsilon))\),
using \(\E_Q\|Y\|_F^2=d\,\|\E_QY^2\|\). Since
\(\Pr_P(A)\le(1+\chi^2)^{1/2}\Pr_Q(A)^{1/2}\), every \(\|Y_i\|_{\rm op}\le\|Y_i\|_F\) is at most
\(R\) with probability at least \(1-\delta/2\), and the truncation bias
\(\|\E Y\mathbf 1\{\|Y\|>R\}\|\le(\E_P\|Y\|_F^2)^{1/2}\Pr_P(\|Y\|>R)^{1/2}\) is at most
\(\eta/8\) for a suitable \(C_1\). Matrix Bernstein \citep{tropp2012user} for the truncated
centered summands gives, with probability at least \(1-\delta/2\),
\(\|\bar Y_n-\E Y\|\le\eta/8+\sqrt{2\sigma^2\log(4d/\delta)/n}+2R\log(4d/\delta)/(3n)\).
Under the displayed condition on \(n\) the right side is at most \(\eta/2\). Weyl's
inequality puts one eigenvalue within \(\eta/2\) of \(\lambda^{k^\ast}\tilde\kappa\) and all
others below \(\eta/2\) in modulus, so the largest-modulus eigenvector is the perturbed
spike, and the Davis--Kahan theorem gives \(\sin\angle(\widehat w,u)\le\varepsilon\). The
factor \(\lambda^{k^\ast}\) cancels between \(\eta\), \(\sigma^2\) and \(R\).

\step{Part 5: optimal time} By part~3, \(\iota_t\propto S_t^{-2}\), which Theorem~\ref{thm:time} maximizes at
\(t^\ast\). \qed

\subsection{Proof of Proposition~\ref{prop:smoothing} (smoothing exposes the partial trace)}\label{pf:prop:smoothing}
For unit \(w\), \(u_w=(\bar\sigma/k^\ast)\He_{k^\ast}(w^\top x/\bar\sigma)\) lies in the degree-\(k^\ast\) chaos of
\(Q_t\), and \(\nabla u_w=\He_{k^\ast-1}(w^\top x/\bar\sigma)\,w\) because
\(\He_{k}'=k\He_{k-1}\). Lemma~\ref{lem:condstein}(1) gives the formula for
\(\widehat C_\infty\) on the sphere; both sides are homogeneous polynomials of degree
\(k^\ast\) in \(w\), hence equal everywhere. Expanding \(\widehat T[(w+\sigma z)^{\otimes k^\ast}]\)
by multilinearity and symmetry, the terms with \(2j\) factors of \(\sigma z\) number
\(\binom{k^\ast}{2j}\), odd moments of \(z\) vanish, and
\(\E z^{\otimes2j}=\sum_{\text{pairings}}\Id^{\otimes j}\) has \((2j-1)!!\) pairings, each
contributing \(\widehat T[w^{\otimes(k^\ast-2j)},\Id^{\otimes j}]\) by symmetry. For finite
\(r\), \(\widehat C_r(w)\) is again a homogeneous polynomial of degree \(k^\ast\) in \(w\),
linear in the symmetric tensor \(w^{\otimes k^\ast}\). The Hessian at \(0\) of its smoothing
comes only from the terms with exactly two factors of \(w\), which number
\(\binom{k^\ast}2\) and contribute
\(\binom{k^\ast}2(k^\ast-3)!!\sigma^{k^\ast-2}\,w^\top Kw\) with
\(K_{bc}=(\bar\sigma/k^\ast)\overline{\inner r{\nabla_x(M_p(x/\bar\sigma))_{bc}}}\). By the definition of
\(X\), \(K=(h/(a\bar\sigma))\bar Y_n\), and the Hessian of \(w^\top Kw\) is \(2K\). For \(\sigma=0\) and
\(k^\ast\ge3\) the Hessian at \(0\) vanishes. At fixed \(w\) the terms with \(j<p\) have
degree at least four in \(w\) and carry at most \(\sigma^{k^\ast-4}\), the term \(j=p+1\) is
constant, and the \(j=p\) term contributes the gradient
\((h/(a\bar\sigma))k^\ast(k^\ast-1)(k^\ast-3)!!\,\sigma^{k^\ast-2}\bar Y_nw\); hence, when
\(\bar Y_nw\ne0\), the normalized gradient converges to \(\bar Y_nw/|\bar Y_nw|\) as
\(\sigma\to\infty\). The power method converges from almost every start to the
eigenvector of largest-modulus eigenvalue when that eigenvalue is simple in
modulus. \qed

\subsection{Proof of Theorem~\ref{thm:twostage} (two-stage learner)}\label{pf:thm:twostage}
\begin{roadmap}
Each stage is analysed conditionally on the previous ones, which is why the three
clean samples are independent. Stage~1 needs only constant accuracy, so it is
Theorem~\ref{thm:multiview}(4) with \(\varepsilon=\frac12\). Stage~2 is a vector mean
whose expectation points exactly at \(u\) (Lemma~\ref{lem:condstein}(2)) with
\(O(d)\) variance, so Chebyshev's inequality suffices. Stage~3 is a scalar least
squares problem whose risk decomposes into a bias term, controlled by the overlap,
and a variance term free of \(d\).
\end{roadmap}

\step{Stage 1} This is Theorem~\ref{thm:multiview}(4) with \(\varepsilon=\frac12\).

\step{Stage 2} Condition on \(S_1\); then \(\widehat w_1\) is fixed with
\(m_1=|\widehat w_1^\top u|\ge\sqrt3/2\). By Lemma~\ref{lem:condstein}(2) and the mean
computation in the proof of Theorem~\ref{thm:multiview}(1),
\(\E\widehat v=\gamma\,\tilde\kappa\,(\widehat w_1^\top u)^{k^\ast-1}u\) with
\(\gamma=h\sigma_0\lambda^{k^\ast-1}/\bar\sigma^2\). Each summand is
\(r\,\He_{k^\ast-1}(\widehat w_1^\top x_t/\bar\sigma)\), averaged over views, and its second moment
is at most \((\E|r|^4)^{1/2}(\E\He_{k^\ast-1}^4)^{1/2}\le K_1d\); here \(\E|r|^4=O(d^2)\),
and \(\E\He_{k^\ast-1}^4\) is a moment of degree \(4(k^\ast-1)\) of a linear combination
of \(Z\) and Gaussians, finite by \(\E|Z|^{4(k^\ast-1)}<\infty\). By Chebyshev's
inequality, \(|\widehat v-\E\widehat v|^2\le K_1d/(\delta n_2)\) with probability at least
\(1-\delta\). The distance from \(\widehat v\) to the line \(\R u\) is at most
\(|\widehat v-\E\widehat v|\), so
\(\sin\angle(\widehat v,u)\le|\widehat v-\E\widehat v|/(|\E\widehat v|-|\widehat v-\E\widehat v|)\). For any
\(\varepsilon_2\in(0,1)\) this is at most \(\sqrt{\varepsilon_2}\) when
\(|\widehat v-\E\widehat v|\le\sqrt{\varepsilon_2}\,|\E\widehat v|/2\), in particular when
\begin{equation}\label{eq:n2}
 n_2\ge\frac{4K_1d}{\delta\varepsilon_2\gamma^2\tilde\kappa^2(3/4)^{k^\ast-1}} .
\end{equation}

\step{Stage 3} Condition on \(\widehat w\) with \(m=|\widehat w^\top u|\), let
\(\phi=h_{k^\ast-1}(\widehat w^\top x/\bar\sigma)\,\widehat w\) and
\(D(m)=\|\phi\|^2_{L^2(P_t)}=\E_Ph_{k^\ast-1}(\widehat w^\top X_t/\bar\sigma)^2\). As in the proof of
Proposition~\ref{prop:neuron},
\(\inner{m_t-m^Q_t}{\phi}_{L^2(P_t)}=(h/(a\bar\sigma))\sqrt{k^\ast}\,e_{k^\ast}(t)\,m^{k^\ast}\) and
\(D(m)=1+\sum_{j=k^\ast}^{2k^\ast-2}b_je_j(t)m^j\). For every \(\beta\),
\(\|m_t-m^Q_t-\beta\phi\|^2=B-G(m)+(\beta-\beta_\ast)^2D(m)\), with
\(G(m)=(k^\ast h^2/(a^2\bar\sigma^2))e_{k^\ast}^2m^{2k^\ast}/D(m)\) and \(\beta_\ast\) the population
minimizer. \(D\) is continuous and positive on \([0,1]\), so \(D_{\min}>0\), and
\(|D(m)-D(1)|\le K_2(1-m^2)\) with \(K_2=\sum_j\frac j2|b_je_j(t)|\), because only even
\(j\) occur (\(h_{k^\ast-1}^2\) is even) and \(1-m^j\le\frac j2(1-m^2)\) for even \(j\). If
\(m^2\ge1-\varepsilon_2\), then \(m^{2k^\ast}\ge1-k^\ast\varepsilon_2\) and
\(D(1)/D(m)\ge1-K_2\varepsilon_2/D_{\min}\), so
\begin{equation}\label{eq:Gm}
 G(m)=G_\ast\,m^{2k^\ast}\frac{D(1)}{D(m)}\ge(1-K_3\varepsilon_2)\,G_\ast,
 \qquad K_3=k^\ast+K_2/D_{\min}.
\end{equation}
The
least-squares coefficient satisfies \(\widehat\beta-\beta_\ast=\bar\xi/\bar q\) with
\(\xi=\inner{x_0-m^Q_t(x_t)}\phi-\beta_\ast|\phi|^2\) and \(q=|\phi|^2\) averaged over \(S_3\);
\(\E\xi=0\), \(\E q=D(m)\), and both variances are at most \(K_4\), because they are
polynomials of degree at most \(2(k^\ast-1)\) in \(\widehat w^\top(x_0-m^Q_t(x_t))\) and
\(\widehat w^\top x_t\), linear combinations of \(Z\) and independent Gaussians with
bounded coefficients, so their second moments involve moments of \(Z\) of order at most
\(4(k^\ast-1)\). By Chebyshev's inequality, with probability at least \(1-\delta\),
\(\bar\xi^2\le2K_4/(\delta n_3)\) and \(\bar q\ge D(m)/2\) once
\(n_3\ge8K_4/(\delta D_{\min}^2)\); on this event
\begin{equation}\label{eq:readout}
 (\widehat\beta-\beta_\ast)^2D(m)\le\frac{\bar\xi^2}{\bar q^2}D(m)\le\frac{8K_4}{\delta n_3D_{\min}} .
\end{equation}

\step{Accuracy bookkeeping} Run Stage~2 with \(\varepsilon_2=\varepsilon/(2K_3)\), which is at
most \(\varepsilon\) because \(K_3\ge k^\ast\ge4\). Condition~\eqref{eq:n2} then reads
\(n_2\ge C_3d/(\delta\varepsilon)\) with \(C_3=8K_1K_3/(\gamma^2\tilde\kappa^2(3/4)^{k^\ast-1})\),
and gives \(\sin^2\angle(\widehat w,u)\le\varepsilon_2\le\varepsilon\). By~\eqref{eq:Gm},
\(G(m)\ge(1-\varepsilon/2)G_\ast\). Take
\(C_4=\max\{16K_4/(D_{\min}G_\ast),\,8K_4/D_{\min}^2\}\); since \(\varepsilon<1\),
\(n_3\ge C_4/(\delta\varepsilon)\) satisfies the requirement
\(n_3\ge8K_4/(\delta D_{\min}^2)\) of Stage~3, and~\eqref{eq:readout} gives
\((\widehat\beta-\beta_\ast)^2D(m)\le(\varepsilon/2)G_\ast\). On the intersection of the
three events, which has probability at least \(1-3\delta\), the conditional denoiser
risk of the realized estimator given all training clean points and views is
\[
 B-G(m)+(\widehat\beta-\beta_\ast)^2D(m)\le B-\Bigl(1-\frac\varepsilon2\Bigr)G_\ast+\frac\varepsilon2G_\ast
 =B-(1-\varepsilon)G_\ast ,
\]
and the conditional score risk is \((a/h)^2\) times it. \qed

\subsection{Proof of Corollary~\ref{cor:dimsep} (dimension separation, \(k^\ast=4\))}\label{pf:cor:dimsep}
The upper bound is Theorem~\ref{thm:twostage} with \(k^\ast=4\) and fixed
\(\varepsilon,\delta\); the mixture has an \(L^2\) likelihood ratio
(Theorem~\ref{thm:noiseop}) and is sub-Gaussian. The statistical-query statement is
Proposition~\ref{prop:sq}. Theorem~\ref{thm:frozen}(1) with \(J=3\) shows that an
orientation-blind dictionary with bounded \(L^4/L^2\) constant and \(D=o(d^3)\) has
risk \((1-o(1))(a/h)^2B\); Theorem~\ref{thm:kernel}(2) with Lemma~\ref{lem:selfreg}
gives the kernel statement; Corollary~\ref{cor:online} gives the online one. \qed
\section{Experimental protocols and reproducibility}\label{app:experiments}
\label{app:reproducibility}

The studies below used fixed protocols, with seeds and resource budgets
recorded before data collection. Raw unit records for the large GPU study are
archived separately; the reported summaries were computed from those records.
The following details specify the experimental comparisons without relying on
the internal organization of the implementation.

\subsection{Covariance-visible studies}
The 168-unit main suite (base seed \(20261029\), one Tesla~T4, \(4.9\) GPU-hours)
and the 60-unit revision suite (base seed \(20261030\), 16-core CPU) used these
settings:
\begin{enumerate}[leftmargin=*]
\item \textbf{Barrier}: \(k\in\{1,2\}\), \(d\in\{8,\dots,256\}\), \(p\in\{64,256,1024\}\),
\(t\in\{0.1,0.3,0.6,1\}\), \(n=1024\), \(n_v=512\), 8 replicates, 8192 test points. A
violation is a mean risk plus three standard errors below the floor.
\item \textbf{Feature budget}: \(k=1\), \(d\in\{6,\dots,32\}\), \(p\in\{2^4,\dots,2^{14}\}\),
\(t\in\{0.3,0.6\}\), \(n=8192\), \(n_v=2048\), 3 replicates.
\item \textbf{Certificate}: PCA risk at \(d=k\), \(n\in\{64,\dots,4096\}\), 32 replicates;
per-cell and simultaneous (Bonferroni over 168 cells, Student \(t_{31}\)) upper limits;
exact first crossing \(d_0\).
\item \textbf{Generation}: three latent laws (product \(k=1\), \(k=2\), and a four-component
law in \(k=2\)), \(d\in\{8,32,128,512\}\), \(n\in\{256,1024\}\), \(p=256\), 6 replicates;
24 log-spaced fitted times in \([0.02,4]\), 400 Euler--Maruyama steps, final Tweedie step;
exact-assignment \(W_2\) on 2048 latent points.
\item \textbf{Staircase}: best-in-span least squares of the exact residual on \(131{,}072\)
inputs; Hermite dictionaries (affine, lower odd levels and a Haar-random fraction of the
next level) and tanh dictionaries of equal dimension; \(t\in\{0.3,0.6,1\}\).
\item \textbf{Data off the subspace}: \(UZ+\sigma_0G_0\), \(\sigma_0\in\{0,0.1,0.3,1\}\),
\(d\in\{8,\dots,512\}\), \(n\in\{256,1024,4096\}\), with an isotropic Gaussian baseline.
\item \textbf{Generation off the subspace}: as item~4 with \(\sigma_0\in\{0,0.1,0.3\}\).
\end{enumerate}
Exact constants were evaluated from the analytic formulas for the posterior
mean, density floor, Hermite energies, Funk--Hecke spectrum, harmonic weights,
staircase levels and time change. Checks, tables and figures were computed
from replicate-level records.

\subsection{Covariance-matched studies}
The covariance-matched studies used the following designs.
\begin{itemize}[leftmargin=*]
\item \textbf{X3}: nested quadratic Hermite dictionaries,
ridge fits on 4,096 pairs, 32,768 test pairs, eight seeds, Holm-corrected one-sided tests.
\item \textbf{X5}:
exponential-kernel ridge regression on noisy inputs, \(d\in\{8,11,16\}\), \(n\le2d^3\),
ridges \(10^{-8}\)--\(10^2\), three seeds; eigenvalue study to \(d=512\).
\item \textbf{Spectral statistics}: single-view recovery curves in units of
\(v_td^2/\bar\kappa_t^2\), four seeds; clean partial-trace overlaps at
\(n=x\cdot2d^2/\kappa_4^2\), \(x\in\{0.5,\dots,8\}\), \(d\in\{64,128,256\}\), eight seeds.
\item \textbf{X4 rate study}: 1,080 cells on two RTX~4090 GPUs under the
second, frozen protocol. The archived run records have a recorded SHA-256 digest.
\item \textbf{Two-stage learner on the X4 protocol}:
identical grid, seeds, budgets, success rule and threshold fit; variants
\(r\in\{1,\infty\}\) with zero, one or three sharpening steps on the validation split,
and a sensitivity run of the one-step full-reuse variant with 8,192-point validation
and calibration sets.
\item \textbf{X1, X2 and the landscape study}: the online, preconditioning
and reuse comparisons described in Section~\ref{sec:experiments}.
\end{itemize}
The identities of Lemma~\ref{lem:condstein},
Theorem~\ref{thm:multiview}(1,3) and Proposition~\ref{prop:smoothing} were also
checked by Monte Carlo.

\paragraph{Auditing the covariance-matched numbers.}
Each covariance-matched protocol was fixed before its study ran. The
replicate-level records of the explicit learners retain the variant,
validation and calibration size, dimension, training size, seed, normalized
gain and squared overlaps. A run succeeds when its normalized gain is at
least \(0.25\). Spectral-transition records contain one row per run. The X4
thresholds and intervals use per-cell success counts; full run histories
are archived with a recorded digest. The sensitivity run reduces each of
the validation and calibration sets to \(8{,}192\) points.

\subsection{Full result tables for the covariance-visible regime}
The tables and figures below were computed from the recorded runs without
manual adjustment.

\subsection{Results}

\paragraph{Barrier.} Across 288 Gaussian-control and ambient cells, 0 fall below the Ky Fan floor by more than three standard errors (no detected violations when this count is zero). The largest PCA/oracle denoiser difference is 3.8e-12, and the largest standardized deviation of PCA risks across $d$ is 3.12.
\begin{table}[H]\centering\small
\caption{Score risk at $t=0.3$, $k=1$: mean (standard error) over replicates, and the Ky Fan floor for the ambient dictionary.}
\begin{tabular}{rrrrrr}\toprule $d$ & $p$ & Gaussian & ambient RF & Ky Fan floor & PCA RF \\ \midrule
8 & 64 & 0.5248 & 0.2887 (0.022) & 0.01083 & 0.003556 (0.00047) \\
8 & 256 & 0.5248 & 0.1334 (0.0064) & 0.003055 & 0.002977 (0.00058) \\
8 & 1024 & 0.5248 & 0.09092 (0.0039) & 0.001052 & 0.003348 (0.00053) \\
16 & 64 & 0.5257 & 0.4923 (0.0095) & 0.05666 & 0.003046 (0.00068) \\
16 & 256 & 0.5257 & 0.3458 (0.015) & 0.01845 & 0.002585 (0.0005) \\
16 & 1024 & 0.5257 & 0.2279 (0.0066) & 0.01478 & 0.002971 (0.00077) \\
32 & 64 & 0.5243 & 0.5238 (0.0033) & 0.07273 & 0.004419 (0.00081) \\
32 & 256 & 0.5243 & 0.5045 (0.0048) & 0.06743 & 0.004306 (0.00071) \\
32 & 1024 & 0.5243 & 0.4637 (0.0053) & 0.04624 & 0.004761 (0.00077) \\
64 & 64 & 0.5228 & 0.5236 (0.0027) & 0.07683 & 0.002848 (0.00052) \\
64 & 256 & 0.5228 & 0.5231 (0.0026) & 0.07613 & 0.00249 (0.00038) \\
64 & 1024 & 0.5228 & 0.5234 (0.0024) & 0.07334 & 0.002571 (0.00036) \\
128 & 64 & 0.521 & 0.5213 (0.0019) & 0.07832 & 0.003565 (0.00064) \\
128 & 256 & 0.521 & 0.5216 (0.0019) & 0.07823 & 0.003575 (0.00062) \\
128 & 1024 & 0.521 & 0.5214 (0.0019) & 0.07787 & 0.003178 (0.00043) \\
256 & 64 & 0.5273 & 0.5274 (0.0022) & 0.079 & 0.002268 (0.00034) \\
256 & 256 & 0.5273 & 0.5277 (0.0023) & 0.07898 & 0.00252 (0.00049) \\
256 & 1024 & 0.5273 & 0.5277 (0.0021) & 0.07894 & 0.002513 (0.00054) \\
\bottomrule\end{tabular}\end{table}

\paragraph{Feature budget.} Cells below the normalized explicit barrier by more than three standard errors: 0; below the Ky Fan floor: 0. Table~\ref{tab:halfgap} gives the width $p_{1/2}$ at which the ambient estimator closes half of the Gaussian--PCA gap, and the reference width $p_N$ at which the explicit barrier vanishes. The largest Ky Fan floor in each row is below the half-gap threshold, so the barrier implies no lower bound on $p_{1/2}$; $p_N$ is a scale, not a necessary width.
\begin{table}[H]\centering\small
\caption{Half-gap width $p_{1/2}$ of the ambient estimator, reference width $p_N$, the half-gap risk threshold and the largest Ky Fan floor over the widths tested.}\label{tab:halfgap}
\begin{tabular}{rrrrrr}\toprule
$d$ & $t$ & $p_{1/2}$ & $p_N$ & half-gap threshold & largest floor \\ \midrule
6 & 0.3 & 64 & 14 & 0.262 & 0.0097 \\
8 & 0.3 & 64 & 34 & 0.262 & 0.0381 \\
12 & 0.3 & 256 & 112 & 0.263 & 0.0589 \\
16 & 0.3 & 512 & 260 & 0.263 & 0.0662 \\
24 & 0.3 & 2048 & 850 & 0.261 & 0.0718 \\
32 & 0.3 & 4096 & 1972 & 0.263 & 0.0741 \\
6 & 0.6 & 64 & 14 & 0.0285 & 0.00112 \\
8 & 0.6 & 128 & 34 & 0.0282 & 0.00762 \\
12 & 0.6 & 512 & 112 & 0.0283 & 0.0125 \\
16 & 0.6 & 1024 & 260 & 0.0285 & 0.0143 \\
24 & 0.6 & 8192 & 850 & 0.0281 & 0.0157 \\
32 & 0.6 & -- & 1972 & 0.0287 & 0.0163 \\
\bottomrule\end{tabular}\end{table}
Descriptive log-log slopes of $p_{1/2}$ in $d$: $t=0.3$: 2.68, $t=0.6$: 3.42.
\begin{figure}[H]\centering\includegraphics[width=\linewidth]{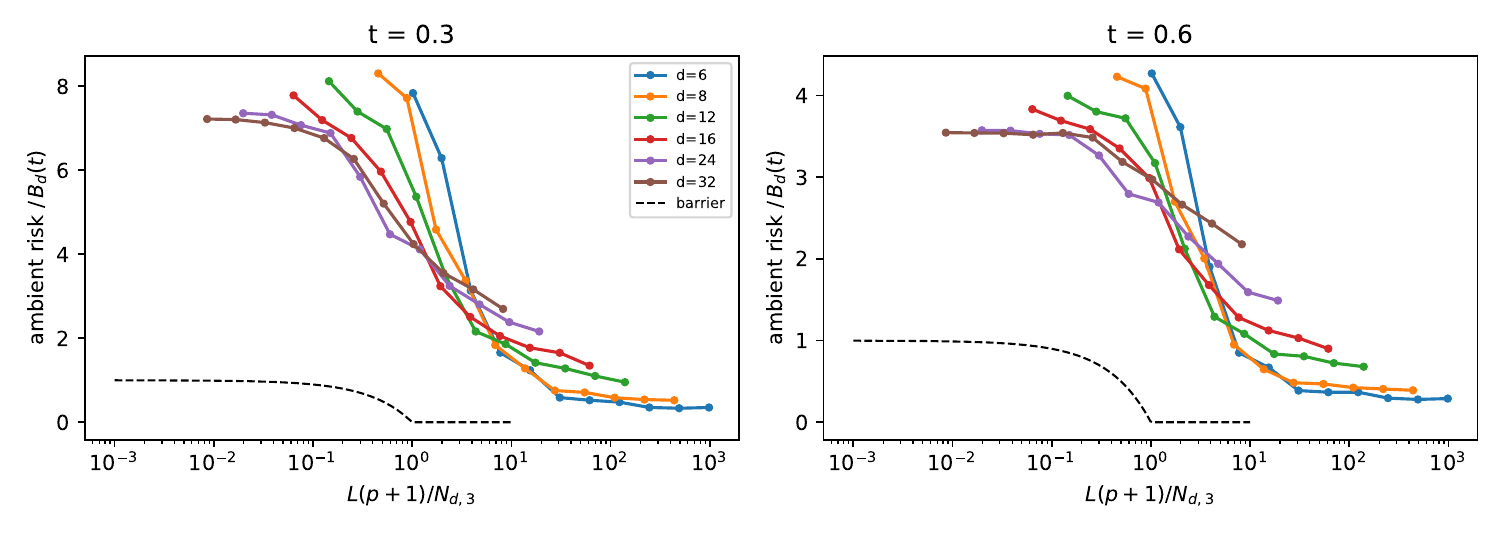}
\caption{Ambient risk normalized by $B_d(t)$ against $L(p+1)/N_{d,3}$; the dashed line is the explicit barrier.}\label{fig:width}\end{figure}

\paragraph{Certificate.} With per-cell normal limits, 102 of 168 cells admit a finite $d_0$; with simultaneous Bonferroni--$t$ limits at family level $1\%$, 100 do. In the regime $n\geq512$ ($k=1$) or $n\geq1024$ and $t\geq0.3$ ($k=2$), 75 and 75 of 75 cells are certified, with largest $d_0$ equal to 40 and 49. Table~\ref{tab:certificate} gives $d_0$ at $p=256$.
\begin{table}[H]\centering\small
\caption{Certified $d_0$ at $p=256$, as per-cell / simultaneous: for every $d\geq d_0$ the explicit floor of the ambient estimator exceeds the upper confidence limit of the PCA risk (Lemma~\ref{lem:monotone}). Per-cell limits are one-sided normal $99\%$ limits; simultaneous limits are Bonferroni--$t$ over all cells. Both are approximate. A dash means no certificate.}\label{tab:certificate}
\begin{tabular}{rrrrrr}\toprule
$k$ & $n$ & $t=0.1$ & $t=0.3$ & $t=0.6$ & $t=1.0$ \\ \midrule
1 & 64 & --/-- & 24/29 & --/-- & --/-- \\
1 & 128 & --/-- & 18/19 & 21/23 & --/-- \\
1 & 256 & 35/91 & 18/18 & 18/19 & --/-- \\
1 & 512 & 20/20 & 17/17 & 17/18 & 25/31 \\
1 & 1024 & 18/19 & 17/17 & 17/17 & 20/20 \\
1 & 2048 & 17/17 & 17/17 & 17/17 & 17/18 \\
1 & 4096 & 17/17 & 16/16 & 17/17 & 17/17 \\
2 & 64 & --/-- & --/-- & --/-- & --/-- \\
2 & 128 & --/-- & --/-- & --/-- & --/-- \\
2 & 256 & --/-- & --/-- & 40/98 & --/-- \\
2 & 512 & --/-- & 35/44 & 21/22 & --/-- \\
2 & 1024 & --/-- & 21/22 & 18/19 & 25/27 \\
2 & 2048 & --/-- & 19/19 & 18/18 & 20/21 \\
2 & 4096 & --/-- & 18/18 & 17/17 & 18/18 \\
\bottomrule\end{tabular}\end{table}

\paragraph{Generation.} The largest off-subspace RMS over all methods is 4.5e-13; the largest standardized deviation of PCA $W_2$ across $d$ is 2.73.
\begin{table}[H]\centering\small
\caption{Latent $W_2$, mean (standard error), at $d=512$ and $n=1024$; the last column is the fresh-versus-fresh reference value.}\label{tab:generation}
\begin{tabular}{lrrrrrr}\toprule latent & Gaussian & ambient RF & PCA RF & KDE & exact score & fresh \\ \midrule
four, $k=2$ & 0.473 (0.08) & 0.473 (0.08) & 0.178 (0.03) & 0.176 (0.02) & 0.125 (0.02) & 0.127 \\
product, $k=1$ & 0.662 (0.007) & 0.662 (0.007) & 0.179 (0.04) & 0.218 (0.02) & 0.112 (0.03) & 0.145 \\
product, $k=2$ & 0.929 (0.006) & 0.929 (0.006) & 0.31 (0.02) & 0.385 (0.01) & 0.279 (0.05) & 0.266 \\
\bottomrule\end{tabular}\end{table}
\begin{figure}[H]\centering\includegraphics[width=\linewidth]{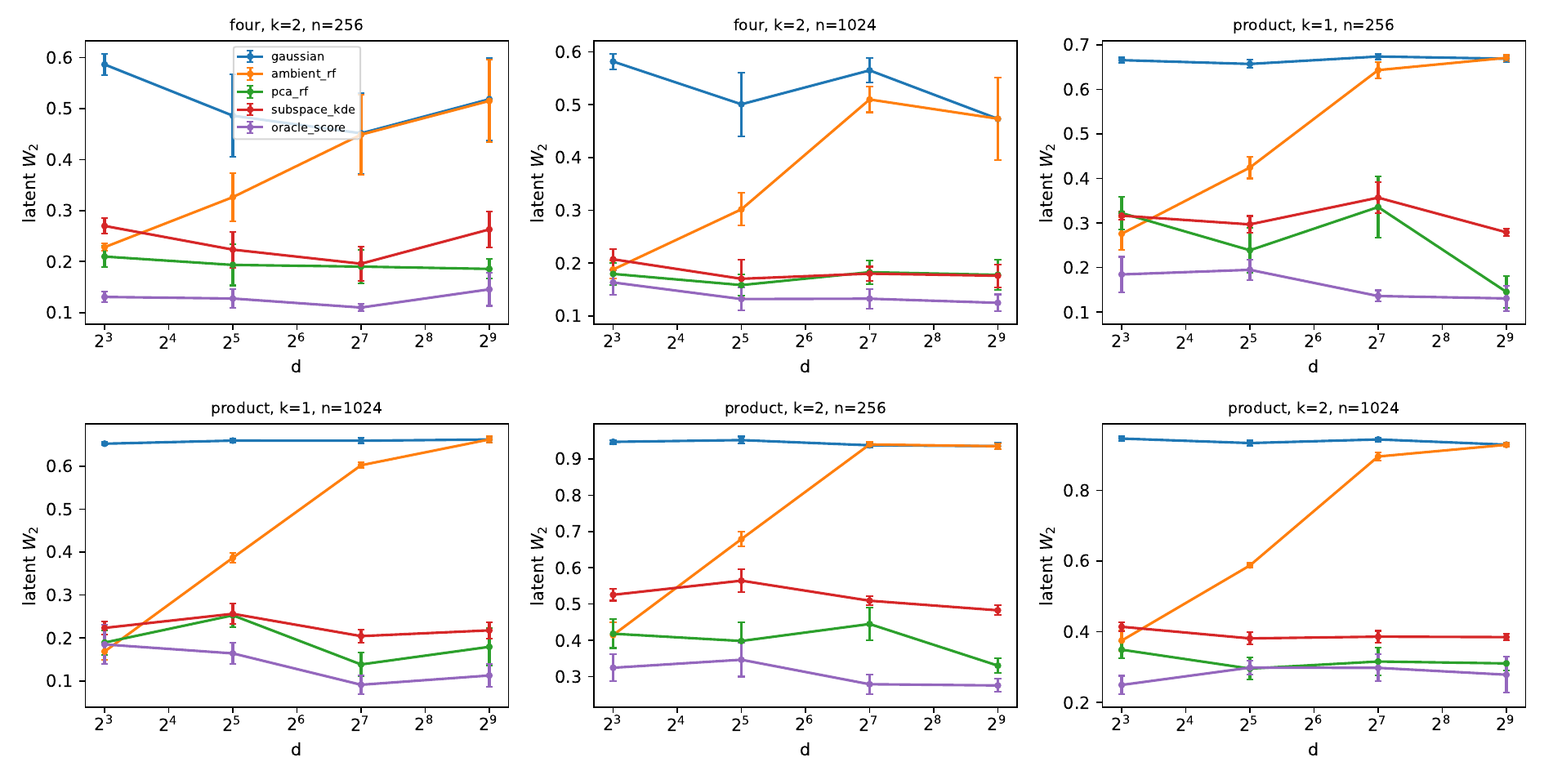}
\caption{Latent $W_2$ (mean and standard error over replicates) against $d$.}\end{figure}

\paragraph{Staircase.} Across 324 best-in-span cells, 0 fall below the Ky Fan floor by more than three standard errors.
\begin{table}[H]\centering\small
\caption{Best-in-span score risk (mean over replicates) of the full Hermite dictionary at level $J$ (dimension $\binom{d+J}{J}$) and of tanh random features of the same dimension, against the dimension-free level $k(a/h)^2\varepsilon_J$ of Theorem~\ref{thm:sharp}.}\label{tab:staircase}
\begin{tabular}{rrrrrrr}\toprule $t$ & $J$ & $d$ & dimension & Hermite & tanh & $k(a/h)^2\varepsilon_J$ \\ \midrule
0.3 & 3 & 6 & 63 & 0.186 & 0.211 & 0.189 \\
0.3 & 3 & 8 & 129 & 0.189 & 0.152 & 0.189 \\
0.3 & 3 & 12 & 377 & 0.189 & 0.147 & 0.189 \\
0.3 & 3 & 16 & 833 & 0.193 & 0.163 & 0.189 \\
0.3 & 3 & 24 & 2625 & 0.192 & 0.193 & 0.189 \\
0.3 & 3 & 32 & 6017 & 0.196 & 0.19 & 0.189 \\
0.3 & 5 & 6 & 315 & 0.0861 & 0.0487 & 0.0878 \\
0.3 & 5 & 8 & 921 & 0.0868 & 0.0472 & 0.0878 \\
0.3 & 5 & 12 & 4745 & 0.0918 & 0.053 & 0.0878 \\
0.6 & 3 & 6 & 63 & 0.0123 & 0.0274 & 0.0127 \\
0.6 & 3 & 8 & 129 & 0.0127 & 0.0178 & 0.0127 \\
0.6 & 3 & 12 & 377 & 0.0125 & 0.0174 & 0.0127 \\
0.6 & 3 & 16 & 833 & 0.0132 & 0.0196 & 0.0127 \\
0.6 & 3 & 24 & 2625 & 0.0125 & 0.0229 & 0.0127 \\
0.6 & 3 & 32 & 6017 & 0.0129 & 0.0222 & 0.0127 \\
0.6 & 5 & 6 & 315 & 0.00372 & 0.00352 & 0.00378 \\
0.6 & 5 & 8 & 921 & 0.00378 & 0.00363 & 0.00378 \\
0.6 & 5 & 12 & 4745 & 0.00395 & 0.00396 & 0.00378 \\
1.0 & 3 & 6 & 63 & 0.000283 & 0.00176 & 0.0003 \\
1.0 & 3 & 8 & 129 & 0.000309 & 0.00116 & 0.0003 \\
1.0 & 3 & 12 & 377 & 0.000292 & 0.00115 & 0.0003 \\
1.0 & 3 & 16 & 833 & 0.000322 & 0.00128 & 0.0003 \\
1.0 & 3 & 24 & 2625 & 0.000285 & 0.0015 & 0.0003 \\
1.0 & 3 & 32 & 6017 & 0.000288 & 0.00144 & 0.0003 \\
1.0 & 5 & 6 & 315 & 4.18e-05 & 0.00013 & 4.2e-05 \\
1.0 & 5 & 8 & 921 & 4.38e-05 & 0.000122 & 4.2e-05 \\
1.0 & 5 & 12 & 4745 & 4.37e-05 & 0.000122 & 4.2e-05 \\
\bottomrule\end{tabular}\end{table}
\begin{figure}[H]\centering\includegraphics[width=\linewidth]{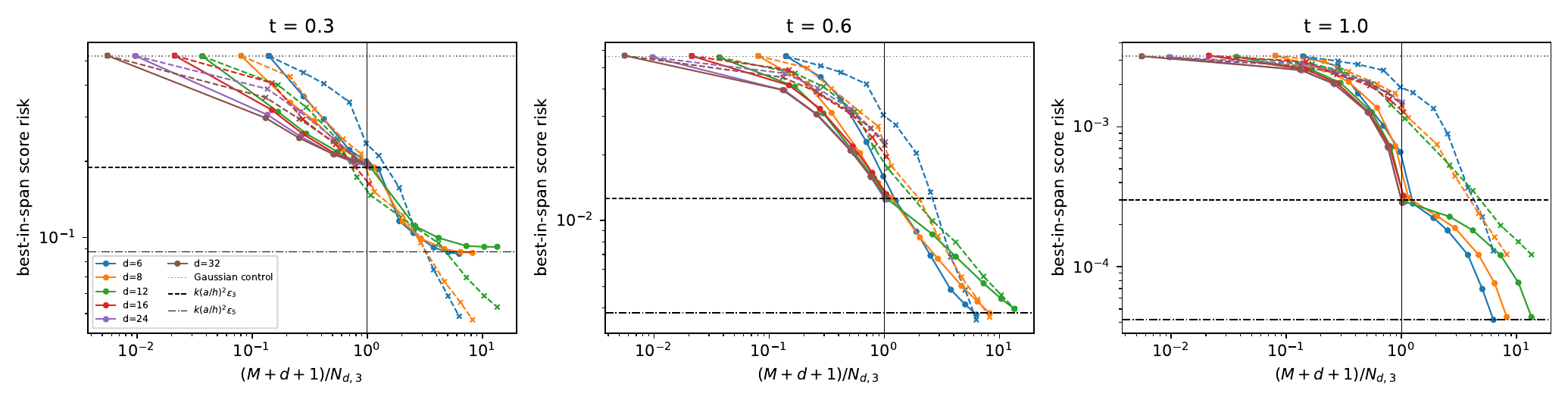}
\caption{Best-in-span score risk of orientation-blind dictionaries against the normalized dimension $(M+d+1)/N_{d,3}$. Circles: affine functions, full lower Hermite levels and a uniformly random subspace of the next level; crosses: tanh random features of the same dimension. Horizontal lines: the Gaussian control and the dimension-free levels of Theorem~\ref{thm:sharp}; the vertical line marks $N_{d,3}$.}\label{fig:staircase}\end{figure}

\paragraph{Data off the subspace.} Across 384 Gaussian-control and ambient cells, 0 fall below the time-changed Ky Fan floor of Proposition~\ref{prop:offsubspace} by more than three standard errors. With $\sigma_0=0$ the largest standardized deviation of PCA risks across $d$ is 2.4.
\begin{table}[H]\centering\small
\caption{Data $UZ+\sigma_0G_0$, $k=1$, $p=256$, $t=0.3$: subspace error $\norm{\sin\Theta}_F^2$ (measured / spiked-covariance reference) and score risks (mean). The barrier column is $\lambda_t^{-2}F^{\mathrm{KF}}_d(L(p+1);t')$.}\label{tab:noisy}
\begin{tabular}{rrrrrrrrr}\toprule $\sigma_0$ & $d$ & $n$ & $\sin^2\Theta$ & reference & Gaussian & ambient & barrier & PCA \\ \midrule
0.1 & 8 & 256 & 8.6e-05 & 0.00013 & 0.513 & 0.169 & 0.00298 & 0.00855 \\
0.1 & 128 & 256 & 0.0019 & 0.0021 & 0.539 & 0.54 & 0.077 & 0.0367 \\
0.1 & 512 & 256 & 0.0083 & 0.0084 & 0.619 & 0.622 & 0.0781 & 0.116 \\
0.1 & 8 & 4096 & 5.2e-06 & 8.3e-06 & 0.508 & 0.12 & 0.00298 & 0.000892 \\
0.1 & 128 & 4096 & 0.00014 & 0.00013 & 0.512 & 0.512 & 0.077 & 0.00251 \\
0.1 & 512 & 4096 & 0.00054 & 0.00053 & 0.515 & 0.515 & 0.0781 & 0.00736 \\
0.3 & 8 & 256 & 0.00076 & 0.0012 & 0.424 & 0.16 & 0.00242 & 0.0177 \\
0.3 & 128 & 256 & 0.018 & 0.019 & 0.602 & 0.603 & 0.0676 & 0.21 \\
0.3 & 512 & 256 & 0.073 & 0.073 & 1.16 & 1.17 & 0.0686 & 0.807 \\
0.3 & 8 & 4096 & 5e-05 & 7.7e-05 & 0.413 & 0.101 & 0.00242 & 0.00143 \\
0.3 & 128 & 4096 & 0.0013 & 0.0012 & 0.428 & 0.428 & 0.0676 & 0.0137 \\
0.3 & 512 & 4096 & 0.005 & 0.0049 & 0.462 & 0.462 & 0.0686 & 0.0514 \\
1.0 & 8 & 256 & 0.011 & 0.018 & 0.0974 & 0.0904 & 0.000303 & 0.0439 \\
1.0 & 128 & 256 & 0.23 & 0.25 & 0.619 & 0.62 & 0.0175 & 0.614 \\
1.0 & 512 & 256 & 0.65 & 0.65 & 2.03 & 2.03 & 0.0178 & 2.03 \\
1.0 & 8 & 4096 & 0.00095 & 0.0012 & 0.0671 & 0.029 & 0.000303 & 0.00352 \\
1.0 & 128 & 4096 & 0.019 & 0.018 & 0.102 & 0.102 & 0.0175 & 0.0412 \\
1.0 & 512 & 4096 & 0.073 & 0.071 & 0.211 & 0.211 & 0.0178 & 0.159 \\
\bottomrule\end{tabular}\end{table}
\begin{figure}[H]\centering\includegraphics[width=\linewidth]{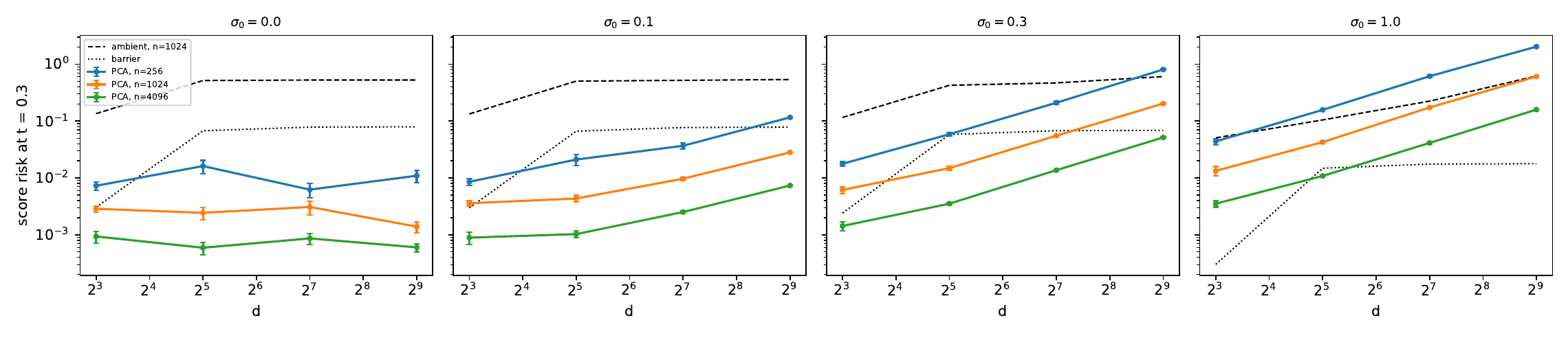}
\caption{Score risk at $t=0.3$ against $d$ for data off the subspace. PCA random features for three training sizes, ambient random features at $n=1024$ and the time-changed barrier for the ambient dictionary.}\label{fig:noisy}\end{figure}

\begin{table}[H]\centering\small
\caption{Latent $W_2$ of generated samples (mean over replicates) for data off the subspace, $k=1$, $n=1024$; the reference law is that of $U^\top X_0^\sigma=Z+\sigma_0\xi$.}\label{tab:noisygen}
\begin{tabular}{rrrrrrrr}\toprule $\sigma_0$ & $d$ & Gaussian & ambient RF & PCA RF & KDE & exact score & fresh \\ \midrule
0.0 & 32 & 0.67 & 0.402 & 0.267 & 0.3 & 0.131 & 0.127 \\
0.0 & 512 & 0.653 & 0.653 & 0.326 & 0.284 & 0.297 & 0.16 \\
0.1 & 32 & 0.66 & 0.407 & 0.263 & 0.291 & 0.124 & 0.122 \\
0.1 & 512 & 0.643 & 0.642 & 0.302 & 0.266 & 0.286 & 0.156 \\
0.3 & 32 & 0.593 & 0.449 & 0.202 & 0.244 & 0.0907 & 0.0938 \\
0.3 & 512 & 0.574 & 0.573 & 0.223 & 0.218 & 0.225 & 0.126 \\
\bottomrule\end{tabular}\end{table}

\paragraph{Staircase.} No best-in-span cell falls below the Ky Fan floor by more
than three standard errors (\(0\) of \(324\)). The full level-\(J\) Hermite
dictionary reaches the dimension-free level of Theorem~\ref{thm:sharp} at every
tested \(d\) (Table~\ref{tab:staircase}): for \(J=3\) its risk lies within
\(8\%\) of \(k(a_t/h_t)^2\varepsilon_3(t)\) from \(d=6\) to \(d=32\) at all three
times, for example \(0.00029\), \(0.0129\) and \(0.196\) at \(d=32\) and
\(t=1,0.6,0.3\) against \(0.00030\), \(0.0127\) and \(0.189\); for \(J=5\) it lies
within \(5\%\) of \(k(a_t/h_t)^2\varepsilon_5(t)\) for \(d\leq12\). The dictionary
dimension needed to reach the level grows from \(63\) to \(6017\) over this range,
as \(N_{d,3}\) does, while the attained risk does not change. The approach to the
level is convex rather than linear in the fraction \(f\) of level three that the
dictionary contains: at \(f=\frac12\) the remaining fraction of the
level-three excess is \(0.25\)--\(0.36\) at \(t=1\), \(0.17\)--\(0.24\) at
\(t=0.6\) and \(0.05\)--\(0.11\) at \(t=0.3\). This is compatible with the lower
bound, which is linear in \(f\) but carries the density factor \(c_t^k\). At
\(t=1\) and \(t=0.6\) the normalized curves for \(d\in\{8,\dots,32\}\) coincide
to within \(0.05\) and \(0.07\), so that the relevant scale of the dictionary is
\(N_{d,3}\) and not \(d\) or \(d^2\). Tanh random features of the same dimension
behave differently at different times. At \(t=1\) and \(t=0.6\) they stay above
the cubic Hermite dictionary, by a factor between about \(4\) and \(6\) at
\(t=1\), and the gap does not close with \(d\). At \(t=0.3\) and
\(8\leq d\leq16\) they fall below the
level-three value (\(0.147\) against \(0.189\) at \(d=12\)), and at \(J=5\) even
below the level-five value, because their span contains components of degree
above \(J\), which carry a large part of the residual at small \(t\). At
\(d\geq24\) this advantage disappears, as a span of dimension \(N_{d,3}\) is a
vanishing fraction of the next level, \(N_{d,5}\). The upper bound concerns
polynomial dictionaries only, so none of these measurements conflicts with it.

\paragraph{Data off the subspace.} No Gaussian-control or ambient cell falls below
the time-changed Ky Fan floor of Proposition~\ref{prop:offsubspace} by more than
three standard errors (\(0\) of \(384\)), and with \(\sigma_0=0\) the PCA risks
again vary across \(d\) only within Monte Carlo error (largest standardized
deviation \(2.4\)). The measured \(\norm{\sin\Theta}_F^2\) agrees with the
spiked-covariance reference once \(d\geq32\) (Table~\ref{tab:noisy}). Ambient
random features show no measurable improvement over the Gaussian control once
\(d\geq128\), at every \(\sigma_0\), \(n\) and \(t\). With \(\sigma_0>0\) every
method's risk grows with \(d\) at fixed \(n\), because the shared baseline
estimates a \(d\)-dimensional mean and an off-subspace variance from noisy data.
This growth is common to the Gaussian control, the PCA estimator and the
oracle-projection estimator: PCA and oracle risks differ by a median of
\(1.3\%\) over the \(144\) cells with \(\sigma_0>0\), and coincide at
\(\sigma_0=0\). The part of the risk that the
barrier concerns, the improvement over the Gaussian control, is nearly
independent of \(d\). At \(t=0.3\) the PCA estimator's absolute improvement varies
by at most \(13\%\) of its \(d=8\) value across \(d\in\{8,\dots,512\}\) for \(\sigma_0\leq0.3\) at every \(n\), and
for \(\sigma_0=1\) it stays within \(20\%\) of its \(d=8\) value while
\(d\sigma_0^2/n\leq1/8\). It collapses once \(d\sigma_0^2/n\geq1/2\), where
\(\norm{\sin\Theta}_F^2\geq0.23\): there the PCA estimator retains at most a
quarter of its \(d=8\) improvement, while the oracle projection retains
\(30\)--\(80\%\). This is the only regime in the grid in which subspace
estimation, rather than feature learning or the baseline, limits the adaptive
estimator. Every cell lies above the detection threshold
\(\lambda/\sigma_0^2>\sqrt{d/n}\) of the spiked-covariance model, with
\(\lambda=\mu^2+\tau^2\), so the collapse is caused by estimation error, not by a
loss of detectability; we do not bound it. In generation
(Table~\ref{tab:noisygen}), PCA random features at \(d=512\) and
\(\sigma_0\in\{0.1,0.3\}\) reach \(W_2\) within \(0.02\) of the exact-score
sampler, and ambient random features again reproduce the Gaussian control. With
three replicates the standard errors of these \(W_2\) values are about
\(0.03\)--\(0.05\); the difference between the exact-score values at \(d=32\) and
\(d=512\) lies within this spread, and the six-replicate main study shows no
trend in \(d\).

\end{document}